\documentclass[10pt,twocolumn,letterpaper]{article}

\usepackage[accsupp]{axessibility} 
\usepackage{cvpr}              
\usepackage{tabularx}
\usepackage{graphicx}
\usepackage{pifont}
\usepackage{wrapfig}
\usepackage{multirow}
\usepackage{makecell}
\usepackage{multirow}
\usepackage{mwe}
\usepackage{amsmath}
\usepackage[export]{adjustbox}
\usepackage{dashrule}
\usepackage{tikz}
\usepackage{arydshln}
\usepackage{xcolor}
\usepackage{titletoc}

\definecolor{cvprblue}{rgb}{0.21,0.49,0.74}
\usepackage[pagebackref,breaklinks,colorlinks,citecolor=cvprblue]{hyperref}

\def\paperID{10648} 
\def\confName{CVPR}
\def\confYear{2024}
 
\newcommand{\modelname}{ESR-NeRF\xspace}
\title{\modelname: Emissive Source Reconstruction Using LDR Multi-view Images}

\author{
    Jinseo Jeong\textsuperscript{1} \quad 
    Junseo Koo\textsuperscript{1} \quad 
    Qimeng Zhang\textsuperscript{2} \quad 
    Gunhee Kim\textsuperscript{1} \\ 
    \textsuperscript{1}Seoul National University \quad 
    \textsuperscript{2}Korea University \\
    {\tt\small jinseo@vision.snu.ac.kr, junseo.koo@vision.snu.ac.kr, zoe1024@korea.ac.kr,  gunhee@snu.ac.kr} \\
    {\small \url{https://jinseo.kr/ESR-NeRF}}
    }

\begin{document}
\maketitle
\begin{abstract}
    Existing NeRF-based inverse rendering methods suppose that scenes are exclusively illuminated by distant light sources, neglecting the potential influence of emissive sources within a scene.
    In this work, we confront this limitation using LDR multi-view images captured with emissive sources turned on and off.
    Two key issues must be addressed: 1) ambiguity arising from the limited dynamic range along with unknown lighting details, and 2) the expensive computational cost in volume rendering to backtrace the paths leading to final object colors.
    We present a novel approach, \modelname, leveraging neural networks as learnable functions to represent ray-traced fields.
    By training networks to satisfy light transport segments, we regulate outgoing radiances, progressively identifying emissive sources while being aware of reflection areas.
    The results on scenes encompassing emissive sources with various properties demonstrate the superiority of \modelname in qualitative and quantitative ways.
    Our approach also extends its applicability to the scenes devoid of emissive sources, achieving lower CD metrics on the DTU dataset.
\end{abstract}
\vspace{-15pt}
\section{Introduction}
\label{sec:intro}
Extensive research has focused on reconstructing 3D object structures~\cite{Park_2019_CVPR,DVR,idr2020,ge2023ref}, material properties~\cite{li2020inverse,refnerf2022,hadadan2023inverse}, and lighting~\cite{liang2023envidr,DE-NeRF,hweberEditableIndoorLight,Garon_2019_CVPR,luan2021unified} from 2D images, applicable across domains including 3D graphics and augmented reality~\cite{ARAH2022,wang2022neural,sotarendering2020,advancesrendering2022}.
This endeavor not only facilitates the creation of life-like virtual objects but also streamlines the process of scene manipulation~\cite{lumigraph2021,sun2022neuconw,wang2023fegr,tang2022nerf2mesh}.
Recent advancements~\cite{ji2022relight,physically2022,wang2021learning,lyu2023dpi} have built on Neural Radiance Fields (NeRF)~\cite{nerf2020} successes in novel view synthesis~\cite{nerf++2020,mipnerf2021,360mipnerf2022,xiangli2022bungeenerf,pointlight2022}.
Significant progress in re-lighting~\cite{nerfw2020,nerfosr2022,mai2023neural} has facilitated scene editing via manipulating the reconstructed light sources.
However, existing methods predominantly deal with the scenes lit by distant sources, like environment maps or collocated flashlights.
Notably, NeRF-based inverse rendering has yet to consider scenes with multiple emissive sources, a common real-world illumination condition.

\begin{figure}[t]
    \centering
    \includegraphics[width=\linewidth]{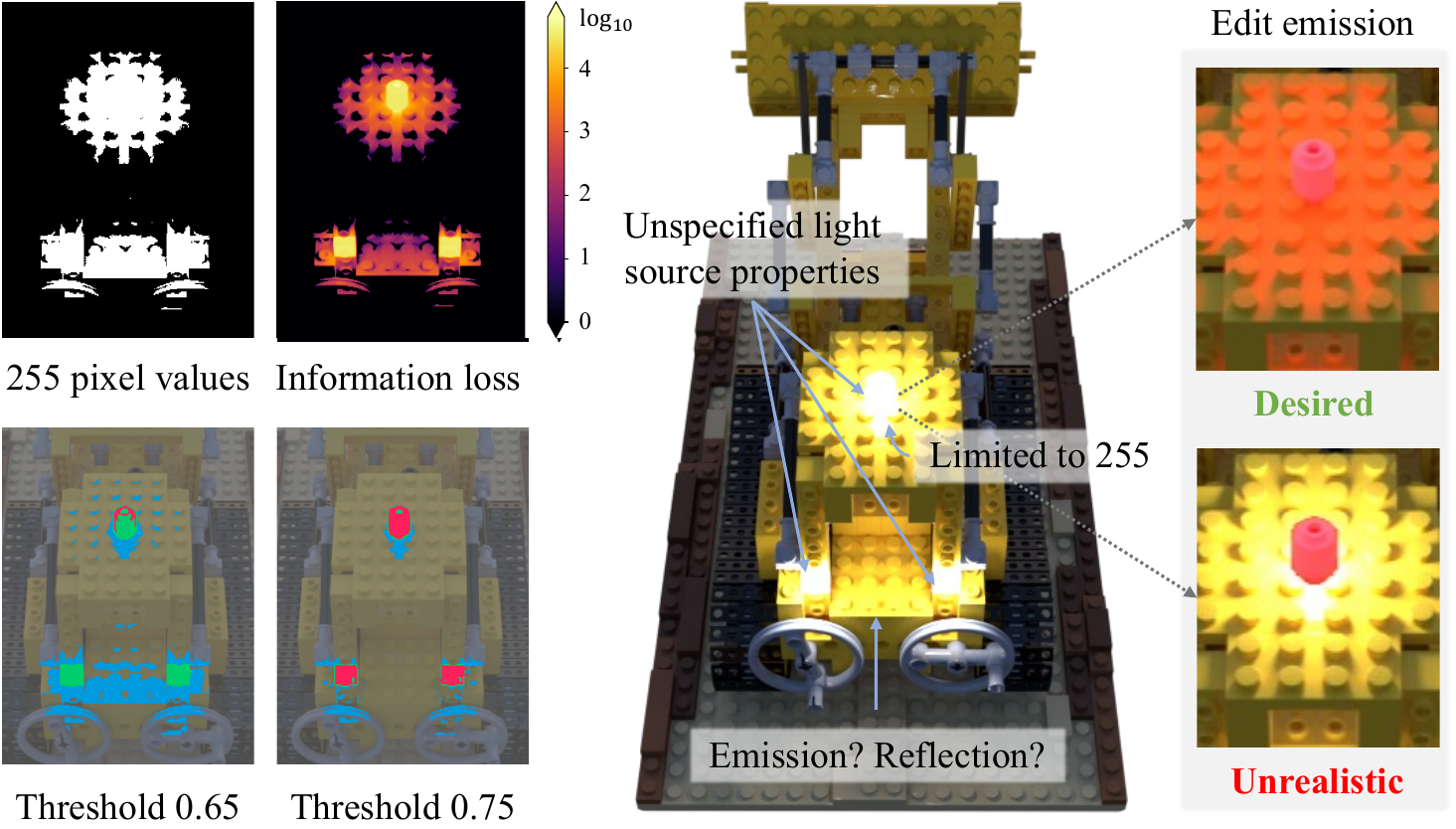}

    \caption{
        Challenges posed by emissive sources in LDR images.
        Green, red, and blue in thresholded images respectively show true positives, false negatives, and false positives of source identification.
        Thresholding values are scaled down divided by 255.
        The contrast between light on and off pixel values is more pronounced in surroundings than emissive sources.
        Inaccurate reconstruction of emissive sources disrupts scene editing, causing reflection areas to stay static while only the source colors change.
    }
    \label{fig:problem formulation}
    \vspace{-15pt}
\end{figure}

Emissive sources in a scene introduce critical challenges: (i) ambiguity in decomposing scene components and (ii) high computational costs for analyzing the causes of pixel colors.
This ambiguity stems from difficulties in identifying emissive source regions, as illustrated in Fig.~\ref{fig:problem formulation}.
Contrary to prior setups~\cite{nerd2021,neural-incident-light-field,nemto,iron-2022,reflectfields2020,twoshot2020}, we allow the possibility of numerous emissive sources throughout the scene.
In standard photographs with pixel values from 0 to 255, the distinction between emissive sources and nearby reflection areas is challenging.
As shown in Fig.~\ref{fig:problem formulation}, relying solely on pixel value thresholding is insufficient for differentiating between emissive sources and their reflections.
Naive inverse path tracing is impractical, due to the computational costs rising exponentially with the number of ray bounces in volume rendering.
This can cause inaccuracy in emissive source reconstruction, yielding unrealistic illumination in reflective areas as users manipulate emissive sources.

To address these challenges, we introduce \modelname (Emissive Sources Reconstructing NeRF), a novel approach capable of reconstructing any number of emissive sources by progressively discovering reflection areas. 
We assume that the scenes are observed in two lighting conditions: one with all emissive sources active and the other with them inactive.
Our approach utilizes neural networks as learnable functions for representing ray-traced fields.
By training networks to satisfy each light transport segment, we sidestep the computational overhead of ray tracing associated with ray bounces.
In this work, we exclusively use low dynamic range (LDR) images, setting us apart from prior mesh-based methods that rely on high dynamic range (HDR) images~\cite{fipt2023,freeviewpoint2021,Recovering2021,inversepathtracing2019}.

Our experiments encompass synthetic and real scenes, ranging from single to multiple lighting configurations with complex reflections.
The scenes vary in light source counts, color, and intensity.
Qualitative and quantitative evaluations show \modelname's superiority over state-of-the-art NeRF-based re-lighting methods.
Furthermore, Chamfer Distance (CD) metrics on the DTU dataset~\cite{DTU} indicate \modelname's competitive performance in scene reconstruction, even without emissive sources. 

We summarize our contributions as follows.
\begin{enumerate}
    \item Our work presents the first NeRF-based inverse rendering that can deal with the scenes with any number of emissive sources, challenging the distant light assumption of previous research.
    \item Unlike existing mesh-based methods relying on HDR images, we use LDR images for the first time, overcoming the poor representation of emissive sources.
    \item We provide a benchmark dataset designed to evaluate the performance of emissive source reconstruction.
    \item Our method is applicable to the scenes with or without emissive sources, achieving superior mesh reconstruction results on the DTU dataset.
\end{enumerate}

\section{Related work}
\label{sec:relatedwork}

\textbf{Neural Rendering}.
Advancements in implicit representations \cite{sitzmann2019siren,tancik2020fourfeat} and volume rendering~\cite{Max} have significantly enhanced neural rendering capabilities, enabling the reconstruction of scene components from 2D images.
One of the key directions is mesh extraction~\cite{wu2022object,wu2023objsdfplus,zhang2023unbiased,neus2,Oechsle2021ICCV, wu2022voxurf}, with methods like NeuS~\cite{wang2021neus} and VolSDF~\cite{volsdf2021} utilizing signed distance function (SDF) values for volume rendering.
Recently, the efficient computation of volume rendering has become a focal point due to the substantial computational cost associated with network inference for ray color calculation~\cite{mueller2022instant,terminerf2021,yu2021plenoctrees}.
Several methods propose to directly predict ray color using the 4D light fields concept~\cite{rayspace2022,lightfield2022,lightfield2021} or leveraging voxel grids for fast inference of spatial features~\cite{tensorf2022,plenoxels2021,dvgo2022,barron2023zipnerf,li2023neuralangelo,cai2023neuda}.
NeuralRadiosity~\cite{radiosity2021} shares similarity with our method, as it predicts ray-traced values instead of explicitly tracing individual rays.
However, they primarily focus on calculating the final object color when all scene information is available.
In contrast, our inverse rendering approach aims to reconstruct emissive sources within a scene, addressing the ambiguities introduced by their presence in LDR images.

\vspace{5pt}
\noindent
\textbf{Inverse Rendering}.
A growing emphasis revolves around the decomposition of materials represented by spatially varying bidirectional reflectance distribution functions (SVBRDF)~\cite{psnerf2022,zhu2022learning,totalrelighting2021}.
To lessen the computational burden in inverse rendering~\cite{nerv2021,media2021,modelingindirect2022,Jin2023TensoIR}, several methods have adopted neural networks as lookup tables~\cite{neuralpil2021} or computational caches~\cite{nerv2021,nerfactor2021,zhang2023neilf++}.
While NeRV~\cite{nerv2021} utilizes caching visibility and NeILF++~\cite{zhang2023neilf++} adopts caching surface point radiance with the inter-reflection loss for incident radiance, 
our method diverges by focusing on tracing radiance origins. Specifically, we aim to identify emissive sources within a scene, moving beyond the simplification of incident radiance calculations.
Several methods rely on diverse known lighting configurations to exploit variations in object appearances~\cite{nelf2021,yang2022s3nerf,Toschi_2023_CVPR,zeng2023nrhints}.
Toggling emissive sources on and off resembles the common one-light-at-a-time (OLAT) technique, as seen in NLT~\cite{lighttransport2021} and ReNeRF~\cite{xu2023renerf}. However, our setting does not need to know light source properties and to toggle lights individually. 
Instead, we allow for toggling all lights together.
Recent works have also jointly reconstruct the mesh, materials, and lighting~\cite{sun2023neuralpbir,Munkberg_2022_CVPR,hasselgren2022nvdiffrecmc,lyu2022nrtf}.
They tackle with images captured under a single unknown lighting condition~\cite{physg2021,nerfactor2021}, assuming that radiance already encodes global illumination~\cite{modelingindirect2022,wu2023nefii}.
However, they confine to the scenes illuminated by far-distant lights, constrained to an 8-bit color spectrum.
Our work considers the presence of multiple emissive sources within a scene captured in LDR images, questioning the prevailing notion that radiance fields trained with the image rendering loss faithfully represents global illumination.
While some methods~\cite{fipt2023,freeviewpoint2021,li2022texir,Recovering2021,inversepathtracing2019} deal with the scenes featuring emissive sources, they work outside the volume rendering framework and depend on HDR input images, assuming prior knowledge of scene geometry.

\begin{table}
    \scriptsize
    \centering
    \begin{tabular}{c|ccc|c}
        \toprule
                                & Voxurf    & TensoIR      & Path Tracing       & \modelname   \\
        \midrule
        Big O                   & n         & $n \cdot  d$ & $(n\cdot d)^{b+1}$ & $n^2\cdot d$ \\
        Indirect illumination   & \ding{56} & \ding{52}    & \ding{52}          & \ding{52}    \\
        BRDF decomposition      & \ding{56} & \ding{52}    & \ding{52}          & \ding{52}    \\
        Emissive source control & \ding{56} & \ding{56}    & \ding{52}          & \ding{52}    \\
        \bottomrule
    \end{tabular}
    \caption{
        Computational cost comparison for inverse rendering methods.
        $n$ is the number of sampled points along a ray, $d$ is the number of scattering rays, and $b$ is the number of ray bounces.
    }
    \label{tab:computational cost}
    \vspace{-15pt}
\end{table}
\begin{figure*}[t!]
    \centering
    \includegraphics[width=\textwidth]{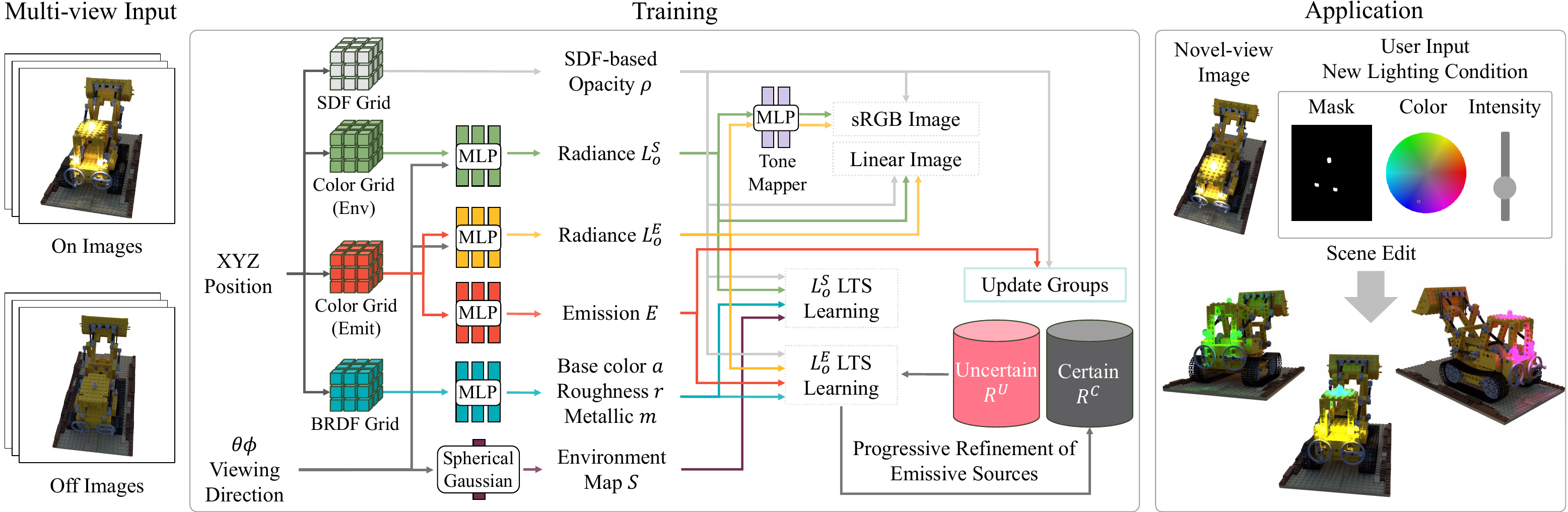}
    \caption{
        The pipeline of emissive source reconstruction. Given LDR images with emissive sources on and off, scene components are reconstructed by synthesizing training images and enforcing LTS requirements. Emissive sources are progressively refined via categorizing training rays into uncertain and certain groups. 
        The scenes can be edited with new lighting conditions using reconstructed emissive sources.
        }
    \label{fig:pipeline}
    \vspace{-10pt}
\end{figure*}

\section{Preliminaries}
\label{preliminaries}


\textbf{Surface Representation}.
Analgous to NeRF~\citep{nerf2020}, neural network $f_{\theta}$ predicts SDF values at arbitrary 3D spatial locations.
NeuS~\citep{wang2021neus} integrates surface representation into volume rendering using the SDF-based opacity $\rho(x)=\max(\frac{-\frac{d\Phi_s}{dx}\left(f(x)\right)}{\Phi_s\left(f(x)\right)},0)$.
Here $\Phi_s(x)=\left(1+e^{-sx}\right)^{-1}$ is the sigmoid function where $s$ controls the sharpness of surfaces.
The color of a ray can be calculated as
\begin{equation}
    \label{eq:volume rendering}
    \small
    \hat{C}(r) = \int_{0}^{\infty} T(r(t))\rho(r(t)) L_o(r(t),\omega_o) \,dt,
\end{equation}
where $\hat{C}(r)$ denotes the predicted ray color, $r(t;c,\omega_o)=c-t\cdot\omega_o$ is the ray with camera center $c$ along direction $\omega_o$, $T(r(t)) = \exp{(\int_{0}^{t} -\rho(r(u))\,du)}$ is the transmittance, and $L_o(r(t),\omega_o)$ is the outgoing radiance. 
Henceforth, we use $x$ to denote a point in $r(t;c,\omega_o)$ for notational simplicity. 

\vspace{5pt}
\noindent
\textbf{Light Transport in Volume Rendering}.
Extracting light sources necessitates analyzing the causes affecting the final ray colors.
Kajiya's rendering equation~\citep{kajiya} factorizes the outgoing radiance $L_o(x,\omega_o)$ into emission and reflections:
\vspace{-2pt}
\begin{equation}
    \label{eq:kajiya}
    \small
    L_o(x,\omega_o)=E(x)+\int_{\Omega}L_i(x,\omega_i) R(x,\omega_o,\omega_i;b) d\omega_i, 
    \vspace{-2pt}
\end{equation}
where $E(x)$ is the emission, $R(x,\omega_o,\omega_i;b)$ represents the SVBRDF parametrized by parameters $b$ with Lambert cosine multiplied, and $L_i(x,\omega_i)$ is the incident radiance.
In volume rendering, computing the incident radiance at point $x$ is akin to evaluating Eq.~\ref{eq:volume rendering}, with $x$ serving as the camera center.
By iteratively factorizing the outgoing radiance in the incident radiance, the contribution of a path length $i$ for a pixel can be decomposed as in Eq.~\ref{eq:light path},
where $\mathcal{H}_{i}=\Pi_{j=1}^{i-1}T(x_{j})\rho(x_{j})R(x_{j},\omega_{j-1},\omega_{j})$ is the path throughput, $S(\omega_i)$ is the environment map strength in direction $\omega_i$, and $V(x,\omega_i)=exp{\left(\int_{0}^{\infty} -\rho(r(u;x_i,-\omega_i))\,du\right)}$ is the visibility of the environment map at point $x$ along direction $\omega_i$:
\vspace{-2pt}
\begin{equation}
    \label{eq:light path}
    \small
    \begin{aligned}
        P_{i} & = \int_{l_1}\int_{\Omega}\cdots\int_{l_{i-1}}\int_{\Omega} (\int_{l_i}T(x_i)\rho(x_i) E(x_i) \,dt_i + \\
                & S(\omega_{i-1})V(x_{i-1},\omega_{i-1})) \mathcal{H}_{i}dt_{1}d\omega_{1}\cdots dt_{i-1}d\omega_{i-1}.
    \end{aligned}
\end{equation}

Extending the analysis to longer light paths, or equivalently, increasing the number of ray bounces, leads to exponential growth in computation complexity.
This poses a challenge when attempting to decompose the influence of unknown emissive sources, as their ability to produce strong reflections makes ignoring indirect illumination infeasible.

\section{Methodology}
\label{Methodology}

None of the previous works address the reconstruction of emissive sources from LDR multi-view images.
Sec. \S~\ref{sec:tonemap} through \S~\ref{sec:edit} detail our method, \modelname, which reconstructs emissive sources without prior knowledge of scene geometry, materials, or lighting specifics (including their location, number, or colors).
We also show how these reconstructed sources can be used for scene editing in \S~\ref{sec:edit}.


\subsection{Learnable Tone-mapper}
\label{sec:tonemap}

Throughout the paper, we use $\mathcal{R}$ to represent camera rays,
$C$ for pixel values, and a binary flag $\mathbb{I}$ to indicate
whether an image is captured with emissive sources on or off.



\begin{figure}
    \centering
    \tiny

    \begin{subfigure}{0.31\linewidth}
        \centering
        \includegraphics[width=\linewidth,valign=m]{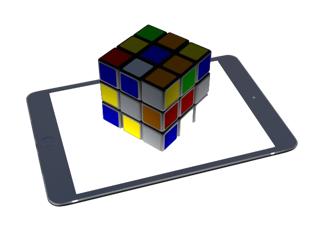}
    \end{subfigure}
    \begin{subfigure}{0.31\linewidth}
        \centering
        \includegraphics[width=\linewidth,valign=m]{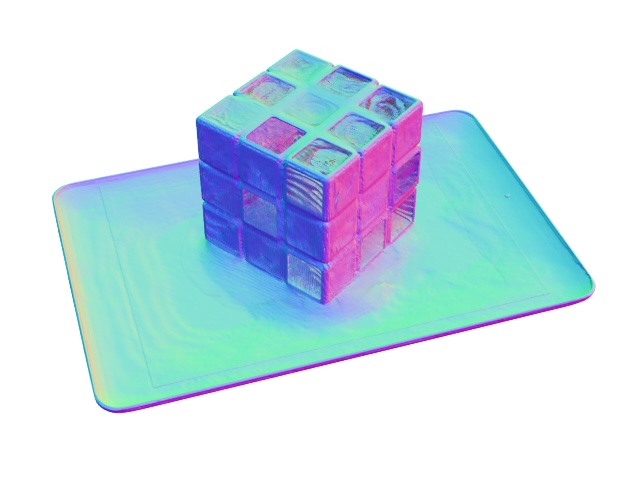}
    \end{subfigure}
    \begin{subfigure}{0.31\linewidth}
        \centering
        \includegraphics[width=\linewidth,valign=m]{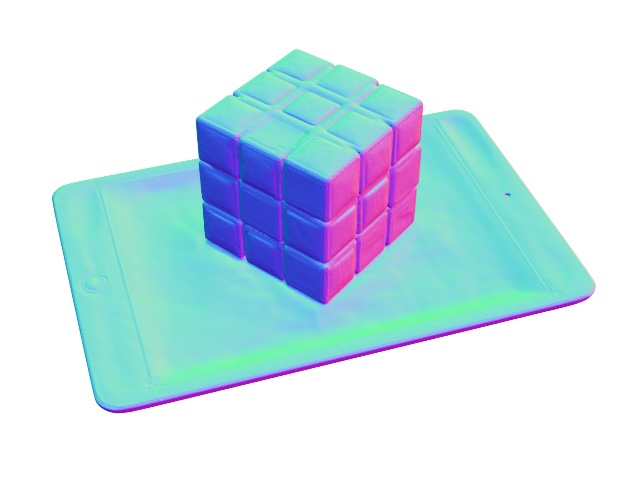}
    \end{subfigure}

    \begin{subfigure}{0.31\linewidth}
        \centering
        \includegraphics[width=\linewidth,valign=m]{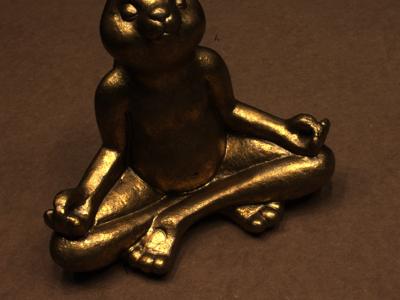}
        \captionsetup{font=footnotesize}
        \caption{Image}
    \end{subfigure}
    \begin{subfigure}{0.31\linewidth}
        \centering
        \includegraphics[width=\linewidth,valign=m]{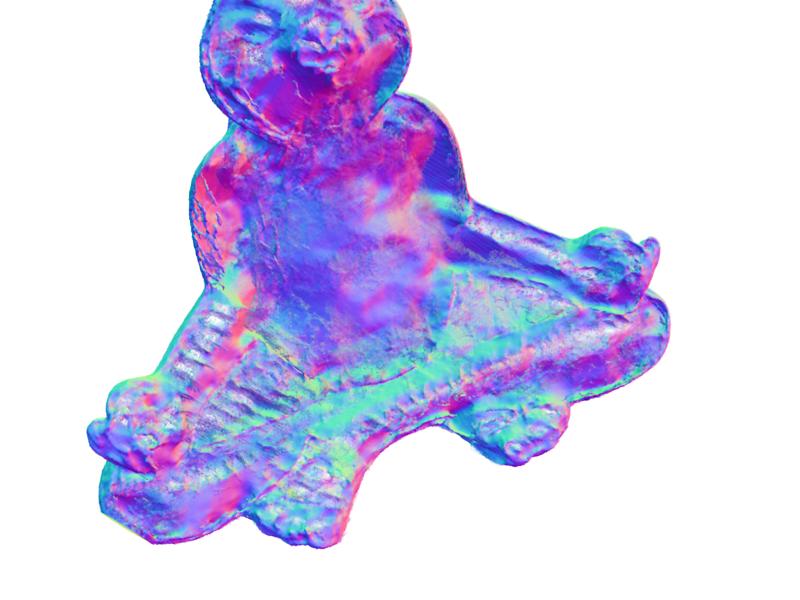}
        \captionsetup{font=footnotesize}
        \caption{w/o $m_\theta$}
    \end{subfigure}
    \begin{subfigure}{0.31\linewidth}
        \centering
        \includegraphics[width=\linewidth,valign=m]{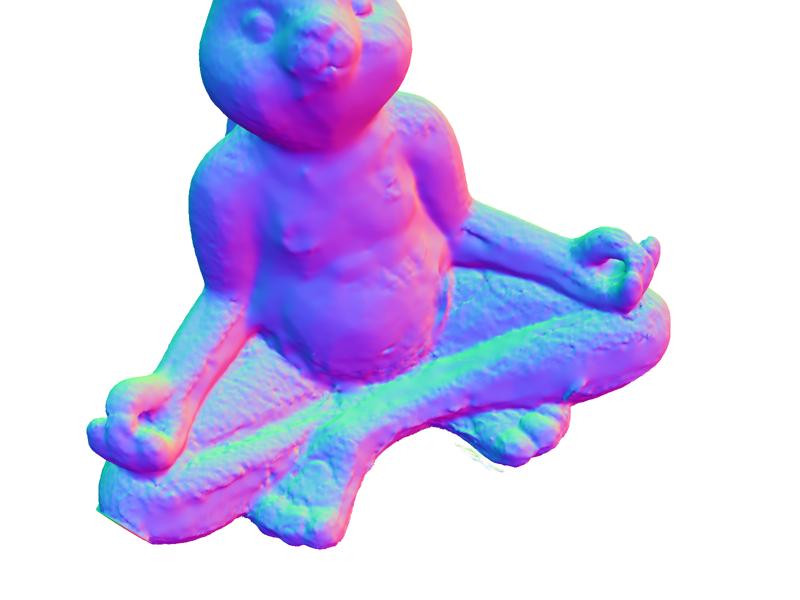}
        \captionsetup{font=footnotesize}
        \caption{\modelname}
    \end{subfigure}

    \vspace{-5pt}
    \caption{Reconstructed surfaces with the learnable tone-mapper.}
    \label{fig:surface}
    \vspace{-15pt}
\end{figure}

To extract HDR values from LDR images, we employ the softplus activation for outgoing radiance prediction and apply a clipping and gamma function $\tau$~\citep{iec61966-2-1} for the rendering loss such that $\hat{C}_{\tau}(r) = \tau(\hat{C}(r))$.
Unlike previous NeRF-based works~\cite{nerv2021,Jin2023TensoIR,mai2023neural,sun2023neuralpbir} that limit radiance to the range of $\left[0,1\right]$, our approach allows for any positive radiance values.
Yet, it creates difficulties in differentiating between the surface weight $T(x)\rho(x)$ and the magnitude of radiance value $L_o(x,\omega_o)$,
since it allows for the possibility of assigning extreme radiance to
the points with low surface weights to render same ray colors.
Such ambiguity poses challenges, particularly in dark and high-contrast scenes, aggravating surface reconstruction (see Fig.~\ref{fig:surface}).
To address this, we introduce a learnable tone-mapper $m_{\theta}: \mathbb{R}^{3}_{+} \rightarrow \left[0,1\right]^{3}$, that takes positionally encoded HDR linear values as input:
\begin{equation}
    \small
    \hat{C}_{m_\theta}(r) = \int_{0}^{\infty} T(x)\rho(x) m_\theta(L_{o}(x,\omega_o)) \,dt,
    \vspace{-2pt}
\end{equation}
\begin{equation}
    \small
    L_o(x,\omega_o) = L_o^{S}(x,\omega_o) + L_o^{E}(x,\omega_o) \cdot \mathbb{I},
\end{equation}
\noindent where $L_o^{S}(x,\omega_o)$ is radiance when emissive sources are turned off, while $L_o^{E}(x,\omega_o)$ stands for radiance added to the scene by emissive sources.
Our rendering loss is then formulated as follows, with $\lambda_\tau$ as a hyper-parameter:
\begin{equation}
    \label{eq:rendering_loss}
    \small
    \mathcal{L}_{\text{render}}=\sum_{r \in \mathcal{R}} (\lVert C(r) - \hat{C}_{m_{\theta}}(r)\rVert_2^2 + \lambda_\tau \lVert C(r) - \hat{C}_{\tau}(r)\rVert_2^2 ).
    \vspace{-2pt}
\end{equation}


\subsection{Learning of Light Transport Segments}
\label{sec:lts}
The computational complexity of object appearance analysis in volume rendering is notably high, as shown in Eq.~\ref{eq:light path}.
We take an alternative approach by leveraging neural networks to represent ray-traced fields, rather than explicitly tracing every rays.
Our distinct contribution to inverse rendering lies in precise adjustment of radiance.
Specifically, we impose constraints on the predicted radiance to satisfy each light transport segments.
The light transport segments (LTS) loss, $\mathcal{L}_{lts}$, plays a pivotal role in our method:
\begin{equation}
    \label{eq:lts_loss_env}
    \small
    \mathcal{L}_{lts}^{S} = \sum_{x,\omega_o} \lVert L_o^S(x,\omega_o) - \hat{L}_o^S(x,\omega_o) \rVert_2^2,
\end{equation}
\begin{equation}
    \label{eq:lts_loss_em}
    \small
    \mathcal{L}_{lts}^{E} = \sum_{x,\omega_o} \lVert L_o^E(x,\omega_o) - \hat{L}_o^E(x,\omega_o) \rVert_2^2,
\end{equation}
\begin{equation}
    \label{eq:lts_env}
    \small
    \begin{aligned}
         & \hat{L}_o^{S}(x,\omega_o) =  \int_{\Omega}\underbrace{ S(\omega_i) V(x, \omega_i) R(x, \omega_o, \omega_i)}_{\text{direct illumination by an environment map}} \,d\omega_i  +     \\
         & \int_{\Omega} \int_{0}^{\infty} \underbrace{T(x')\rho(x')L_o^{S}(x',-\omega_i) \,dt 'R(x, \omega_o, \omega_i)}_{\text{indirect illumination by an environment map}} \,d\omega_i.\
    \end{aligned}
\end{equation}
\begin{equation}
    \label{eq:lts_emit}
    \small
    \begin{aligned}
         & \hat{L}_o^{E}(x,\omega_o) = \underbrace{E(x)}_{\text{emission}} +                                                                                                                       \\
         & \int_{\Omega} \int_{0}^{\infty}\underbrace{T(x')\rho(x')L_o^{E}(x',-\omega_i) \,dt' R(x, \omega_o, \omega_i)}_{\text{direct \& indirect illumination by emissive sources}} \,d\omega_i.
    \end{aligned}
\end{equation}

We ensure consistency between the radiance directly predicted by the network $L_o(x,\omega_o)$ and the radiance achievable based on the scene context $\hat{L}_o(x,\omega_o)$.
Previous approaches have focused on matching $\hat{L}_o(x,\omega_o)$
to training views, overlooking the relations to $L_o(x,\omega_o)$.
This hinders the restoration of HDR radiance by supervising scene components to LDR training views.
In contrast, our LTS loss enables volumetric energy \textit{transfer} of radiance, adjusting outgoing radiance based on their interrelations.

To implement this concept, we train six dedicated networks for SDF $f(x)$, SVBRDF parameters $b(x)$, emission $E(x)$, environment map $S(\omega_i)$, outgoing radiances $L^S_o(x,\omega_o)$ and $L^E_o(x,\omega_o)$, to adhere to these LTS requirements.
For the environment map, we represent it using 48 Spherical Gaussians~\citep{SG} : $\sum_{k=1}^{M} \mu_k e^{\lambda_k(\omega_i\cdot \xi_k -1)}$, followed by the softplus activation.
$\mu\in\mathbb{R}^3$, $\lambda\in\mathbb{R}_+$, and $\xi\in\mathbb{S}^2$ respectively denote the lobe amplitude, sharpness, and axis.

\subsection{Progressive Discovery of Reflection Areas}
\label{sec:prog}
\setlength{\columnsep}{8pt}
\begin{wrapfigure}{l}{0.48\linewidth}
    \vspace{-10pt}
    \centering
    \begin{subfigure}{0.48\linewidth}
        \includegraphics[width=\linewidth]{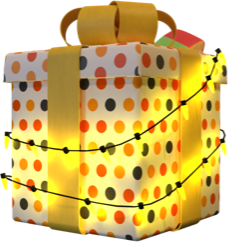}
    \end{subfigure}
    \begin{subfigure}{0.48\linewidth}
        \includegraphics[width=\linewidth]{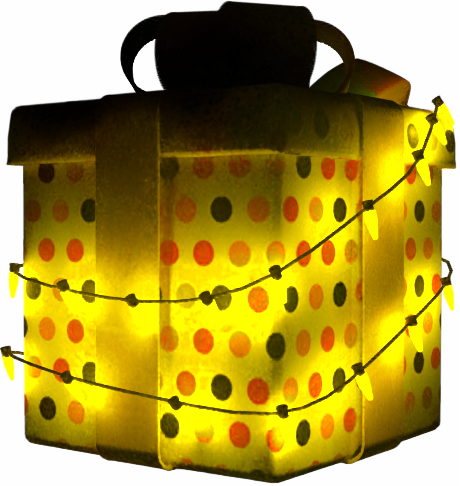}
    \end{subfigure}
    \caption{Left: Image with active emissive sources. Right: Identified emissive sources w/o progressive discovery of reflection areas.}
    \label{fig:em_every}
    \vspace{-10pt}
\end{wrapfigure}

\begin{figure}
    \centering
    \includegraphics[width=\linewidth]{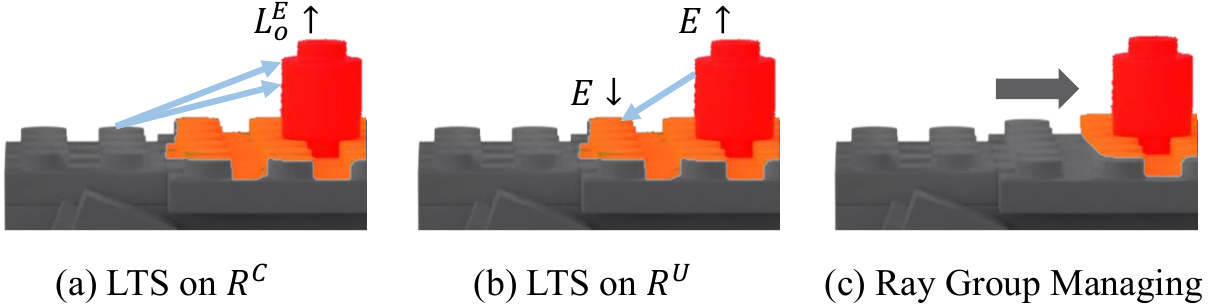}
    \vspace{-15pt}
    \caption{Illustration of the progressive emissive source reconstruction with reflection awareness. Gray color represents the areas belonging to the certain group, while the red (emissive sources) and orange (their reflections) areas belong to the uncertain group.}
    \label{fig:reflection_aware}
    \vspace{-10pt}
\end{figure}

Relying solely on LTS is insufficient for addressing ambiguity arising
from low pixel values of emissive sources and intense reflections in
adjacent regions, often leading to confusion between emission and reflection.
The right image in Fig.~\ref{fig:em_every} shows self-emitting objects restored with the naive LTS loss.
While emissive sources are small, large areas affected by them are also identified as emissive sources.
We propose a reflection-aware progressive approach for precise identification of emissive sources.
By leveraging LTS learning, we extend the regions that can be regarded as reflection areas.
Fig.~\ref{fig:reflection_aware} illustrates our progressive algorithm.

\vspace{5pt}
\noindent
\textbf{Reflection-Aware Emission Refinement}.
Since surface points are unknown and are updated during learning, we opt to utilize rays rather than surface points.
This process involves categorizing training rays into two groups: uncertain ($\mathcal{R}^U$) and certain ($\mathcal{R}^C$).
The certain group contains the rays confidently identified as reflection, aiding the transfer of radiance energy to nearby points.
For the points in the certain group, we use the Eq.~\ref{eq:lts_certain} instead of Eq.~\ref{eq:lts_emit} to exclusively attibute outgoing radiances to reflections.
Satisfying the LTS loss on the certain group results in adjusting the outgoing radiances of influential points, as illustrated in Fig.~\ref{fig:reflection_aware}(a):
\begin{equation}
    \label{eq:lts_certain}
    \small
    \hat{L}_o^{E}(x,\omega_o) = \int_{\Omega} \int_{0}^{\infty}T(x')\rho(x')L_o^{E}(x',-\omega_i) \,dt' R(x, \omega_o, \omega_i) \,d\omega_i.
\end{equation}

The uncertain group includes the rays indicating the areas that are undetermined yet as reflection or emission.
Using Eq.~\ref{eq:lts_uncertain} to compute $\hat{L}_o^E(x,\omega_o)$, this group adjusts emissions $E(x)$ based on the radiance updates by the certain group, where ``sg'' represents the stop-gradient:
\begin{equation}
    \label{eq:lts_uncertain}
    \hspace{-3pt}
    \small
    \begin{aligned}
         & \hat{L}_o^{E}(x,\omega_o) = E(x) +                                                                                                  \\
         & \text{sg}\left(\int_{\Omega} \int_{0}^{\infty}T(x')\rho(x')L_o^{E}(x',-\omega_i) \,dt' R(x, \omega_o, \omega_i) \,d\omega_i\right).
    \end{aligned}
\end{equation}
As shown in Fig.~\ref{fig:reflection_aware}(b), this leads to increased emissions for the regions whose radiances are adjusted to account for the reflections in the certain group.
Conversely, emissions decrease for the regions where there is little change in outgoing radiance, but incident radiances are increased by surrounding influential points.

\vspace{5pt}
\noindent
\textbf{Ray Group Management}.
As emissions and radiances are adjusted, the groups are dynamically updated at predefined training intervals through the following process.
Within the uncertain group, we evaluate the expected emission strength of rays, retaining only those above a threshold $k_i$.
Rays below this threshold are then merged to the certain group:
\begin{equation}
    \label{eq:group_update}
    \small
    \mathcal{R}^U_i = \{r | \max_{RGB}\left(\int_{0}^{\infty} T(x)\rho(x) E(x) \, dt\right) \geq k_i, r \in \mathcal{R}^U_{i-1}\} ,
\end{equation}
\begin{equation}
    \small
    \mathcal{R}^C_i = \left(\mathcal{R}^U_{i-1} - \mathcal{R}^{U}_{t}\right) \cup \mathcal{R}^C_{i-1} .
\end{equation}

Subsequently, newly added rays to the certain group can be used to localize influential points and update their outgoing radiances.
This iterative process progressively refines the separation between reflective and emissive regions, attaining more accurate identification of emissive sources.

\vspace{5pt}
\noindent
\textbf{LTS Loss Decomposition}.
The LTS loss, as detailed in Eq.~\ref{eq:lts_decomp}, can be decomposed using a stop-gradient operation to refine the adjustment process.
\begin{equation}
    \label{eq:lts_decomp}
    \small
    \begin{aligned}
        \mathcal{L}_{lts}^{E} = \sum_{x,\omega_o} ( & \lambda_{l} \lVert \text{sg}(L_o^E(x,\omega_o)) - \hat{L}_o^E(x,\omega_o) \rVert_1 + \\
                                                    & \lambda_{r} \lVert L_o^E(x,\omega_o) - \text{sg}(\hat{L}_o^E(x,\omega_o)) \rVert_1).
    \end{aligned}
\end{equation}

We prioritize $\lambda_l$ to enhance the update of scene context, affecting other points' radiance given the predicted $L_o(x,\omega_o)$.
$\lambda_r$ prevents severe deviation of every $L_o(x,\omega_o)$ within the current scene context.
This  aligns with our focus on HDR source reconstruction from LDR images, addressing under-represented information in training data.



\subsection{Training Details}
\label{sec:training details}
We employ the Voxurf architecture~\citep{wu2022voxurf} as backbone and adopt the simplified Disney BRDF model~\citep{DisneyBRDF} for SVBRDF representation, with parameters including base color $\in [0, 1]^3$, roughness $\in [0, 1]$, and metallic $\in [0, 1]$.
The learnable tone-mapper, structured as a two-layer MLP, is  utilized for the rendering loss only.
Initially, we pre-train our networks using the rendering loss, subsequently integrating the basic LTS loss (Eq.~\ref{eq:lts_loss_env} and Eq.~\ref{eq:lts_loss_em}) into our training regimen.
This phase transitions to the reflection-aware progressive training scheme, where we adopt the $\ell_1$ loss due to its empirical stability in refining emissive source reconstruction.
We use a smoothing regularization to promote local consistency in normals, BRDFs, and emissions.
To ensure view-consistent labeling of 3D points as either reflective or emissive, we implement the emission suppression loss for points beloning to the certain group:
\begin{equation}
    \label{eq:em_suppression}
    \small
    \mathcal{L}_{supp}^{E} =\sum_{r\in R_t^C}\lVert\int_{0}^{\infty}T(x)\rho(x)E(x) \,dt \rVert_2^2,
\end{equation}
The threshold $k_i$ linearly increases with each time step $t$, utilizing a grid search within a range of $[10^{-3}, 10^{-5}]$ to find the slope.
We construct mini-batches via stratified sampling within each group.
For a detailed description of our training procedure, please refer to  Appendix.

\begin{table*}[t]
    \vspace{-10pt}
    \centering
    \scriptsize
    \setlength{\tabcolsep}{3pt}
    \begin{tabular}{@{}c*{11}{>{\centering\arraybackslash}p{13pt}}>{\centering\arraybackslash}p{13pt}|*{12}{>{\centering\arraybackslash}p{13pt}}}
        \toprule
                   & \multicolumn{12}{c|}{White colored} & \multicolumn{12}{c}{Vivid colored}                                                                                                                                                                                                                                                                                                                                                                                                                                                                                                                                                                                                                                                                                                                                                                               \\
                   & \multicolumn{2}{c}{Lego}            & \multicolumn{2}{c}{Gift}           & \multicolumn{2}{c}{Book}        & \multicolumn{2}{c}{Cube}        & \multicolumn{2}{c}{Billboard}   & \multicolumn{2}{c|}{Balls}      & \multicolumn{2}{c}{Lego}        & \multicolumn{2}{c}{Gift}         & \multicolumn{2}{c}{Book}        & \multicolumn{2}{c}{Cube}        & \multicolumn{2}{c}{Billboard}   & \multicolumn{2}{c}{Balls}                                                                                                                                                                                                                                                                                                                                                                                                                                \\
        \cmidrule(lr){2-3} \cmidrule(lr){4-5} \cmidrule(lr){6-7} \cmidrule(lr){8-9} \cmidrule(lr){10-11} \cmidrule(lr){12-13} \cmidrule(lr){14-15} \cmidrule(lr){16-17}\cmidrule(lr){18-19}\cmidrule(lr){20-21}\cmidrule(lr){22-23}\cmidrule(lr){24-25}
                   & IoU                                 & MSE                                & IoU                             & MSE                             & IoU                             & MSE                             & IoU                             & MSE                              & IoU                             & MSE                             & IoU                             & MSE                             & IoU                             & MSE                             & IoU                             & MSE                             & IoU                             & MSE                             & IoU                             & MSE                              & IoU                             & MSE                             & IoU                             & MSE                             \\
        \midrule
        Twins      & 0.22                                & 20.19                              & \textcolor{teal}{\textbf{0.49}} & 8.59                            & 0.63                            & 3.91                            & \textcolor{teal}{\textbf{0.95}} & 31.83                            & 0.69                            & 1.12                            & 0.90                            & 0.06                            & 0.25                            & 6.96                            & \textcolor{teal}{\textbf{0.24}} & 6.09                            & 0.55                            & 2.63                            & 0.95                            & 10.64                            & 0.09                            & 0.75                            & 0.83                            & \textcolor{teal}{\textbf{0.04}} \\
        NeILF++    & 0.43                                & 20.88                              & 0.07                            & 9.38                            & \textcolor{teal}{\textbf{0.95}} & 4.64                            & 0.93                            & 32.67                            & 0.01                            & 1.95                            & \textcolor{teal}{\textbf{0.91}} & 0.80                            & 0.30                            & 7.65                            & 0.09                            & 6.86                            & \textcolor{teal}{\textbf{0.95}} & 3.36                            & 0.94                            & 11.49                            & 0.02                            & 1.57                            & 0.92                            & 0.78                            \\
        TensoIR    & \textcolor{teal}{\textbf{0.71}}     & \textcolor{teal}{\textbf{20.13}}   & 0.15                            & \textcolor{teal}{\textbf{8.55}} & \textcolor{teal}{\textbf{0.95}} & \textcolor{teal}{\textbf{3.87}} & \textcolor{teal}{\textbf{0.95}} & \textcolor{teal}{\textbf{31.73}} & \textcolor{teal}{\textbf{0.76}} & \textcolor{teal}{\textbf{1.11}} & \textcolor{blue}{\textbf{0.95}} & \textcolor{teal}{\textbf{0.05}} & \textcolor{teal}{\textbf{0.33}} & \textcolor{teal}{\textbf{6.93}} & 0.15                            & \textcolor{teal}{\textbf{6.05}} & \textcolor{teal}{\textbf{0.95}} & \textcolor{teal}{\textbf{2.59}} & \textcolor{teal}{\textbf{0.96}} & \textcolor{teal}{\textbf{10.60}} & \textcolor{teal}{\textbf{0.77}} & \textcolor{teal}{\textbf{0.74}} & \textcolor{blue}{\textbf{0.95}} & \textcolor{blue}{\textbf{0.03}} \\
        \midrule
        \modelname & \textcolor{blue}{\textbf{0.81}}     & \textcolor{blue}{\textbf{8.38}}    & \textcolor{blue}{\textbf{0.60}} & \textcolor{blue}{\textbf{3.49}} & \textcolor{blue}{\textbf{0.96}} & \textcolor{blue}{\textbf{1.19}} & \textcolor{blue}{\textbf{0.97}} & \textcolor{blue}{\textbf{17.87}} & \textcolor{blue}{\textbf{0.84}} & \textcolor{blue}{\textbf{0.46}} & \textcolor{blue}{\textbf{0.95}} & \textcolor{blue}{\textbf{0.04}} & \textcolor{blue}{\textbf{0.51}} & \textcolor{blue}{\textbf{5.48}} & \textcolor{blue}{\textbf{0.59}} & \textcolor{blue}{\textbf{2.50}} & \textcolor{blue}{\textbf{0.96}} & \textcolor{blue}{\textbf{0.51}} & \textcolor{blue}{\textbf{0.97}} & \textcolor{blue}{\textbf{7.94}}  & \textcolor{blue}{\textbf{0.88}} & \textcolor{blue}{\textbf{0.26}} & \textcolor{teal}{\textbf{0.94}} & \textcolor{blue}{\textbf{0.03}} \\
        \bottomrule
    \end{tabular}
    \caption{
        Results of emissive source identification. \modelname outperforms state-of-the-art re-lighting methods in reconstructing emissive sources, regardless of their color.
        The IoU measures the source area identification (a higher value is better), and the MSE quantifies the difference between reconstructed images and HDR ground truth images (a lower value is better).
    }
    \label{tab:performance}
    \vspace{-10pt}
\end{table*}

\subsection{Scene Editing}
\label{sec:edit}
Reconstructed emissive sources enable scene editing;
users select emissive sources using binary masks $M_{j=1\dots N}$ and specify lighting conditions using colors $c_{j=1\dots N}$ and intensities $i_{j=1\dots N}$ within the HSV color space~\cite{HSV}.

We identify the rays in the uncertain group that match $M$ by projecting expected surface points $p$ of the rays onto the camera with the pose $\mathbf{R}|\mathbf{t}$:
\begin{equation}
    \small
    p = \int_{0}^{\infty} T(x)\rho(x)x \,dt,
\end{equation}
\begin{equation}
    \small
    \mathbb{I}_j^{hit}(x)=\text{interp}(M_j,p')>0, \,\text{where } p'=\mathbf{K}[\mathbf{R}|\mathbf{t}][p|1]^T .
\end{equation}


For the rays satisfying $\mathbb{I}_j^{hit}(x)$, we apply the designated lighting conditions.
The new emission values are computed by substituting the original hue (H) and saturation (S) of $E(x)$ with the user-specified color $c_j$ and adjusting the value (V) of $v(x)$ with the new intensity $i_j$:
\begin{equation}
    \small
    \hspace{-3pt}
    E(x) = \text{hsv\_to\_rgb}\left(\left[c_{j} | \left(v(x) \times i\right)\right]\right) \cdot \mathbb{I}_{j}^{hit} + E(x) \cdot \neg \mathbb{I}_{j}^{hit} .
\end{equation}
These modifications influence scene appearance by optimizing the loss in Eq.~\ref{eq:lts_edit}.
During this process, all networks, except for $L_o^E(x,\omega_o)$, are frozen:
\begin{equation}
    \label{eq:lts_edit}
    \vspace{-5pt}
    \small
    \mathcal{L}_{edit} = \sum_{x,\omega_o} \lVert L_o^E(x,\omega_o) - \text{sg}(\hat{L}_o^E(x,\omega_o)) \rVert_2^2 .
\end{equation}

\begin{figure}[b]
    \vspace{-14pt}
    \centering
    \includegraphics[width=1\linewidth]{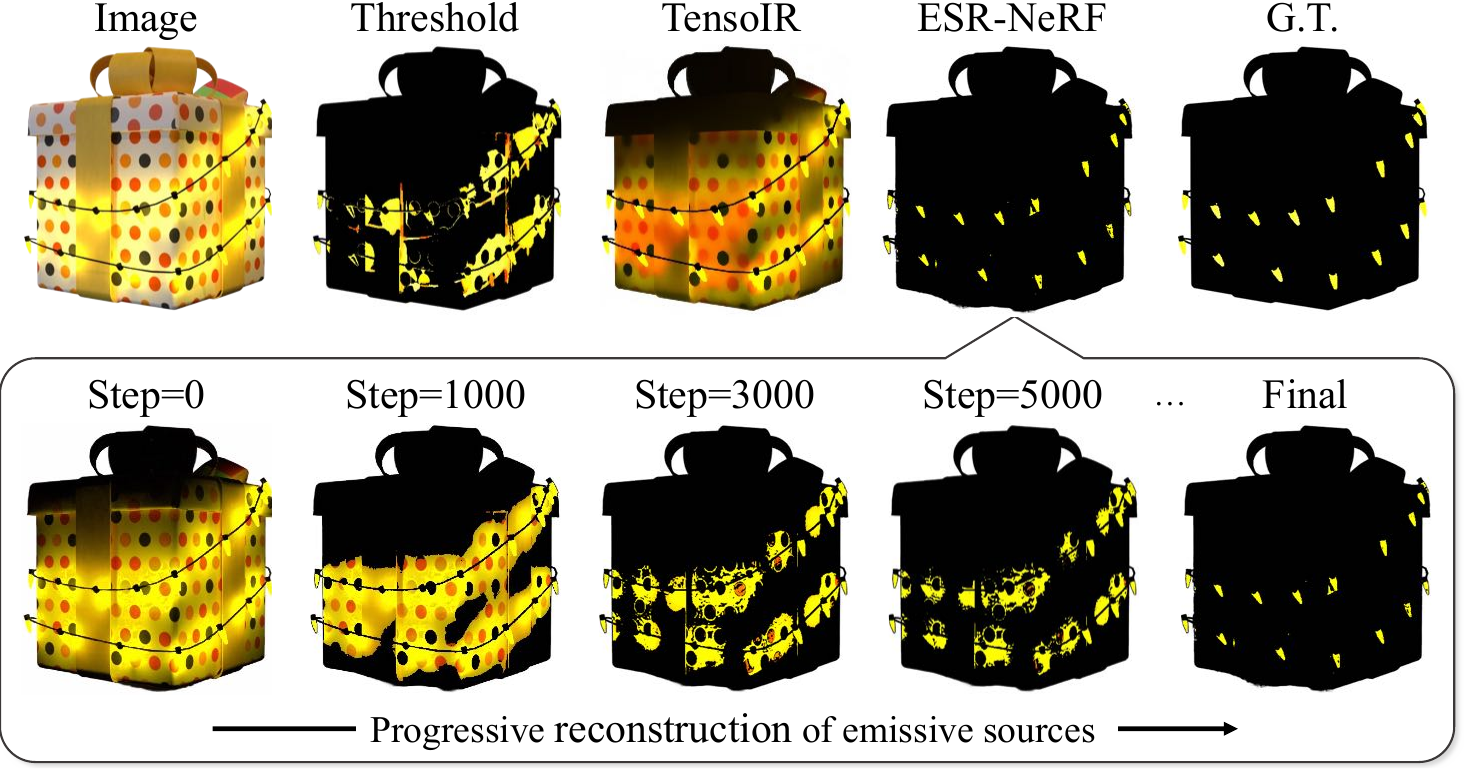}
    \caption{
        Comparison of identified emissive sources. \modelname excels through the reflection-aware progressive refinement.
    }
    \label{fig:emission recon}
    \vspace{-4pt}
\end{figure}

\begin{figure*}
    \vspace{-10pt}
    \centering
    \footnotesize
    \renewcommand{\arraystretch}{0} 
    \begin{tabular}{@{} c@{\hspace{1pt}} *{6}{c@{\hspace{3.5pt}}} @{}}
                                                  & \multicolumn{1}{c}{Image}                                                                        & \multicolumn{1}{c}{Twins}                                                                        & \multicolumn{1}{c}{PaletteNeRF}                                                                        & \multicolumn{1}{c}{TensoIR}                                                                        & \multicolumn{1}{c}{\modelname}                                                                  & \multicolumn{1}{c}{G.T.}                                                                      \\
        \rotatebox[origin=c]{90}{\makecell{Lego}} & \includegraphics[valign=m,width=.15\linewidth]{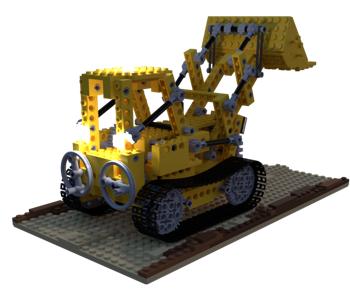} & \includegraphics[valign=m,width=.15\linewidth]{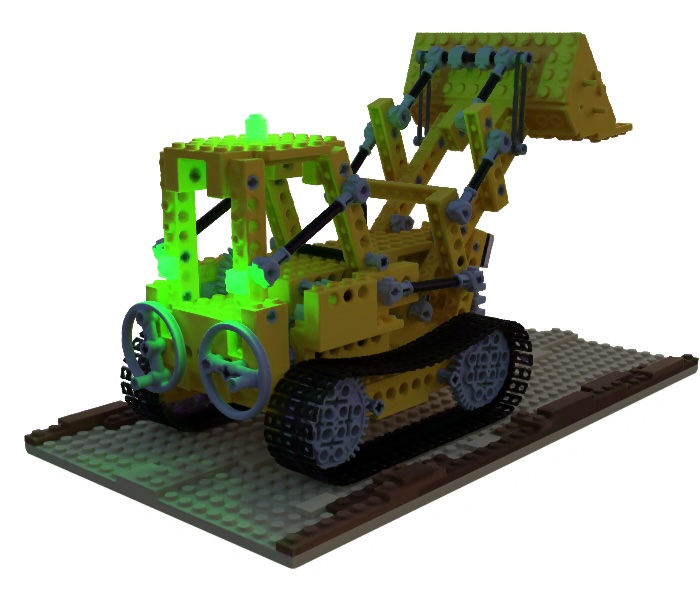} & \includegraphics[valign=m,width=.15\linewidth]{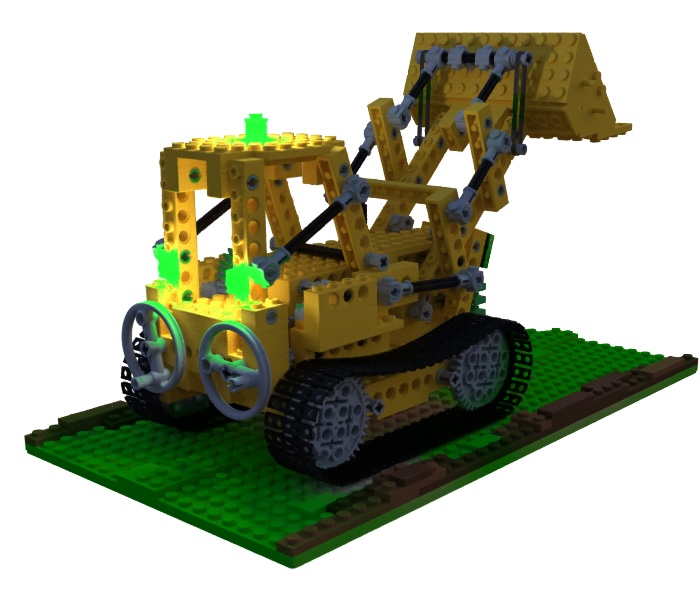} & \includegraphics[valign=m,width=.15\linewidth]{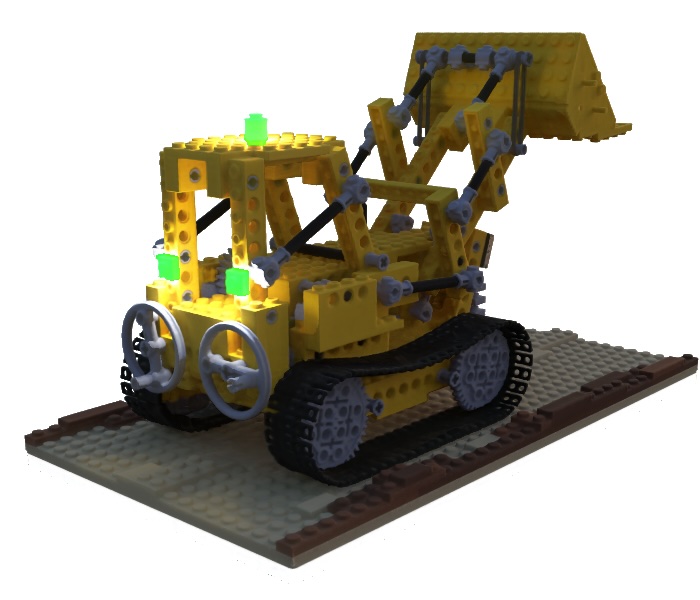} & \includegraphics[valign=m,width=.15\linewidth]{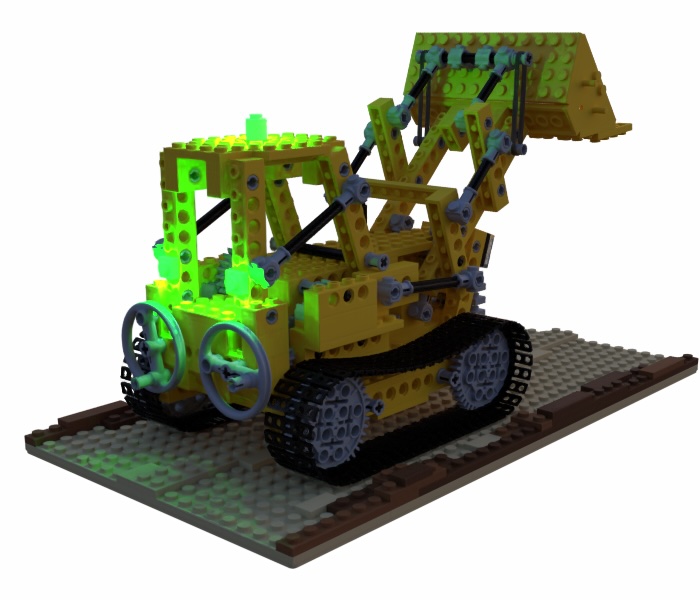} & \includegraphics[valign=m,width=.15\linewidth]{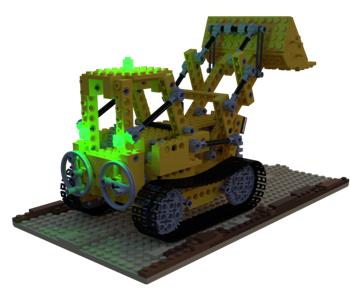} \\
        \rotatebox[origin=c]{90}{\makecell{Cube}} & \includegraphics[valign=m,width=.15\linewidth]{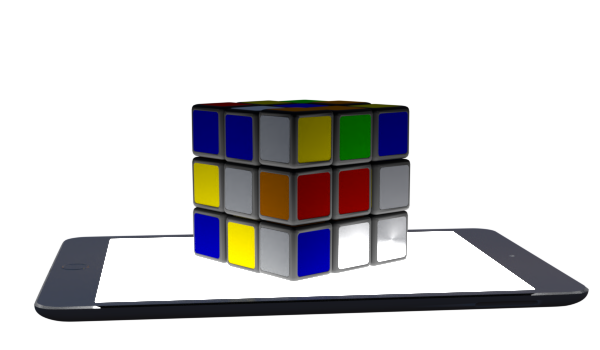}        & \includegraphics[valign=m,width=.15\linewidth]{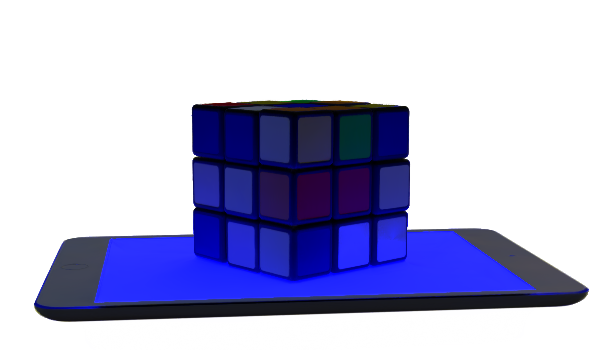}        & \includegraphics[valign=m,width=.15\linewidth]{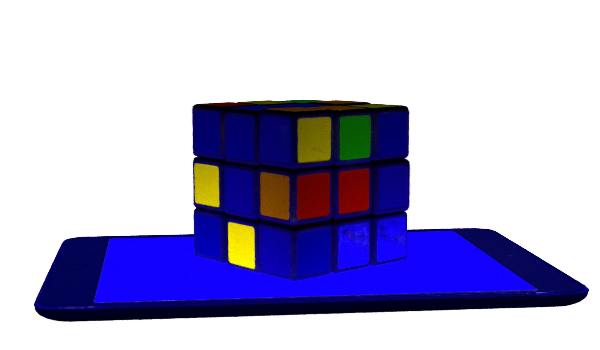}        & \includegraphics[valign=m,width=.15\linewidth]{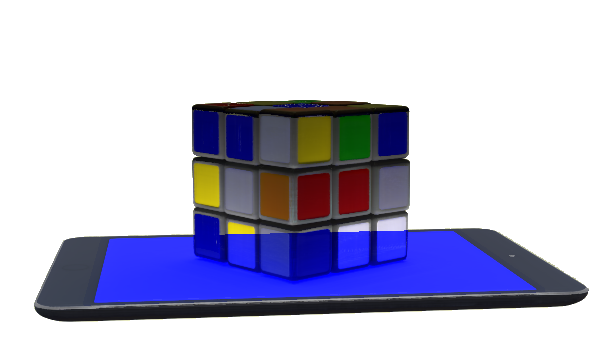}        & \includegraphics[valign=m,width=.15\linewidth]{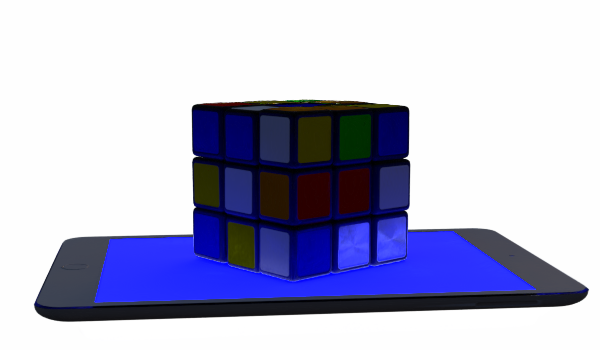}        & \includegraphics[valign=m,width=.15\linewidth]{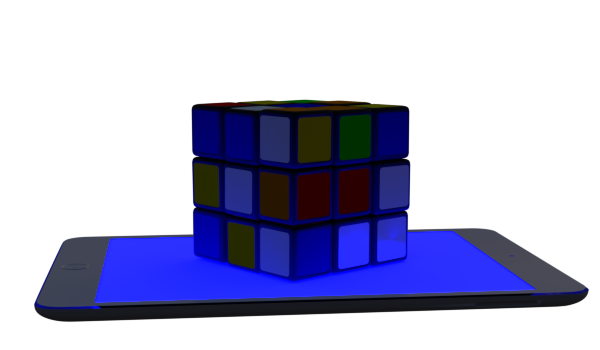}        \\
        \rotatebox[origin=c]{90}{\makecell{Gift}} & \includegraphics[valign=m,width=.15\linewidth]{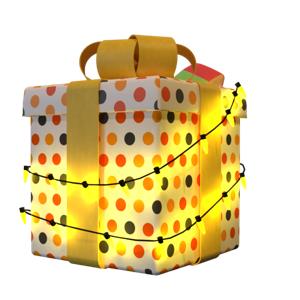}  & \includegraphics[valign=m,width=.15\linewidth]{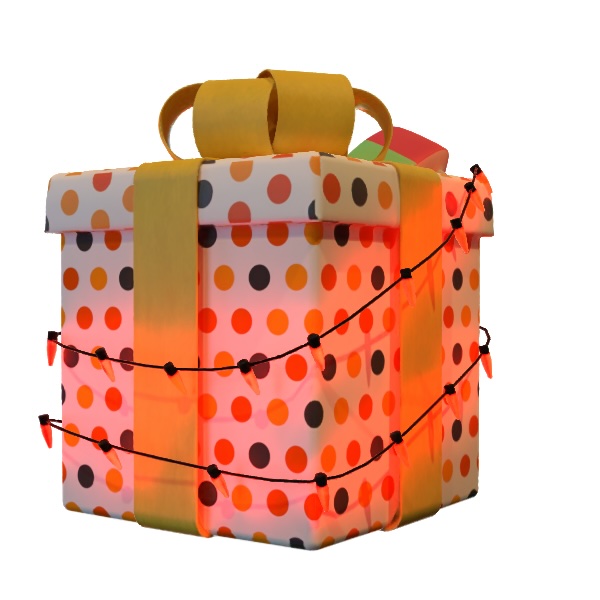}  & \includegraphics[valign=m,width=.15\linewidth]{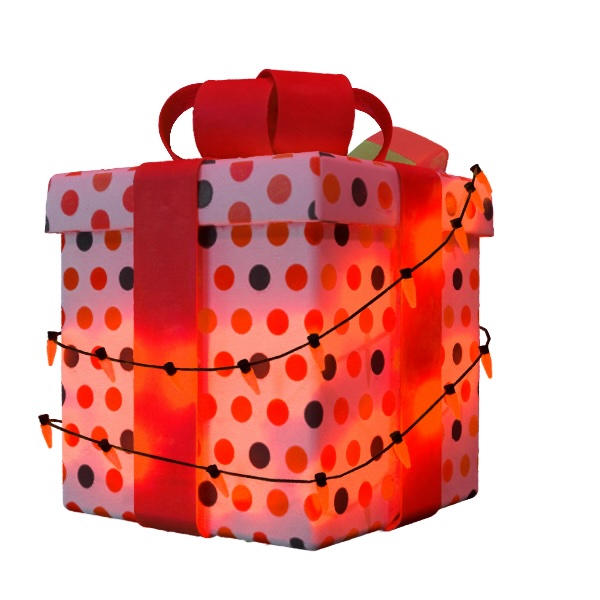}  & \includegraphics[valign=m,width=.15\linewidth]{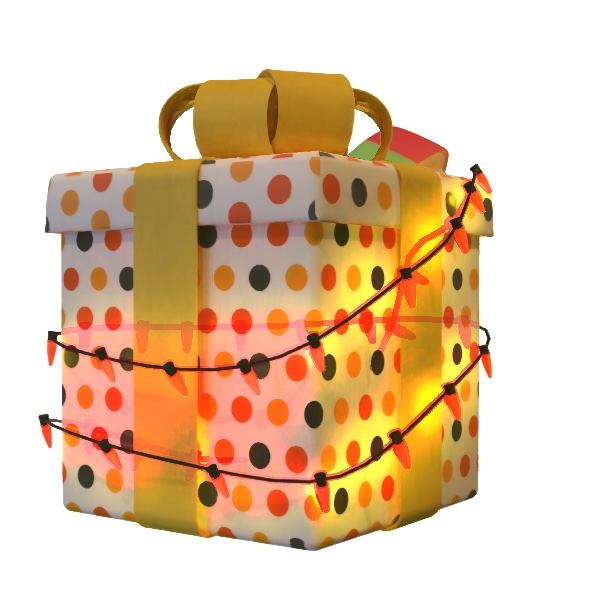}  & \includegraphics[valign=m,width=.15\linewidth]{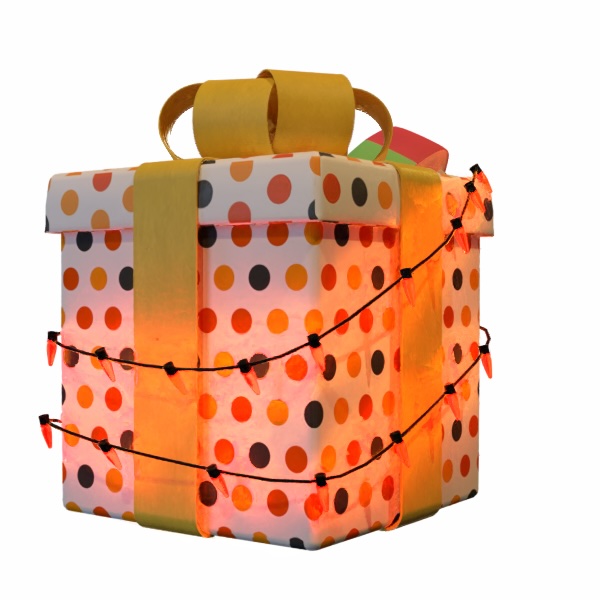}  & \includegraphics[valign=m,width=.15\linewidth]{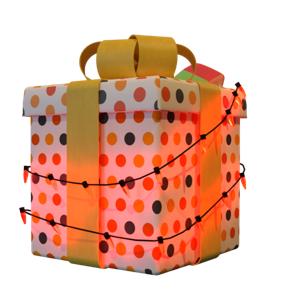}  \\
    \end{tabular}

    \hdashrule{\linewidth}{0.5pt}{1.5mm}
    \vspace{-7pt}

    \begin{tabular}{@{} *{3}{c@{\hspace{3.5pt}}} @{}}
        \includegraphics[valign=m,width=.3\linewidth]{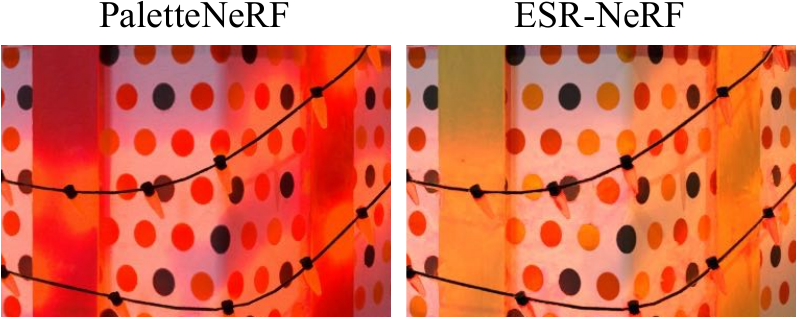} & \includegraphics[valign=m,width=.3\linewidth,height=1.2cm]{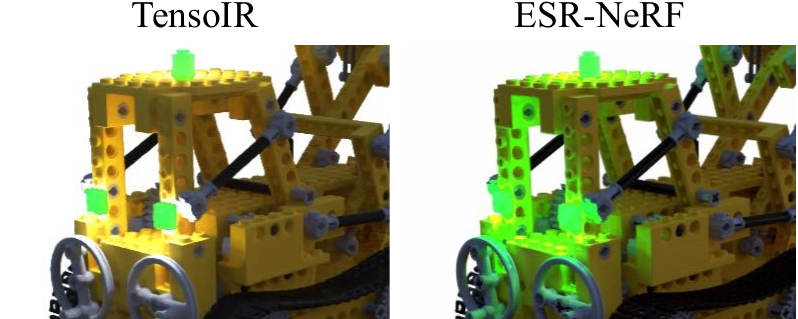} & \includegraphics[valign=m,width=.3\linewidth,height=1.2cm]{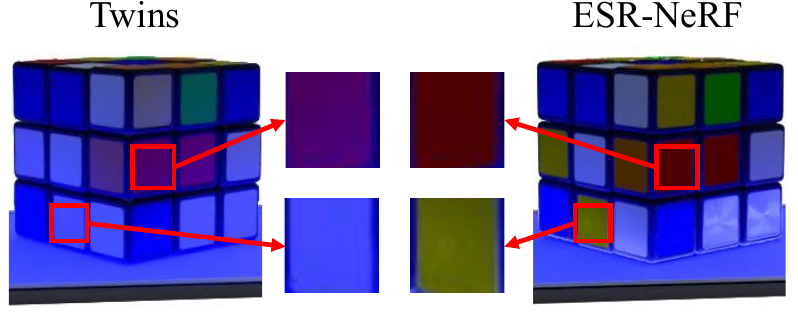} \\
    \end{tabular}

    \vspace{-5pt}
    \caption{
        Comparison of scene editing. \modelname provides precise source control and faithfully represents reflection effects.
        For easy comparison in the Cube scene of low intensity, the bottom-right images are presented with a 40\% increased brightness.
    }
    \label{fig:relight}
    \vspace{-15pt}
\end{figure*}

\section{Experiments}
\label{sec:exp}

We assess \modelname in reconstructing emissive sources by focusing on both identification and intensity restoration.
To showcase its effectiveness, we conduct a range of experiments, including scene editing, ablation studies, illumination decomposition, and surface reconstruction, providing both quantitative and qualitative results.

\subsection{Experiment Settings}

We curate 6 diverse synthetic scenes, each with 200 training images evenly distributed between on and off lighting conditions.
To evaluate the robustness of our approach against light colors, we consider two distinct settings of white colored and vivid colored emissive sources, resulting in a total of 12 scenes.  
The vivid colors are selected with full saturation in the HSV color space.
We measure source identification and radiance reconstruction using IoU and MSE metrics on novel view test images, comparing against ground truth data from Blender-rendered emission masks and EXR files.
The emission strengths, the maximum EXR file values,  range from 2 to 200.
For quantitative scene editing evaluation, we alter the white-colored sources to various colors—red, green, blue, cyan, magenta, yellow—and adjust intensities to half or double their original values.
Qualitative results include scene editing for vividly colored sources and real scenes captured with a Fuji 100s camera using Philips smart bulbs as emissive sources.
Quantitative assessments are based on 50 test images from novel camera poses, except for MSE measured for 25 test images.
We denote the best performance with blue and the second-best with green.
Additionally, we utilize the DTU dataset~\citep{DTU} to evaluate \modelname's performance in surface reconstruction tasks where emissive sources are absent.

\vspace{5pt}
\noindent
\textbf{Baselines}.
We select two state-of-the-art re-lighting methods, TensoIR~\cite{Jin2023TensoIR} and NeILF++~\cite{zhang2023neilf++}, that do not require prior lighting information.
For thorough evaluation, we also implement a simple method, Twins, where separate models are trained under light on and off conditions.
The Twins utilize the radiance discrepancies between the on and off models to distinguish and adjust emissive sources.
For scene editing, we add NeRF-W~\cite{nerfw2020} and PaletteNeRF~\cite{kuang2023palettenerf} as baselines.
Both NeRF-W and Twins adopt the Voxurf~\citep{wu2022voxurf} architecture for fair comparison.
For methods unable to individually control emissive sources, all sources are adjusted together to match the last lighting condition by a user.
For the DTU dataset, we include state-of-the-art surface reconstruction methods that use object masks, such as NeuS~\citep{wang2021neus} and Voxurf, as well as Neural-PBIR~\cite{sun2023neuralpbir}, that jointly reconstructs surfaces, materials, and environment maps.

\begin{table}[t]
    \centering
    \scriptsize
    \begin{tabular}{c*{8}{p{12pt}}}
        \toprule
                    & \multicolumn{2}{c}{NV}           & \multicolumn{2}{c}{NV + I}        & \multicolumn{2}{c}{NV+ C}        & \multicolumn{2}{c}{NV + I + C}                                                                                                                                                  \\
        \cmidrule(l){2-3} \cmidrule(l){4-5} \cmidrule(l){6-7} \cmidrule(l){8-9}
                    & PSNR                             & LPIPS                             & PSNR                             & LPIPS                             & PSNR                             & LPIPS                             & PSNR                             & LPIPS                             \\
        \midrule
        Twins       & 36.52                            & 0.0141                            & \textcolor{teal}{\textbf{27.91}} & \textcolor{teal}{\textbf{0.0252}} & \textcolor{teal}{\textbf{31.02}} & \textcolor{teal}{\textbf{0.0252}} & \textcolor{teal}{\textbf{28.21}} & \textcolor{teal}{\textbf{0.0310}} \\
        NeRF-W      & 36.44                            & 0.0142                            & 24.77                            & 0.0417                            & -                                & -                                 & -                                & -                                 \\
        NeILF++     & 24.40                            & 0.0556                            & 24.71                            & 0.0579                            & 24.06                            & 0.0750                            & 23.24                            & 0.0770                            \\
        TensoIR     & \textcolor{teal}{\textbf{38.04}} & \textcolor{teal}{\textbf{0.0103}} & 27.28                            & 0.0418                            & 26.36                            & 0.0505                            & 25.18                            & 0.0531                            \\
        PaletteNeRF & 33.66                            & 0.0233                            & 23.27                            & 0.0483                            & 24.44                            & 0.0646                            & 22.58                            & 0.0703                            \\
        \midrule
        \modelname  & \textcolor{blue}{\textbf{38.79}} & \textcolor{blue}{\textbf{0.0083}} & \textcolor{blue}{\textbf{29.99}} & \textcolor{blue}{\textbf{0.0193}} & \textcolor{blue}{\textbf{31.73}} & \textcolor{blue}{\textbf{0.0196}} & \textcolor{blue}{\textbf{31.63}} & \textcolor{blue}{\textbf{0.0199}} \\
        \bottomrule
    \end{tabular}
    \caption{
        Scene editing results.
        NV: novel view synthesis, I: intensity editing, and C: color editing.
        A higher PSNR or lower LPIPS value is better.
    }
    \label{tab:relight}
    \vspace{-16pt}
\end{table}

\subsection{Results}

\textbf{Emissive Source Recosntruction}.
Tab.~\ref{tab:performance} shows that our approach excels in accurately identifying emissive source regions and restoring their intensity, regardless of the source color.
While TensoIR and NeILF++ can restore emissions by modifying their physical rendering equations, they suffer from emissive source ambiguity, leading to near-zero IoU performance (see Appendix).
For a comprehensive comparison, we report the best performance of the baseline methods using thresholding on the reconstructed emission strength at 0.01 intervals.
\modelname consistently outperforms the baselines in identifying emissive source regions across all scenes.
Our method also achieves significantly lower MSE values for restoring LDR to HDR images compared to the baselines, demonstrating its effectiveness of handling the ill-posed nature of the scenes with emissive sources.
This is visually confirmed in Fig.~\ref{fig:emission recon}, where \modelname surpasses the baselines in a complex scene with numerous small light bulbs.

\begin{figure}[b]
    \vspace{-17pt}
    \centering
    \footnotesize
    \renewcommand{\arraystretch}{1.2} 
    \begin{tabular}{@{} *{4}{c@{\hspace{3pt}}} @{}}
        \multicolumn{1}{c}{Image}                                                                   & \multicolumn{1}{c}{Emission}                                                            & \multicolumn{1}{c}{Re-light (ours)}                                                       & \multicolumn{1}{c}{Re-light (G.T.)}                                                     \\
        \includegraphics[width=.24\linewidth]{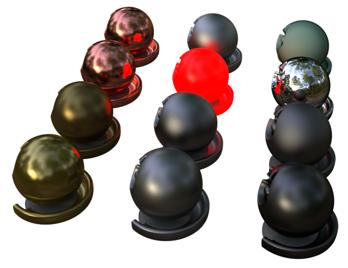} & \includegraphics[width=.24\linewidth]{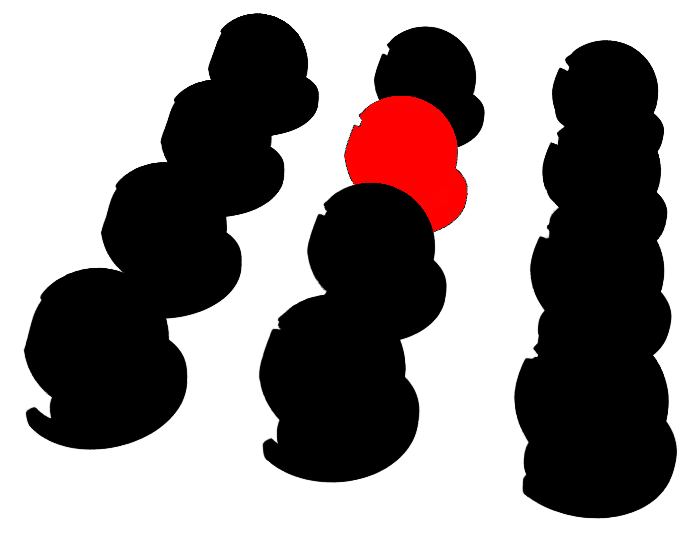} & \includegraphics[width=.24\linewidth]{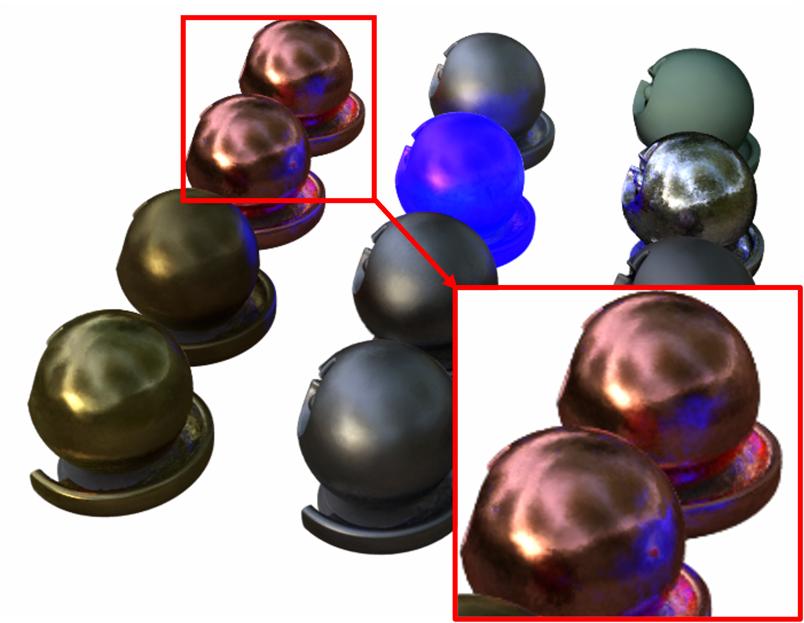} & \includegraphics[width=.24\linewidth]{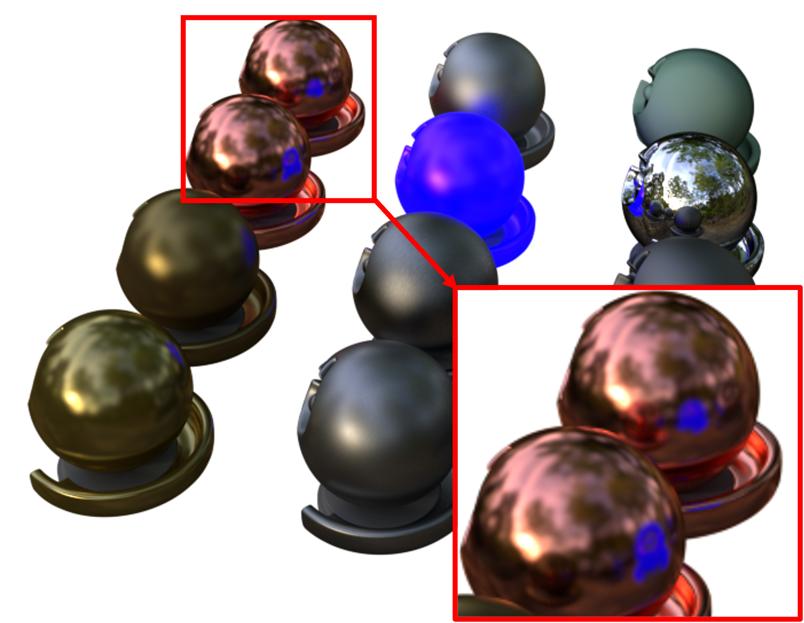} \\
    \end{tabular}
    \vspace{-7pt}
    \caption{
        Reconstructed emitter and re-lighting at novel view.
    }
    \label{fig:material}
    \vspace{-6pt}
\end{figure}

\vspace{5pt}
\noindent
\textbf{Scene Editing}.
\begin{figure}[b]
    \vspace{-22pt}
    \centering
    \footnotesize
    \renewcommand{\arraystretch}{1.5} 
    \begin{tabular}{@{} c@{\hspace{5pt}} *{4}{c@{\hspace{2pt}}} @{}}
                                           & \multicolumn{1}{c}{Lego}                                                                   & \multicolumn{1}{c}{Gift}                                                                     & \multicolumn{1}{c}{Book}                                                                    & \multicolumn{1}{c}{Billboard}                                                                  \\
        \rotatebox[origin=c]{90}{Image}    & \includegraphics[valign=m,width=.11\textwidth]{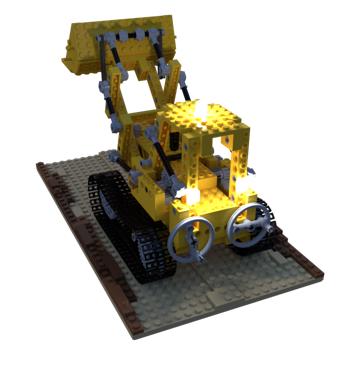}   & \includegraphics[valign=m,width=.11\textwidth]{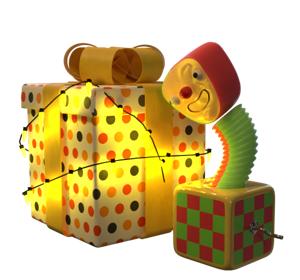}   & \includegraphics[valign=m,width=.11\textwidth]{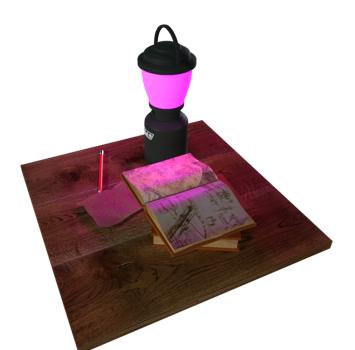} & \includegraphics[valign=m,width=.11\textwidth]{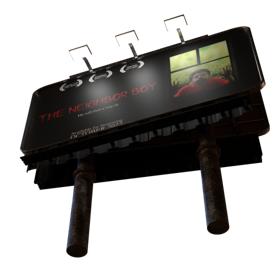}   \\
        \rotatebox[origin=c]{90}{Emission} & \includegraphics[valign=m,width=.11\textwidth]{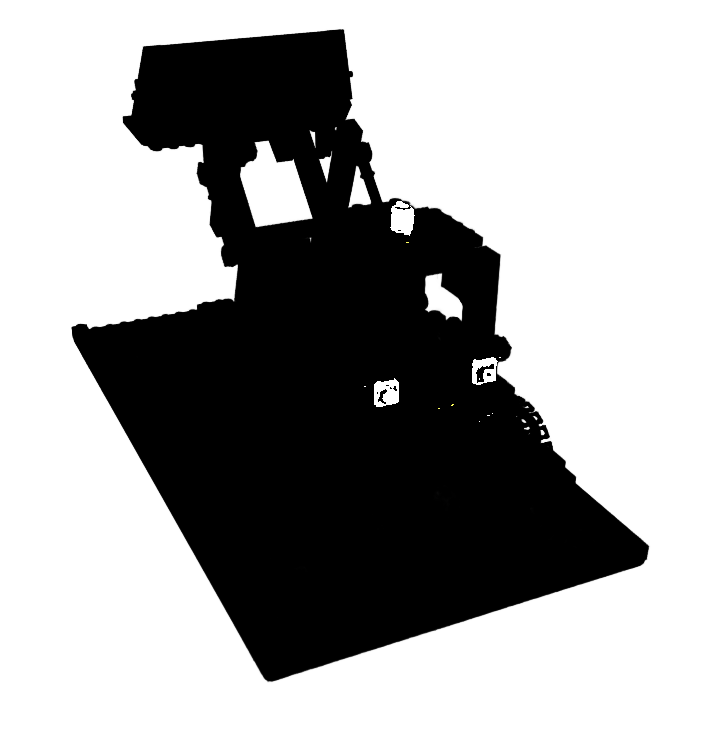}    & \includegraphics[valign=m,width=.11\textwidth]{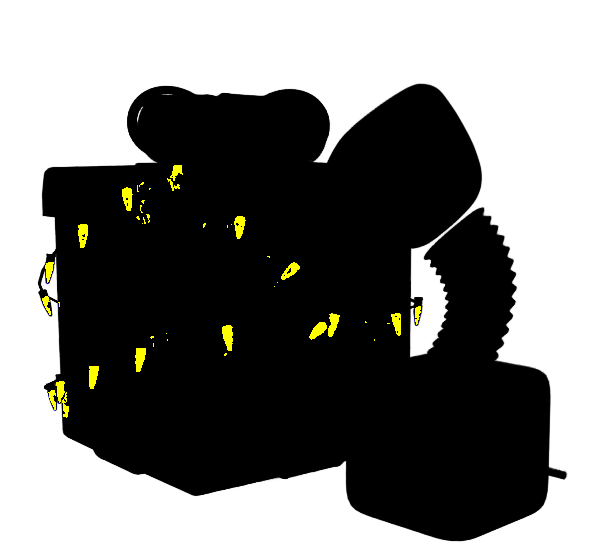}    & \includegraphics[valign=m,width=.11\textwidth]{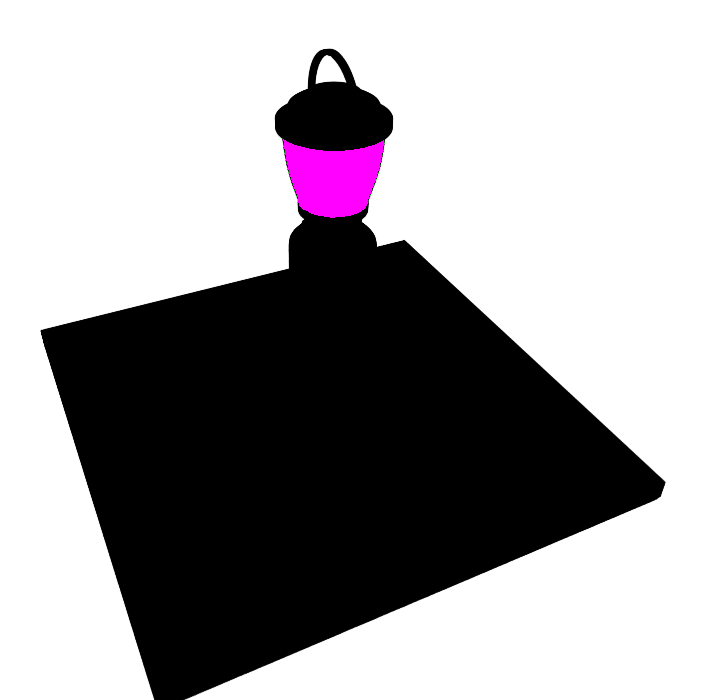}     & \includegraphics[valign=m,width=.11\textwidth]{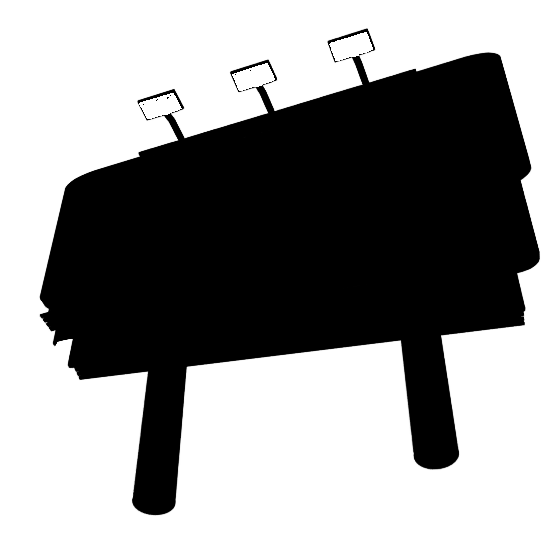}    \\
        \rotatebox[origin=c]{90}{Edited}   & \includegraphics[valign=m,width=.11\textwidth]{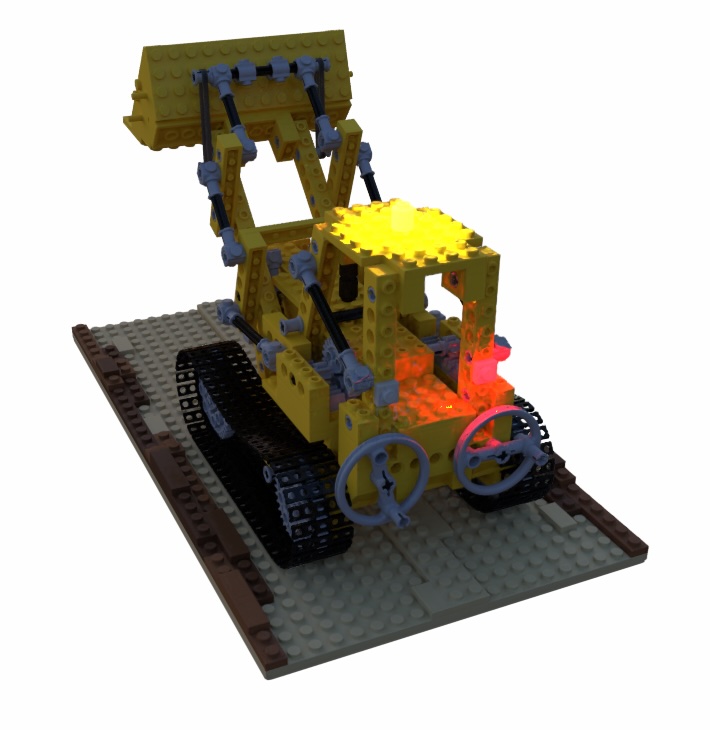} & \includegraphics[valign=m,width=.11\textwidth]{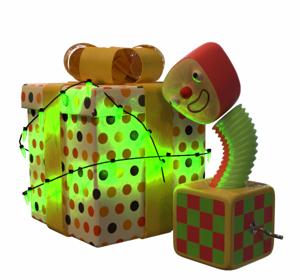} & \includegraphics[valign=m,width=.11\textwidth]{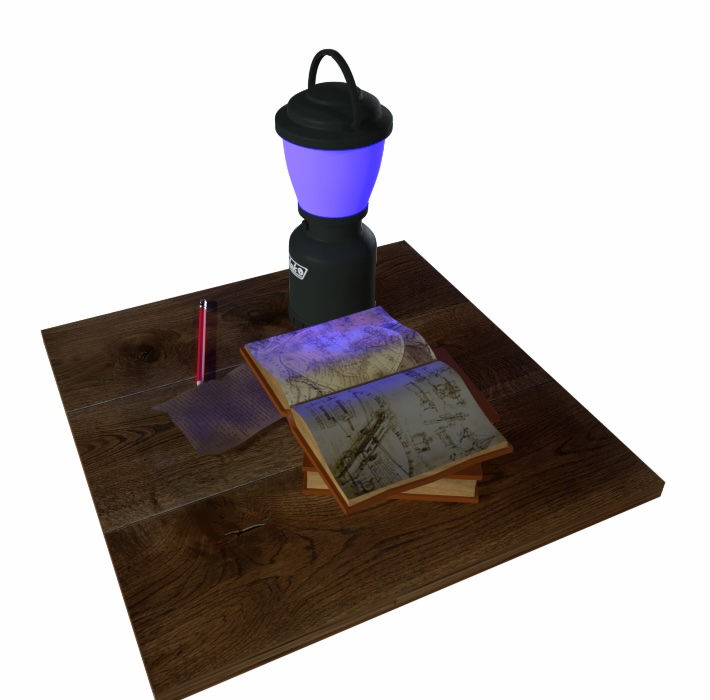}  & \includegraphics[valign=m,width=.11\textwidth]{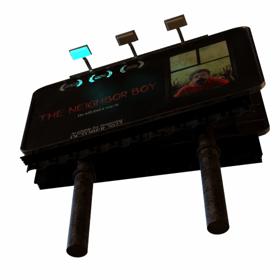} \\
    \end{tabular}
    \vspace{-7pt}
    \caption{Results of source reconstruction and scene editing.}
    \label{fig:relight2}
    \vspace{-6pt}
\end{figure}
Tab.~\ref{tab:relight} and Fig.~\ref{fig:relight} showcase the scene editing results under novel lighting conditions.
Baseline methods struggle to adapt to lighting changes due to their inability to reconstruct emissive sources accurately.
For example, in the Lego scene, TensoIR fails to adjust the illumination in surrounding regions when the color of emissive sources is changed, and in the Cube scene, both the hidden iPad screen and the cube surface covered by the user input mask change together.
Twins introduces blue light onto yellow and red surfaces, leading to unintended white and purple appearances, even though there should be no reflection.
PaletteNeRF, which manipulates scenes through re-colorization, lacks precise control over illumination, as seen in the synchronous color changes in the yellow ribbon and lighting.
In contrast, \modelname demonstrates superior performance in scene editing outshining all baselines thanks to the accurate identification of emissive sources, as detailed in Table~\ref{tab:relight}.
\modelname effectively balances source reconstruction and novel view synthesis, ensuring high performance in both tasks.
NeRF-W is excluded from color adjustments since it doesn't support direct color change through interpolating latent variables learned with light on and off conditions.

Fig.~\ref{fig:material} to~\ref{fig:relight2} present additional examples of emissive source reconstruction and scene editing results.
Fig.~\ref{fig:relight3} shows results on real scenes, for which
due to the impracticality of precise control over smart bulb colors, we offer emission reconstruction results with pseudo ground truth data.
Our method effectively identifies emissive sources in real scenes, while it faces challenges in capturing complex reflections within light bulbs, as evident in the bright spot at the center of the bulbs in the ground truth edit results.

\vspace{3pt}
\noindent
\textbf{Ablation Analysis}.
Progressive refinement with the stop-gradient operation in Eq.~\ref{eq:lts_decomp}  improves the identification of emissive sources and reduces MSE values.
Without $m_\theta$, surface reconstructions become unreliable, complicating the accurate reconstruction of emissive sources.
This issue is evident from the CD metrics and illustrated in Fig.~\ref{fig:surface}.
Further analyses are provided in  Appendix.

\begin{figure}[b]
    \vspace{-13pt}
    \centering
    \includegraphics[width=\linewidth]{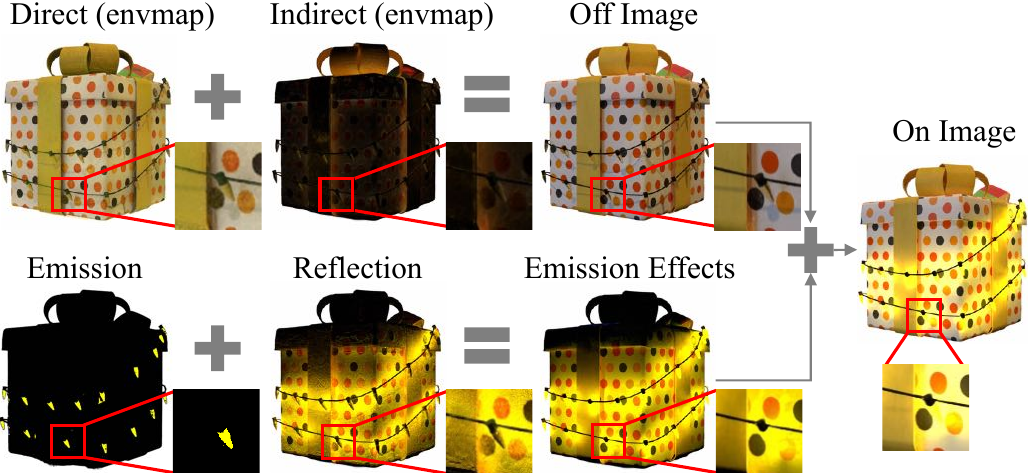}
    \vspace{-15pt}
    \caption{An example of illumination decomposition.}
    \label{fig:illum decomp}
    \vspace{-4pt}
\end{figure}

\vspace{3pt}
\noindent
\textbf{Illumination Decomposition}.
Fig.~\ref{fig:illum decomp} demonstrates \modelname's  decomposition of scene illumination into direct and indirect lighting from an environment map, as well as emissions and their reflections.
The shadow behind the yellow ribbon in the direct figure and the illumination in the indirect figure showcase \modelname's ability to model both direct and indirect illumination.
The reflection figure shows that our method accurately captures how emissive sources contribute to reflections on nearby regions.

\begin{table}
    \vspace{-8pt}
    \footnotesize
    \renewcommand{\arraystretch}{1.5} 
    \begin{tabular}{@{} c@{\hspace{5pt}} *{4}{c@{\hspace{4pt}}} @{}}
                                            & \multicolumn{1}{c}{Image}                                                                                       & \multicolumn{1}{c}{Emission}                                                                          & \multicolumn{1}{c}{Edited}                                                                               & \multicolumn{1}{c}{Pseudo G.T.}                                                                                         \\
        \rotatebox[origin=c]{90}{Jobs}      & \includegraphics[valign=m,width=.225\linewidth]{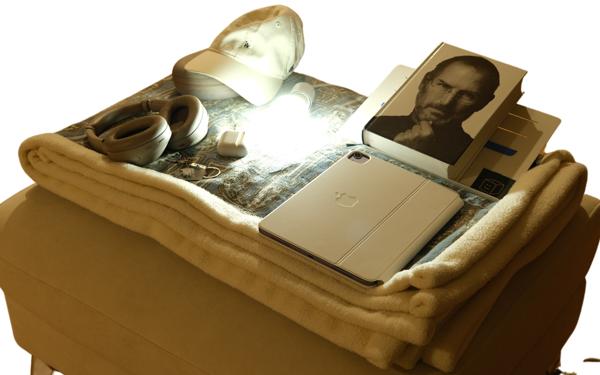}               & \includegraphics[valign=m,width=.225\linewidth]{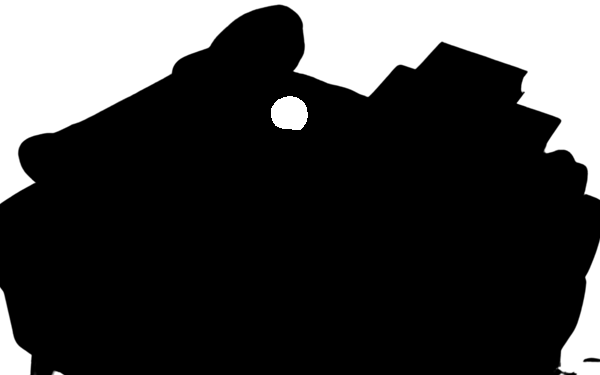}          & \includegraphics[valign=m,width=.225\linewidth]{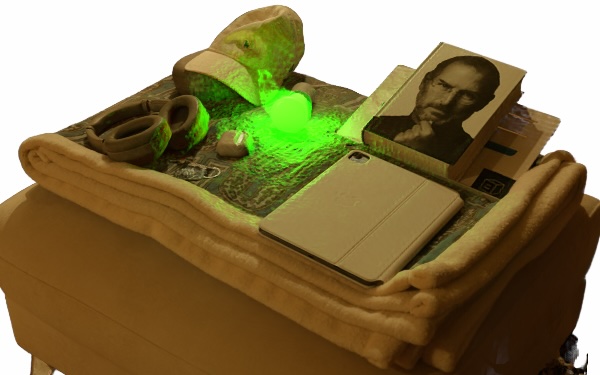}             & \includegraphics[valign=m,width=.225\linewidth]{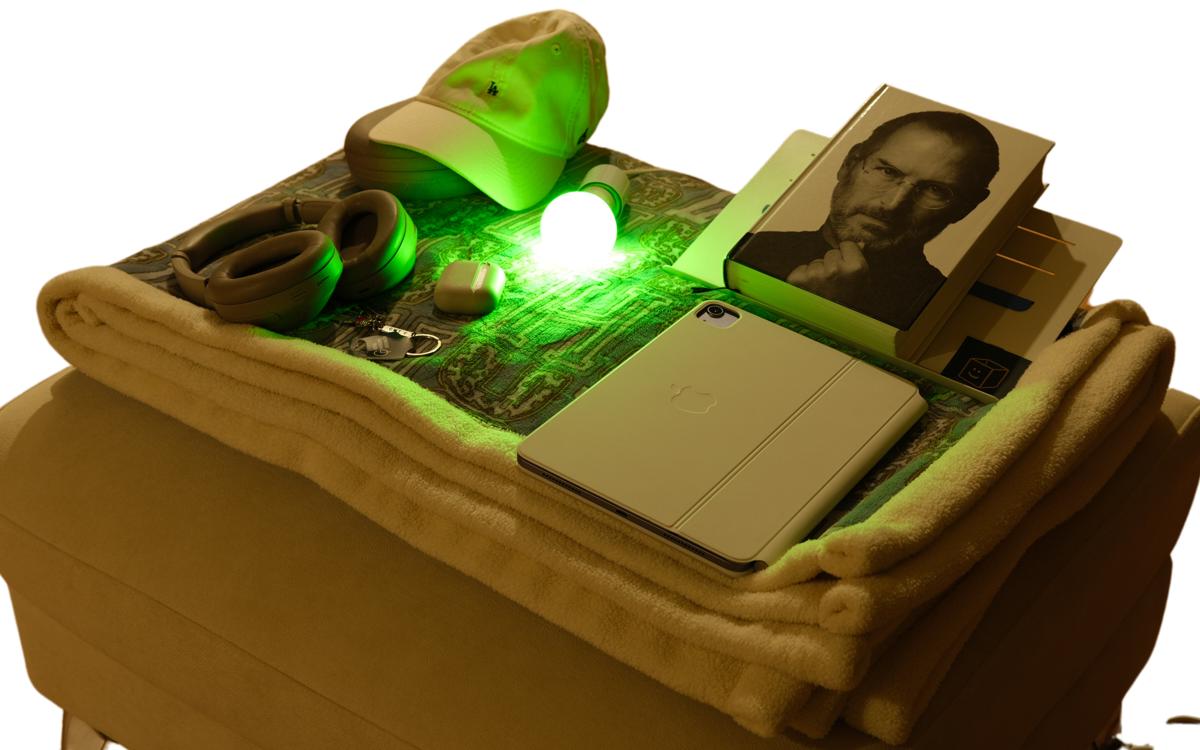}                           \\
        \rotatebox[origin=c]{90}{Dolls}     & \includegraphics[valign=m,width=.225\linewidth]{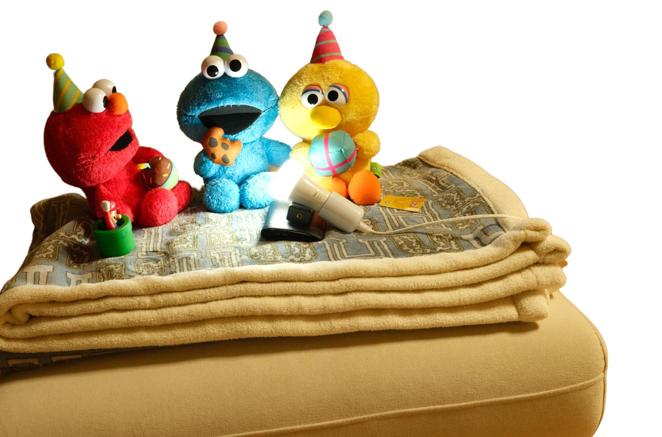}     & \includegraphics[valign=m,width=.225\linewidth]{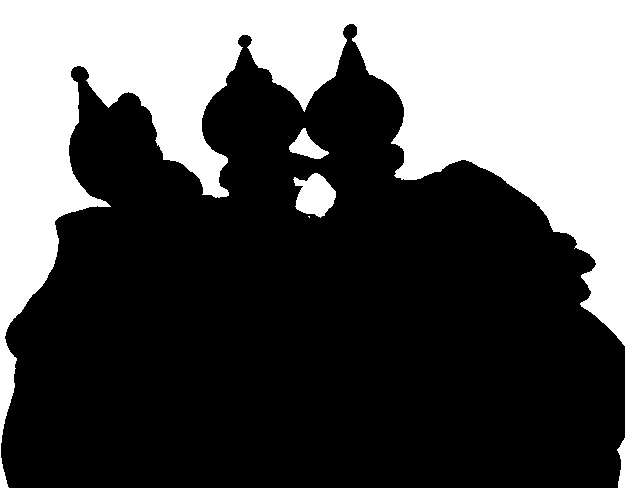}         & \includegraphics[valign=m,width=.225\linewidth]{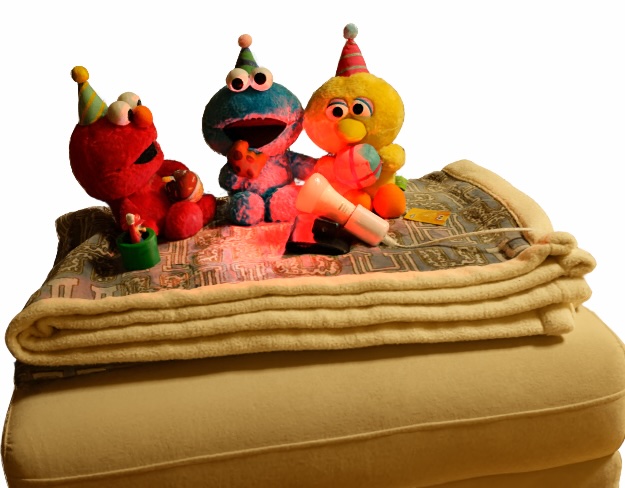}     & \includegraphics[valign=m,width=.225\linewidth]{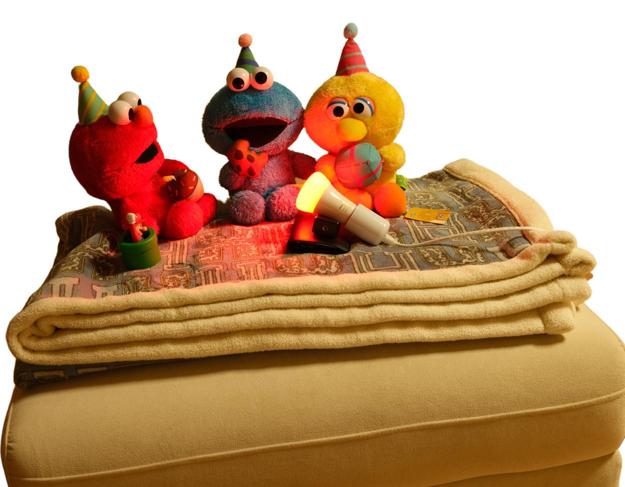}                 \\
        \rotatebox[origin=c]{90}{Cosmetics} & \includegraphics[valign=m,width=.225\linewidth]{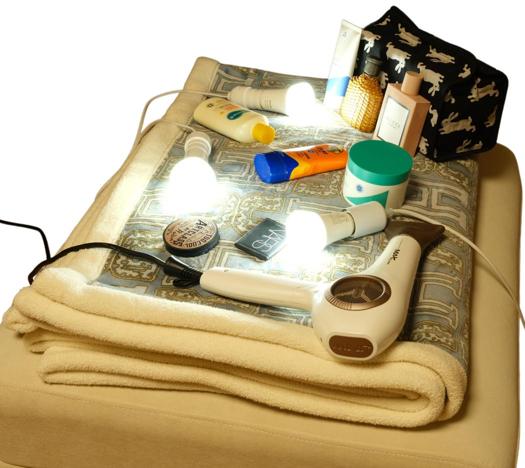} & \includegraphics[valign=m,width=.225\linewidth]{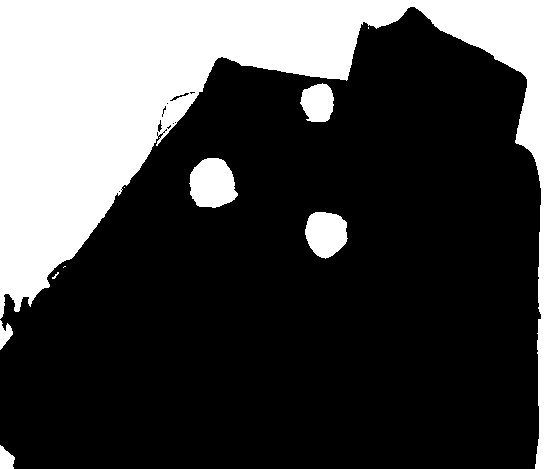} & \includegraphics[valign=m,width=.225\linewidth]{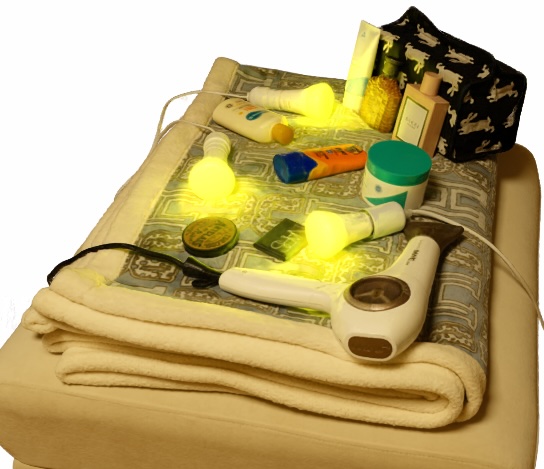} & \includegraphics[valign=m,width=.225\linewidth]{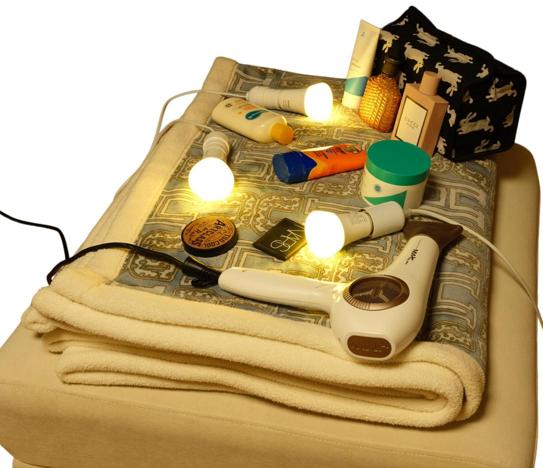} \\
    \end{tabular}
    \vspace{-2pt}
    \captionof{figure}{Source reconstruction and scene editing on real scenes.}
    \label{fig:relight3}
    \vspace{-17pt}
\end{table}

\vspace{3pt}
\noindent
\textbf{Surface Reconstruction}.
Interestingly, our approach can be applied to the scenes without emissive sources to enhance surface reconstruction, as evidenced by the lower CD values in Tab.~\ref{tab:DTU CD} on the DTU dataset.
For this experiment, we use Eq.~\ref{eq:lts_loss_env} to \ref{eq:lts_emit} without our progressive refinement technique.
\modelname's ability to adjust interrelated outgoing radiances helps  prevent surface formations where radiances cannot be produced, considering the predicted scene context.
Additional visualizations of the normals, BRDF, and environment maps are provided in  Appendix.

\begin{table}
    \vspace{-10pt}
    \centering
    \scriptsize
    \setlength{\tabcolsep}{4pt}
    \begin{tabular}{c|*{3}{c}|c}
        \toprule
        Scan & NeuS                            & Voxurf                          & Neural-PBIR                     & \modelname                      \\
        \midrule
        24   & 0.83                            & 0.65                            & \textcolor{blue}{\textbf{0.57}} & \textcolor{teal}{\textbf{0.58}} \\
        37   & 0.98                            & \textcolor{teal}{\textbf{0.74}} & 0.75                            & \textcolor{blue}{\textbf{0.71}} \\
        40   & 0.56                            & \textcolor{teal}{\textbf{0.39}} & \textcolor{blue}{\textbf{0.38}} & \textcolor{blue}{\textbf{0.38}} \\
        55   & 0.37                            & \textcolor{teal}{\textbf{0.35}} & 0.36                            & \textcolor{blue}{\textbf{0.33}} \\
        63   & 1.13                            & \textcolor{teal}{\textbf{0.96}} & 1.04                            & \textcolor{blue}{\textbf{0.93}} \\
        65   & \textcolor{teal}{\textbf{0.59}} & 0.64                            & 0.73                            & \textcolor{blue}{\textbf{0.57}} \\
        69   & \textcolor{blue}{\textbf{0.60}} & 0.85                            & \textcolor{teal}{\textbf{0.65}} & 0.78                            \\
        83   & 1.45                            & 1.58                            & \textcolor{teal}{\textbf{1.28}} & \textcolor{blue}{\textbf{1.18}} \\
        97   & \textcolor{blue}{\textbf{0.95}} & 1.01                            & \textcolor{teal}{\textbf{0.97}} & \textcolor{blue}{\textbf{0.95}} \\
        105  & 0.78                            & \textcolor{teal}{\textbf{0.68}} & 0.76                            & \textcolor{blue}{\textbf{0.58}} \\
        106  & \textcolor{blue}{\textbf{0.52}} & 0.60                            & \textcolor{teal}{\textbf{0.53}} & 0.54                            \\
        110  & 1.43                            & 1.11                            & \textcolor{blue}{\textbf{0.84}} & \textcolor{teal}{\textbf{1.08}} \\
        114  & \textcolor{teal}{\textbf{0.36}} & 0.37                            & 0.38                            & \textcolor{blue}{\textbf{0.33}} \\
        118  & \textcolor{teal}{\textbf{0.45}} & \textcolor{teal}{\textbf{0.45}} & 0.46                            & \textcolor{blue}{\textbf{0.40}} \\
        122  & \textcolor{teal}{\textbf{0.45}} & 0.47                            & 0.49                            & \textcolor{blue}{\textbf{0.44}} \\
        \midrule
        mean & 0.77                            & 0.72                            & \textcolor{teal}{\textbf{0.68}} & \textcolor{blue}{\textbf{0.65}} \\
        \bottomrule
    \end{tabular}
    \vspace{-4pt}
    \caption{
        Results of surface reconstruction via the Chamfer distance on the DTU dataset. A lower value is better.
    }
    \label{tab:DTU CD}
    \vspace{-5pt}
\end{table}

\begin{table}
    \centering
    \scriptsize
    \setlength{\tabcolsep}{4pt}

    \begin{minipage}{.69\linewidth}
        \centering
        \begin{tabular}{c|cc|cc}
            \toprule
                            & \multicolumn{2}{c|}{White} & \multicolumn{2}{c}{Vivid}                                     \\
                            & IoU $\uparrow$             & MSE $\downarrow$          & IoU $\uparrow$ & MSE $\downarrow$ \\
            \midrule
            w/o progressive & 0.40                       & 9.92                      & 0.41           & 3.93             \\
            w/o sg          & 0.71                       & 6.45                      & 0.60           & 3.47             \\
            \midrule
            \modelname      & 0.86                       & 5.24                      & 0.81           & 2.79             \\
            \bottomrule
        \end{tabular}
    \end{minipage}
    \hfill
    \begin{minipage}{.3\linewidth}
        \centering
        \begin{tabular}{c|c}
            \toprule
                           & DTU             \\
                           & CD $\downarrow$ \\
            \midrule
            w/o $m_\theta$ & 0.93            \\
            w/o LTS        & 0.71            \\
            \midrule
            \modelname     & 0.65            \\
            \bottomrule
        \end{tabular}
    \end{minipage}

    \label{tab:ablation}
    \vspace{-4pt}
    \caption{Ablation studies on the sruface reconstruction (left) and the emissive source reconstruction (right).}
    \vspace{-10pt}

\end{table}

\section{Conclusion}
\label{sec:conclusion}
We present \modelname as the first NeRF-based inverse rendering method for the scenes with emissive sources.
Our approach uses LDR images, eliminating the need of HDR images to reconstruct emissive sources.
Furthermore, we demonstrate the application of reconstructed sources in scene editing, enabling color and intensity modifications.

\noindent
\textbf{Limitations}.
Future work could explore using a single lighting condition to disentangle emissive sources, environmental lighting, and object texture. 
It is also promising to address the challenge of volume ray tracing in unbounded scenes to extend to indoor scenes.
Additionally, LTS based re-lighting may be weak in representing new colors that traverse unobserved light paths during training. 
An alternative approach could be extracting emission texture maps and modifying it using the engines such as Blender~\cite{blender} or Mitsuba~\cite{Mitsuba3}. 
More details on alternative re-lighting methods and radiance fine-tuning are provided in  Appendix.
\section{Acknowledgements}
\label{sec:acks}

This work was supported by Samsung Electronics MX, 
Basic Science Research Program through the National Research Foundation of Korea(NRF) funded by the Ministry of Education(RS-2023-00274280),
and Institute of Information \& Communications Technology Planning \& Evaluation (IITP) grant funded by the Korea government (MSIT) (No.~2019-0-01082, SW StarLab; No.~2022-0-00156, Fundamental research on continual meta-learning for quality enhancement of casual videos and their 3D metaverse transformation).
Gunhee Kim is the corresponding author.

{
    \small

}

\clearpage
\startcontents
\setcounter{page}{1}
\maketitlesupplementary

\printcontents{}{-1}{}

\section{Appendix}

\subsection{Implementation Details}

\textbf{Training Procedure}.
Our implementation builds upon Voxurf~\cite{wu2022voxurf}, excluding its dual-color network feature.
We adhere to the coarse and fine processing stages described in Voxurf before initiating our LTS learning-based training strategy.
Additionally, we compute ray colors using alpha masks to filter out points in empty space, aligning with practices in previous studies~\cite{wu2022voxurf,tensorf2022,Jin2023TensoIR}.
The LTS learning training procedure with progressive refinement approach is:

\begin{enumerate}
    \item Initialize ray groups: Uncertain rays $R^U_0=R$ and certain rays $R^C_0=\emptyset$.
    \item Form mini-batches using stratified sampling within each ray group.
    \item Calculate the rendering loss, $\mathcal{L}_{render}$.
    \item For rays in the mini-batch, uniformly sample 100 points to evaluate the LTS loss, $\mathcal{L}_{lts}$.
    \item Compute the surface normal at sampled points.
    \item For $L_o^E(x,\omega_o)$, sample an additional viewing direction on the upper hemisphere at these points.
    \item For $\hat{L}_o^E(x,\omega_o)$, sample 256 rays on the upper hemisphere at these points to compute incident radiance.
    \item Calculate $\mathcal{L}_{lts}$, considering the group membership of each point.
    \item Update network parameters.
    \item Adjust ray groups at specified training intervals.
    \item Repeat steps 2 through 10 until training ends.
\end{enumerate}

\vspace{5pt}
\noindent
\textbf{Discretization}.
Following NeuS~\cite{wang2021neus}, we approximate ray color computation using $N$ discrete points sampled along the ray, denoted as $\{x_i=c-t_i\omega_o | i=1,...,N, t_i<t_{i+1}\}$:
\begin{equation}
    \small
    \hat{C}(r)=\sum_{i=1}^{N}T_i\alpha_iL_o(x_i,\omega_o),
\end{equation}
\begin{equation}
    \small
    \alpha_i = \max\left(\frac{\Phi_s(f(x_i)) - \Phi_s(f(x_{i+1}))}{\Phi_s(f(x_i))}, 0\right),
\end{equation}
\begin{equation}
    \small
    T_i = \prod_{j=1}^{i-1}(1-\alpha_j).
\end{equation}
$\alpha$ is the discrete equivalent of the SDF-based opacity, $\rho$.

For reflections in $\hat{L}_o(x,\omega)$, we employ Monte Carlo sampling, uniformly sampling directions $\omega_i$ around the normal $n$ at point $x$ on the upper hemisphere.
While the current implementation of \modelname doesn't include importance sampling for incident rays, incorporating it in future work for variance reduction may enhance overall performance.
\begin{equation}
    \small
    \hat{L}_o(x,\omega_o)=E(x)+\frac{1}{M}\sum_{j=1}^{M}\left(\frac{L_i(x,\omega_j) R(x,\omega_o,\omega_j)}{\frac{1}{2\pi}}\right).
\end{equation}

\vspace{5pt}
\noindent
\textbf{Simplified Diseny BRDF}.
We adopt the simplified Disney principled BRDF function~\cite{walter}, parameterized by base color $b$, metallic $m$, and roughness $r$.
\begin{equation}
    \small
    \begin{aligned}
        R(x,\omega_o,\omega_i) & = \frac{D(h,n,r)F(\omega_o,h,b,m)G(\omega_o,\omega_i,h,r)}{4(n\cdot \omega_o)} \\
                               & + \left(n\cdot \omega_i\right)\left(1 - m\right) \left(\frac{b}{\pi}\right) ,
    \end{aligned}
\end{equation}
The half vector $h$ is defined as $h=\frac{\omega_o + \omega_i}{\lVert \omega_o + \omega_i \rVert_2}$.
Following NeILF++~\cite{zhang2023neilf++}, the normal distribution function $D$ is approximated using Spherical Gaussian:
\begin{equation}
    \small
    D(h,n,r) = \frac{1}{\pi r^4}\exp(\frac{2}{r^4}(h\cdot n - 1)),
\end{equation}
The Fresnel term $F$ is calculated as follows:
\begin{equation}
    \small
    \begin{aligned}
        F(\omega_o,h,b,m) & =F_0 + (1-F_0)(1-(\omega_o \cdot h)^5), \\
        \text{where }F_0  & =0.04(1-m) + bm,
    \end{aligned}
\end{equation}
The geometry term $G$ adopts the GGX function~\cite{DisneyBRDF}.
\begin{equation}
    \small
    \begin{aligned}
        G(\omega_o, \omega_i,n,r) & = \frac{(n\cdot \omega_o)(n \cdot \omega_i)}{\left( (n\cdot \omega_o)(1-k) + k \right) \left((n\cdot\omega_i)(1-k)+k\right)}, \\
        \text{where } k           & = \frac{r^2}{2}.
    \end{aligned}
\end{equation}
For simplicity, our BRDF model incorporates the Lambert cosine term $(n\cdot\omega_i)$.

\vspace{5pt}
\noindent
\textbf{Gamma Correction}.
To ensure HDR linear color space for outgoing radiance, we apply the standard gamma correction as defined by IEC~\cite{iec61966-2-1} to ray colors before calculating the rendering loss.
The gamma-corrected sRGB color, given a linear color $C_{\text{linear}}$, is computed as follows:
\begin{equation}
    \tau(C_{\text{linear}}) =
    \begin{cases}
        12.92C_{\text{linear}}                 & \text{if } C_{\text{linear}} \leq 0.0031308, \\
        1.055C_{\text{linear}}^{1/2.4} - 0.055 & \text{if } C_{\text{linear}} > 0.0031308.
    \end{cases}
\end{equation}

\vspace{5pt}
\noindent
\textbf{RGB to HSV}.
For scene editing tasks, we utilize the HSV color model~\cite{HSV}.
The hue ($H\in[0,1]$), saturation ($S\in[0,1]$), and value ($V\in\mathbb{R}_{+}$) are calculated using the following method:

\begin{equation}
    \small
    \begin{aligned}
        M & = \max(R,G,B), \\
        m & = \min(R,G,B), \\
        C & = M - m.
    \end{aligned}
\end{equation}

\begin{equation}
    \small
    \begin{aligned}
        H  & = (H' / 6.0) \mod 1.0, \\
        H' & =
        \begin{cases}
            0               & \text{if } C = 0, \\
            \frac{G - B}{C} & \text{if } M=R,   \\
            \frac{B-R}{C}+2 & \text{if } M=G,   \\
            \frac{R-G}{C}+4 & \text{if } M=B.
        \end{cases}
    \end{aligned}
\end{equation}

\begin{equation}
    \small
    S =
    \begin{cases}
        0           & \text{if } V = 0, \\
        \frac{C}{V} & \text{otherwise},
    \end{cases}
\end{equation}

\begin{equation}
    \small
    V = \max(R,G,B)
\end{equation}

\vspace{5pt}
\noindent
\textbf{HSV to RGB}.
Once the color is replaced and intensity is adjusted in the HSV space, the conversion back to RGB is performed as:
\begin{equation}
    \small
    \begin{aligned}
        m          & = V - C,                                    \\
        H'         & = H \times 6.0,                             \\
        C          & = S\times V,                                \\
        X          & = C\times (1 - \lvert H'\mod 2 - 1 \rvert), \\
        (R',G',B') & =
        \begin{cases}
            (C, X, 0) & \text{if } 0\leq H' < 1, \\
            (X, C, 0) & \text{if } 1\leq H' < 2, \\
            (0, C, X) & \text{if } 2\leq H' < 3, \\
            (0, X, C) & \text{if } 3\leq H' < 4, \\
            (X, 0, C) & \text{if } 4\leq H' < 5, \\
            (C, 0, X) & \text{if } 5\leq H' < 6.
        \end{cases}
    \end{aligned}
\end{equation}

\begin{equation}
    \small
    (R,G,B) =(R'+m,G'+m,B'+m) .
\end{equation}

\subsection{Dataset Details}
\label{sec:dataset details}
\textbf{Dataset Construction}.
This section outlines the dataset used for training and evaluation.
Each scene in our dataset comprises 200 training images, with an equal split between two lighting conditions: ``on'' and ``off''.
Emission masks are utilized as ground truth for emissive source identification, while EXR files with linear pixel values assess the accuracy of the reconstructed strength of emission and reflection.
All data are rendered using the Cycles path tracing in Blender~\cite{blender}, with settings that could artificially alter scene illumination are disabled, such as incdient light clamping and the Filmic transform.
For scene editing under novel lighting conditions, we introduce a variety of test scenarios, including intensity editing, color editing, and combined intensity and color editing, each with 50 images.
We derive these scenarios from 25 unique camera positions from the novel view evaluation dataset, each under two different lighting conditions,
Intensity adjustments are made relative to the original scene's emissive source strength, with ``0'' indicating ``light off'' and ``1'' matching the ``light on'' intensity.
We test intensity adjustments at half (0.5) and double (2.0) the original levels.
In scenes allowing individual source adjustments, we include an additional intensity condition where lights are selectively turned off (0.0).
For color editing, we select six colors—red, green, blue, cyan, magenta, and yellow—to demonstrate the effects of various light source colors on scene illumination.

\vspace{5pt}
\noindent
\textbf{Scene Characteristics}.
Our scenes are meticulously crafted using assets from Blendswap and cgtrader, with licensing details and the count of emissive sources detailed in Tab.~\ref{tab:scene-characteristics}.
Below, we describe the unique aspects of each scene.

\begin{table*}
    \centering
    \scriptsize
    \begin{tabular}{@{}l|c|p{13cm}@{}}
        \toprule
        \textbf{Scene Name}   & \textbf{Num Lights} & \textbf{License}                                                                                                      \\
        \midrule
        Lego                  & 3                   & By Heinzelnisse (CC-BY-NC): \url{https://www.blendswap.com/blend/11490}                                               \\
        \midrule
        \multirow{3}{*}{Gift} & \multirow{3}{*}{29} & By juan215 (Royalty Free):                                                                                            \\
                              &                     & \url{https://www.cgtrader.com/free-3d-models/household/household-tools/gift-box-aeb8f01e-929f-4041-9117-bcea21f3c813} \\
                              &                     & By MiriamAHoyt (CC-0): \url{https://blendswap.com/blend/21434}                                                        \\
        \midrule
        \multirow{3}{*}{Book} & \multirow{3}{*}{1}  & By lakerice (CC-0): \url{https://blendswap.com/blend/22197}                                                           \\
                              &                     & By 3dfiles (CC-BY): \url{https://blendswap.com/blend/28034}                                                           \\
                              &                     & By bloknayrb (CC-BY): \url{https://www.blendswap.com/blend/26172}                                                     \\
        \midrule
        \multirow{2}{*}{Cube} & \multirow{2}{*}{1}  & By 4NDR31JK (CC-BY): \url{https://www.blendswap.com/blend/30149}                                                      \\
                              &                     & By sriniwasjha (CC-BY): \url{https://blendswap.com/blend/18409}                                                       \\
        \midrule
        Billboard             & 6                   & By M0h4wkAD3 (CC-BY-NC-SA): \url{https://blendswap.com/blend/27481}                                                   \\
        \midrule
        Balls                 & 1                   & By elbrujodelatribu (CC-0): \url{https://blendswap.com/blend/10120}                                                   \\

        \bottomrule
    \end{tabular}
    \caption{Number of emissive sources and licenses of objects used in scenes.}
    \label{tab:scene-characteristics}
\end{table*}

\begin{itemize}
    \item \textbf{LEGO}:
          This scene showcases three emissive sources, all starting with the same color and intensity.
          The intricate designs of the LEGO bricks create complex reflection effects.
          The emissive sources in these scenes are tested for both collective and individual adjustments.

          \begin{figure}[h]
              \centering
              \begin{subfigure}{0.42\linewidth}
                  \includegraphics[width=\linewidth,valign=m]{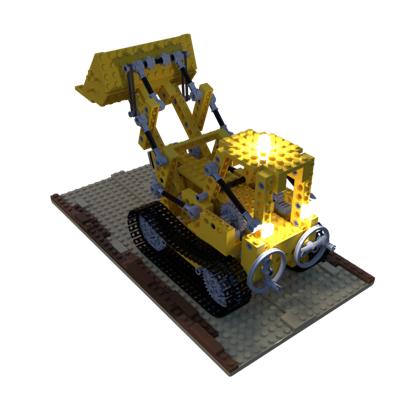}
                  \caption{Lego (white)}
              \end{subfigure}
              \begin{subfigure}{0.42\linewidth}
                  \includegraphics[width=\linewidth,valign=m]{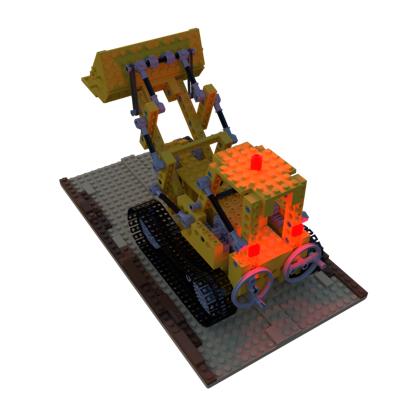}
                  \caption{Lego (vivid)}
              \end{subfigure}
          \end{figure}

    \item \textbf{Gift}:
          Featuring a gift box, a toy, and numerous small bulbs, this scene presents a challenge with its multitude of tiny light bulbs and extensive reflection areas.

          \begin{figure}[h!]
              \centering
              \begin{subfigure}{0.42\linewidth}
                  \includegraphics[width=\linewidth,valign=m]{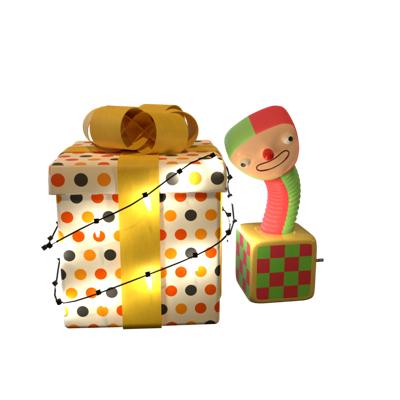}
                  \caption{Gift (white)}
              \end{subfigure}
              \begin{subfigure}{0.42\linewidth}
                  \includegraphics[width=\linewidth,valign=m]{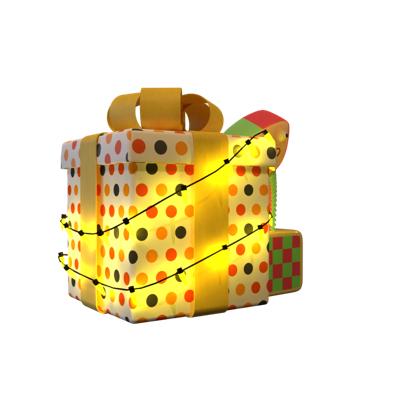}
                  \caption{Gift (vivid)}
              \end{subfigure}
          \end{figure}

    \item \textbf{Book}:
          The Book scene features a single large light source consisting of a lamp, a book, and a pencil.
          The emphasis here is on identifying and restoring the very large emissive source.

          \begin{figure}[h!]
              \centering
              \begin{subfigure}{0.42\linewidth}
                  \includegraphics[width=\linewidth,valign=m]{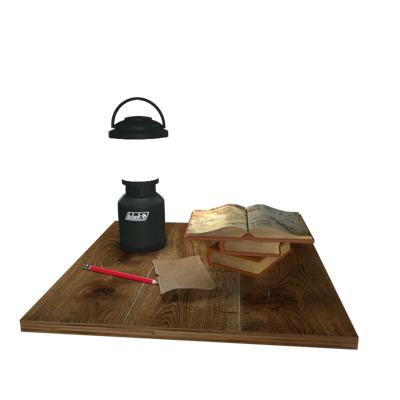}
                  \caption{Book (white)}
              \end{subfigure}
              \begin{subfigure}{0.42\linewidth}
                  \includegraphics[width=\linewidth,valign=m]{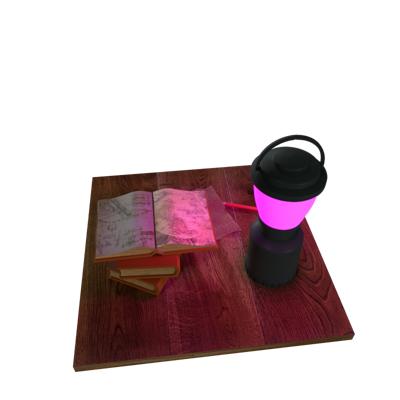}
                  \caption{Book (vivid)}
              \end{subfigure}
          \end{figure}

    \item \textbf{Cube}:
          Comprising a tablet PC and a cube, this scene is marked by its sophisticated reflection effects, especially on the cube surfaces which varying albedo.

          \begin{figure}[h!]
              \centering
              \begin{subfigure}{0.42\linewidth}
                  \includegraphics[width=\linewidth,valign=m]{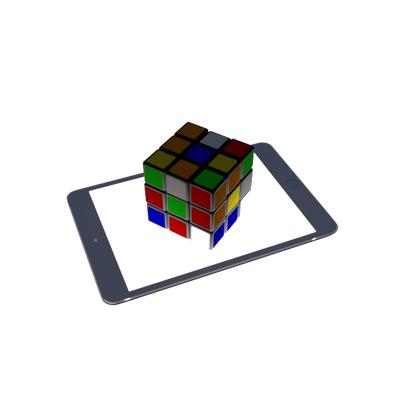}
                  \caption{Cube (white)}
              \end{subfigure}
              \begin{subfigure}{0.42\linewidth}
                  \includegraphics[width=\linewidth,valign=m]{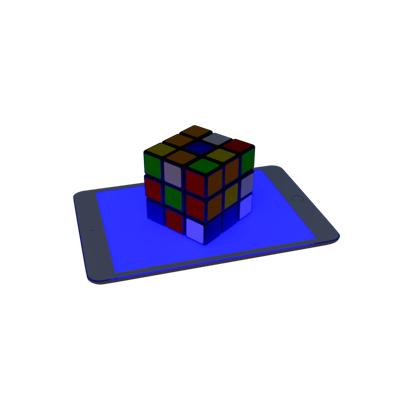}
                  \caption{Cube (vivid)}
              \end{subfigure}
          \end{figure}

    \item \textbf{Billboard}:
          This scene includes two billboards, each equipped with three emissive sources, summing up to six sources.
          The lights are positioned to shine downwards from the billboards' tops.
          We adjust the emissive sources collectively and individually.
          Individual adjustments are performed for three light groups by pairing the light sources of the front and back billboards.

          \begin{figure}[h!]
              \centering
              \begin{subfigure}{0.42\linewidth}
                  \includegraphics[width=\linewidth,valign=m]{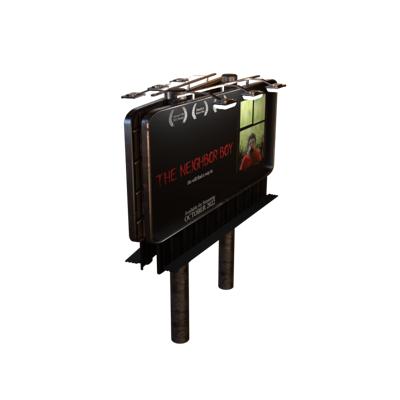}
                  \caption{Billboard (white)}
              \end{subfigure}
              \begin{subfigure}{0.42\linewidth}
                  \includegraphics[width=\linewidth,valign=m]{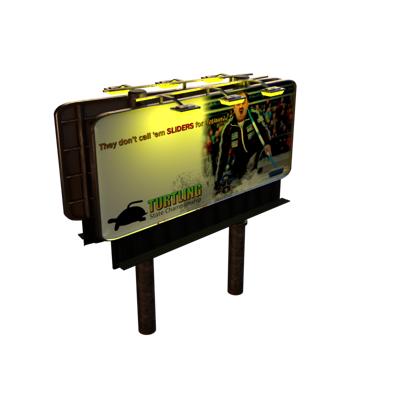}
                  \caption{Billboard (vivid)}
              \end{subfigure}
          \end{figure}

    \item \textbf{Balls}:
          This is the material balls scene in NeRF, with the modification of the red ball as an emissive source.

          \begin{figure}[h!]
              \centering
              \begin{subfigure}{0.42\linewidth}
                  \includegraphics[width=\linewidth,valign=m]{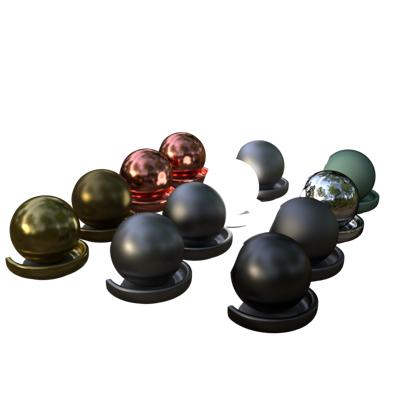}
                  \caption{Balls (white)}
              \end{subfigure}
              \begin{subfigure}{0.42\linewidth}
                  \includegraphics[width=\linewidth,valign=m]{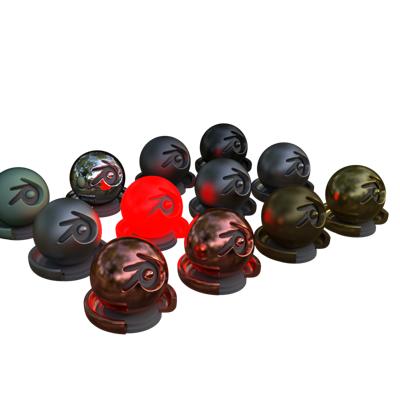}
                  \caption{Balls (vivid)}
              \end{subfigure}
          \end{figure}

\end{itemize}

\begin{figure*}
    \centering
    \scriptsize
    \renewcommand{\arraystretch}{3} 
    \begin{tabular}{@{} *{7}{c@{\hspace{2pt}}} @{}}
        \multicolumn{1}{c}{Image}                                                                                   & \multicolumn{1}{c}{Normal}                                                                            & \multicolumn{1}{c}{Base Color}                                                                           & \multicolumn{1}{c}{Roughness}                                                                            & \multicolumn{1}{c}{Metallic}                                                                            & \multicolumn{1}{c}{Env. Map}                                                                      \\
        \includegraphics[valign=m,width=.15\textwidth]{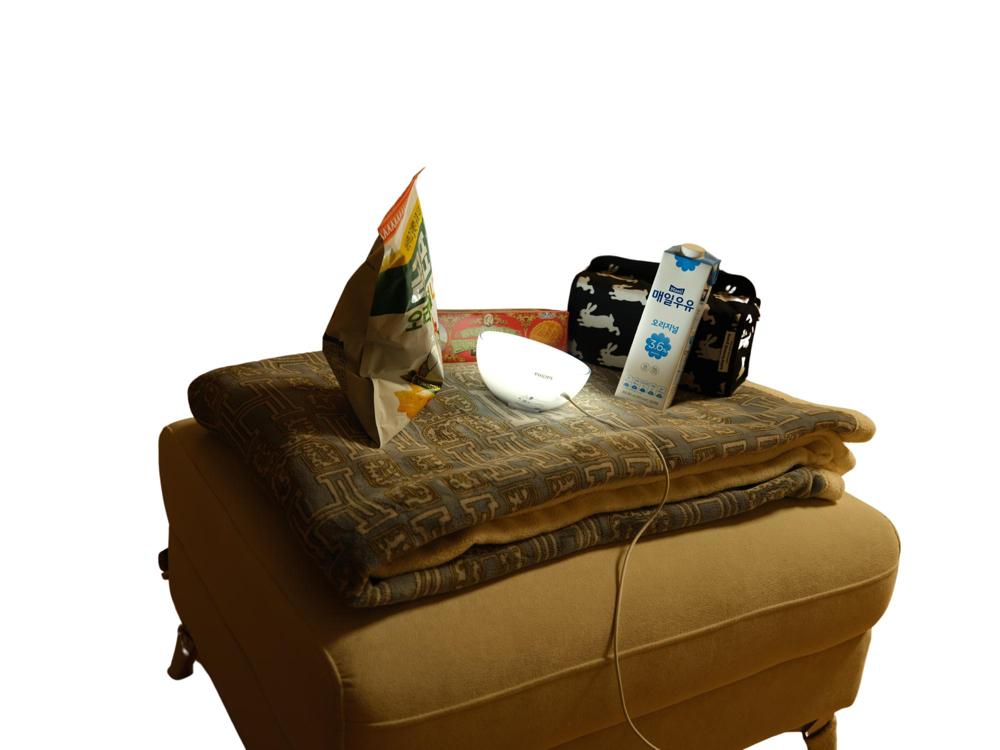}        & \includegraphics[valign=m,width=.15\textwidth]{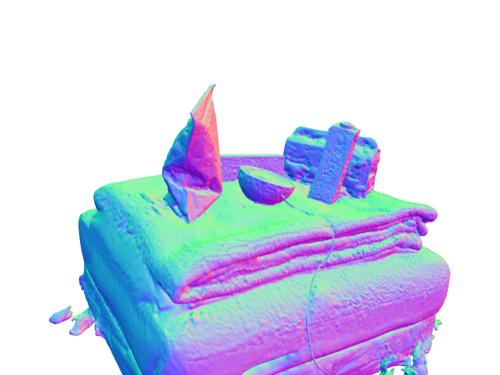}        & \includegraphics[valign=m,width=.15\textwidth]{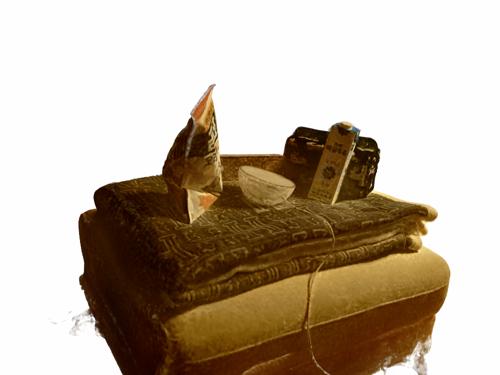}        & \includegraphics[valign=m,width=.15\textwidth]{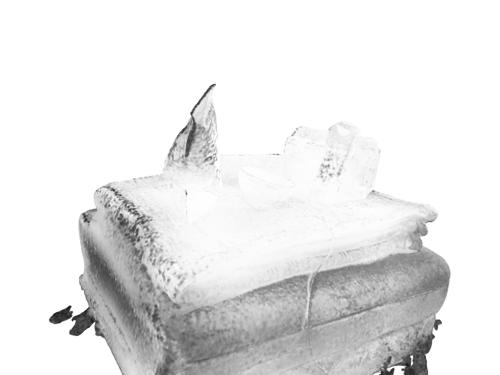}        & \includegraphics[valign=m,width=.15\textwidth]{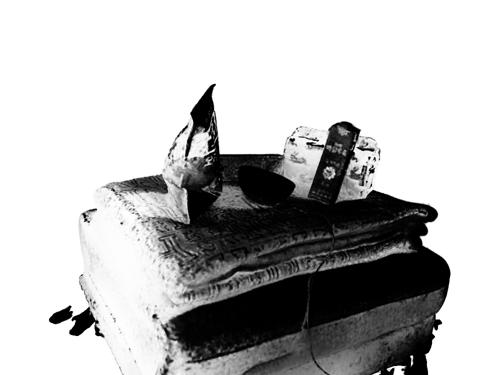}        & \includegraphics[valign=m,width=.225\textwidth]{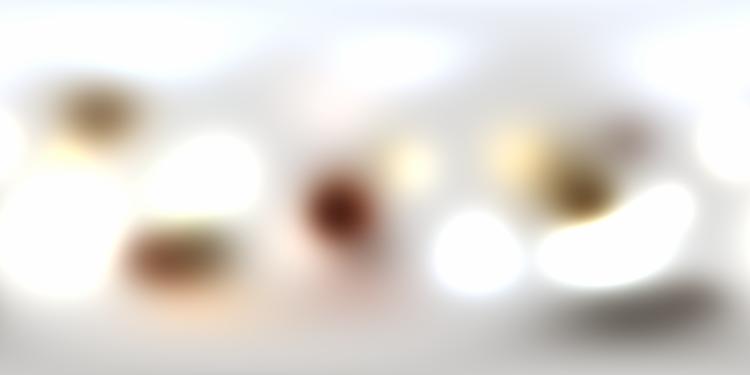}   \\
        \includegraphics[valign=m,width=.15\textwidth]{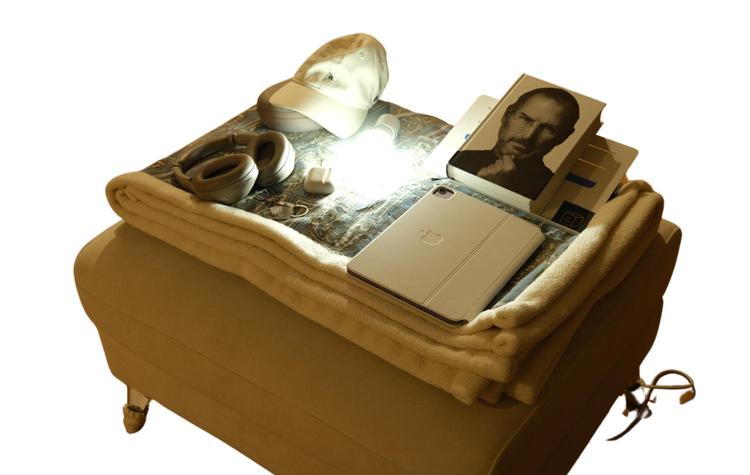}     & \includegraphics[valign=m,width=.15\textwidth]{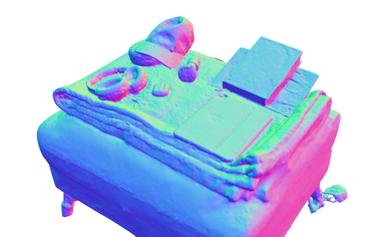}     & \includegraphics[valign=m,width=.15\textwidth]{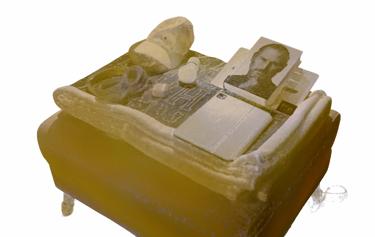}     & \includegraphics[valign=m,width=.15\textwidth]{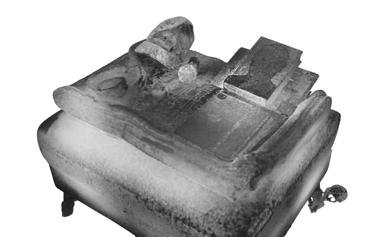}     & \includegraphics[valign=m,width=.15\textwidth]{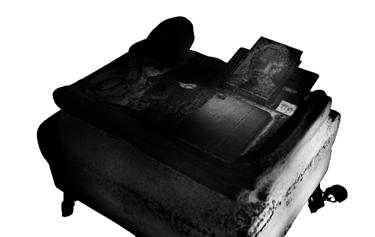}     & \includegraphics[valign=m,width=.225\textwidth]{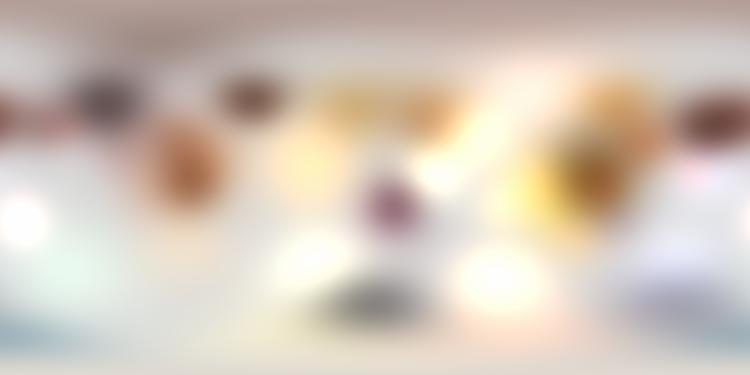}     \\
        \includegraphics[valign=m,width=.15\textwidth]{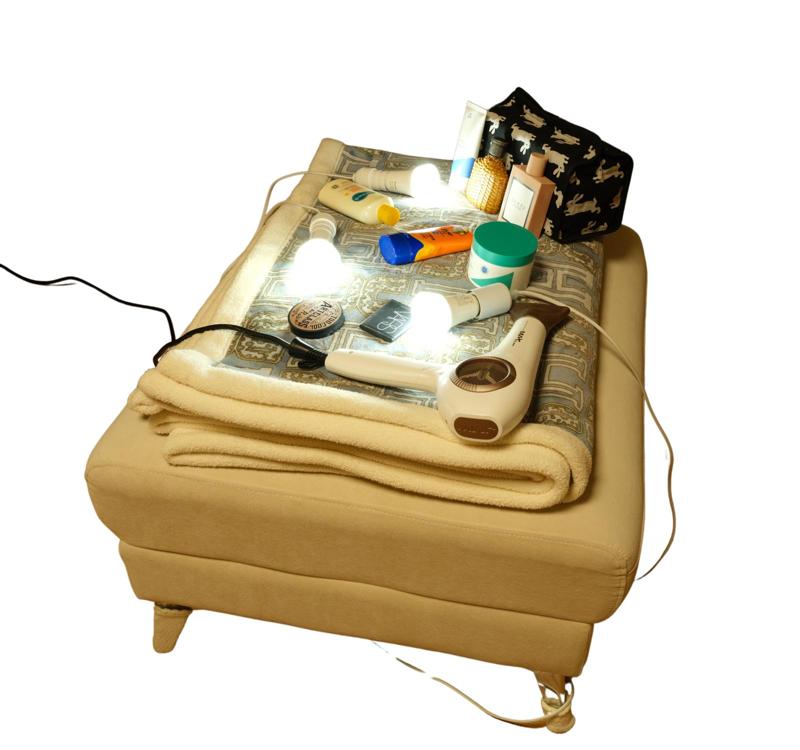} & \includegraphics[valign=m,width=.15\textwidth]{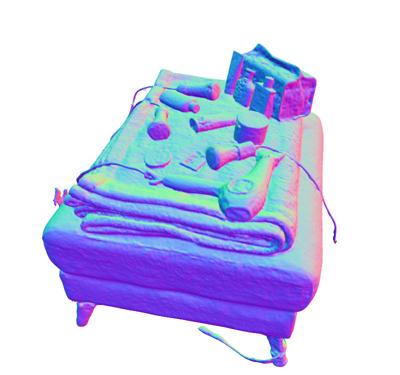} & \includegraphics[valign=m,width=.15\textwidth]{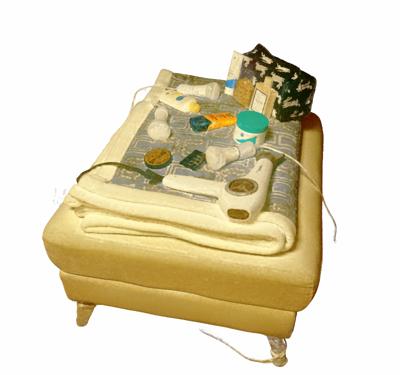} & \includegraphics[valign=m,width=.15\textwidth]{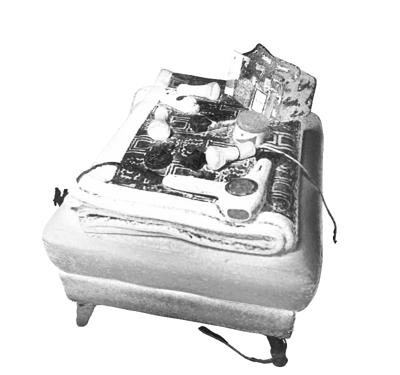} & \includegraphics[valign=m,width=.15\textwidth]{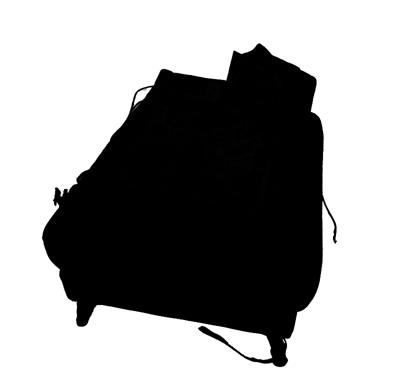} & \includegraphics[valign=m,width=.225\textwidth]{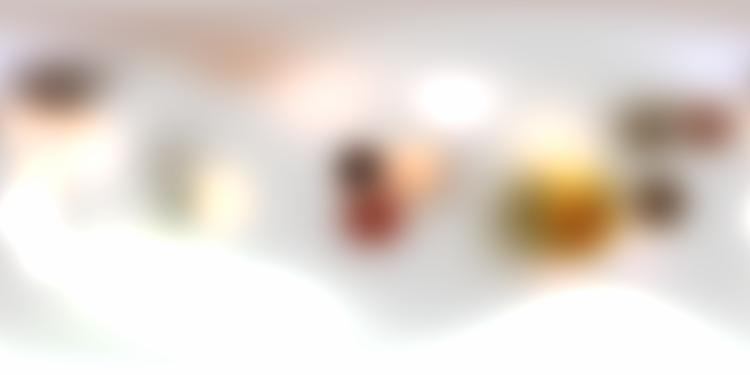} \\
        \includegraphics[valign=m,width=.15\textwidth]{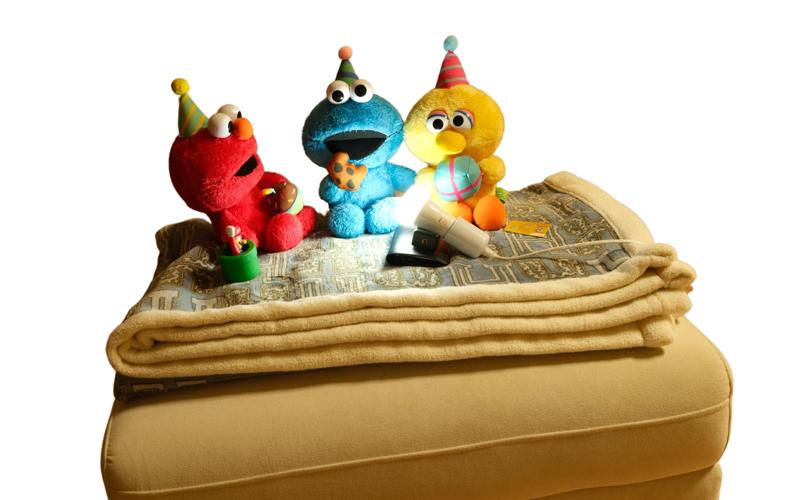}    & \includegraphics[valign=m,width=.15\textwidth]{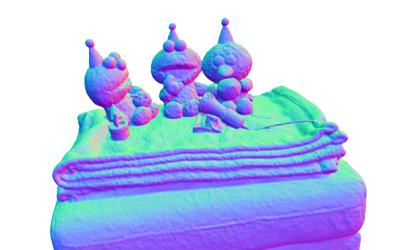}    & \includegraphics[valign=m,width=.15\textwidth]{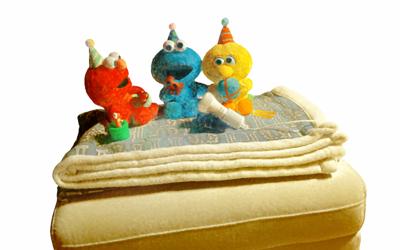}    & \includegraphics[valign=m,width=.15\textwidth]{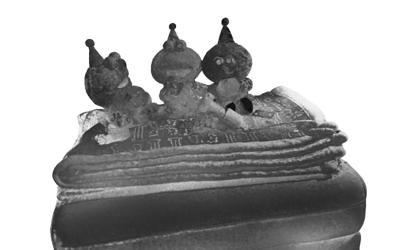}    & \includegraphics[valign=m,width=.15\textwidth]{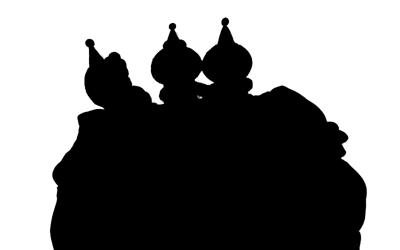}    & \includegraphics[valign=m,width=.225\textwidth]{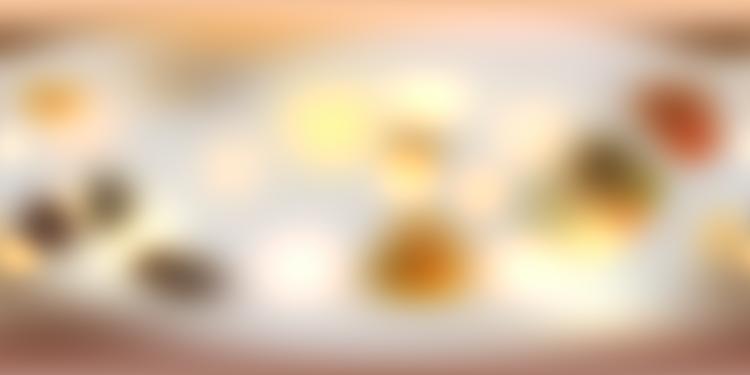}    \\
    \end{tabular}
    \caption{Decomposed scene components on real scenes}
    \label{fig:real_decomp}
\end{figure*}

\begin{figure*}
    \centering
    \scriptsize
    \renewcommand{\arraystretch}{3} 
    \begin{tabular}{@{} *{7}{c@{\hspace{2pt}}} @{}}
        \multicolumn{1}{c}{Image}                                                                                   & \multicolumn{1}{c}{Emission}                                                                     & \multicolumn{1}{c}{Edit 1}                                                                     & \multicolumn{1}{c}{Edit 2}                                                                     \\
        \includegraphics[valign=m,width=.22\textwidth]{figures/appendix/Real/snacks_resize/input_resize.jpg}        & \includegraphics[valign=m,width=.22\textwidth]{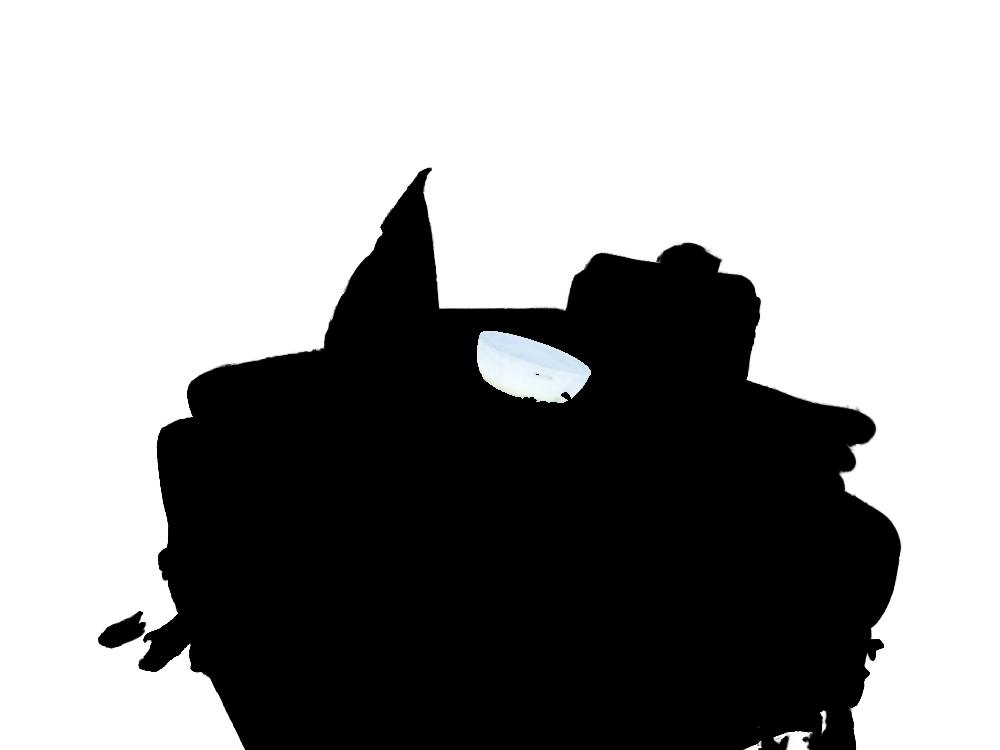}        & \includegraphics[valign=m,width=.22\textwidth]{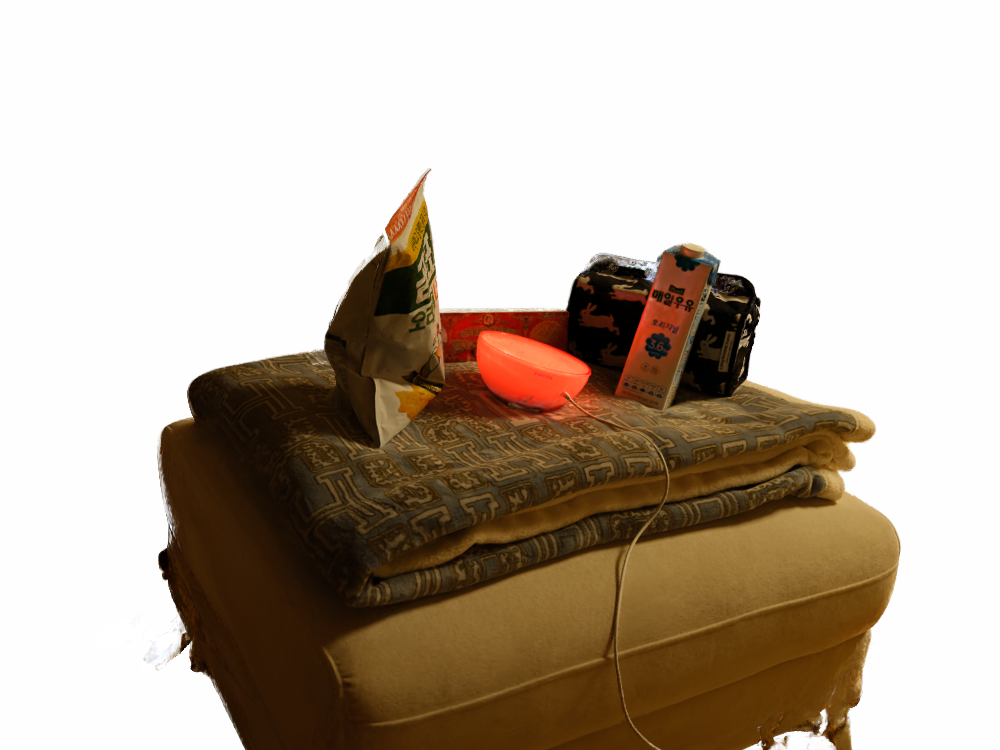}          & \includegraphics[valign=m,width=.22\textwidth]{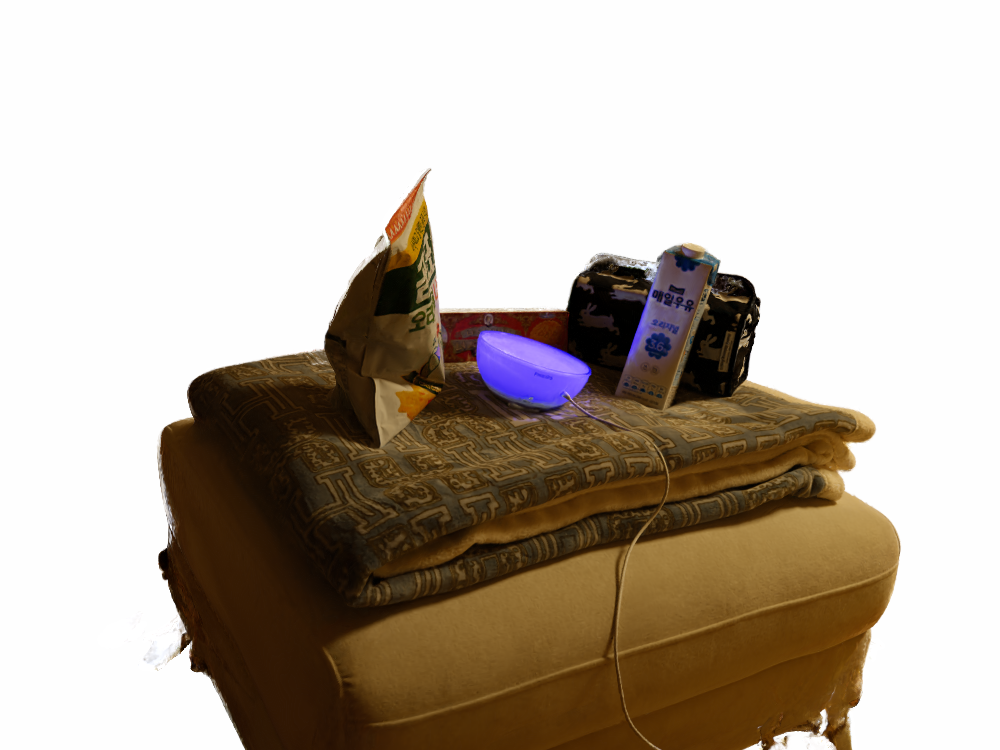}         \\
        \includegraphics[valign=m,width=.22\textwidth]{figures/appendix/Real/jobs_resize/input_crop_resize.jpg}     & \includegraphics[valign=m,width=.22\textwidth]{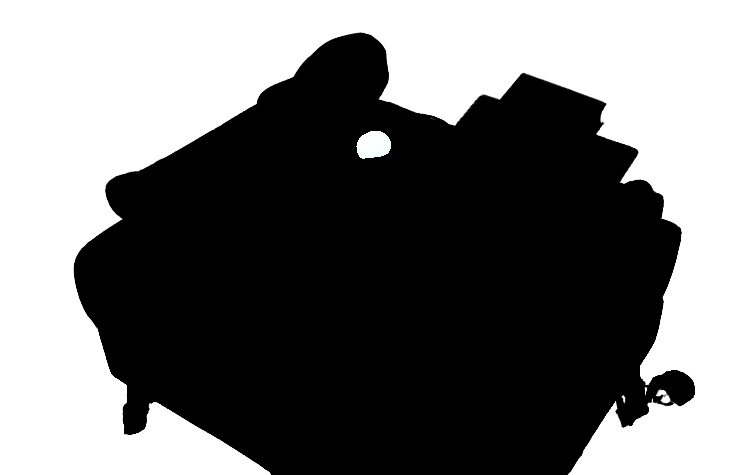}     & \includegraphics[valign=m,width=.22\textwidth]{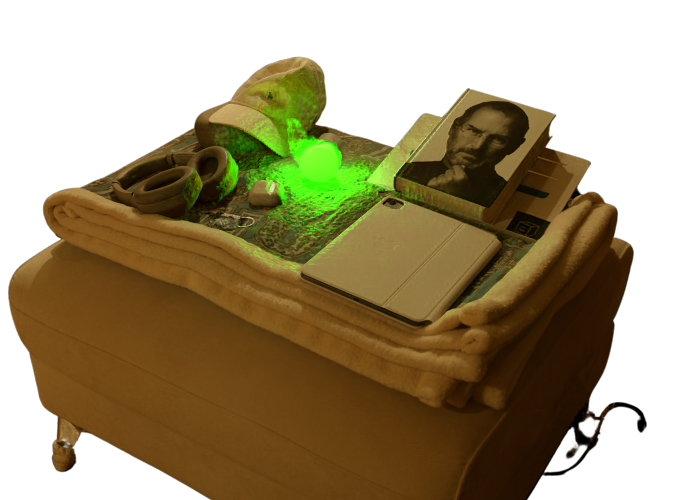} & \includegraphics[valign=m,width=.22\textwidth]{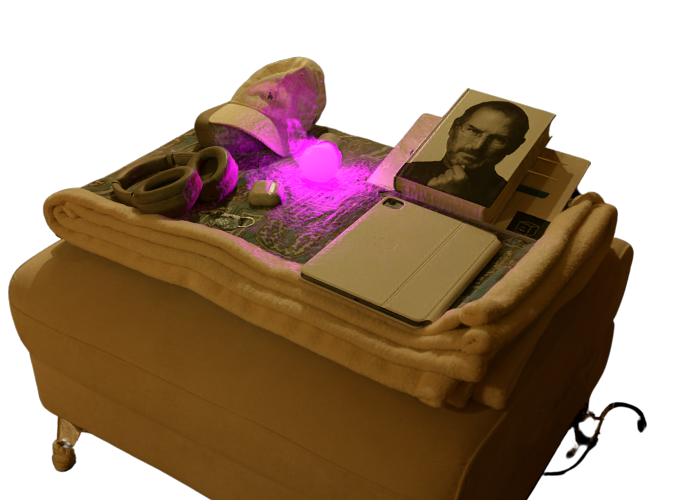} \\
        \includegraphics[valign=m,width=.22\textwidth]{figures/appendix/Real/cosmetic_resize/input_crop_resize.jpg} & \includegraphics[valign=m,width=.22\textwidth]{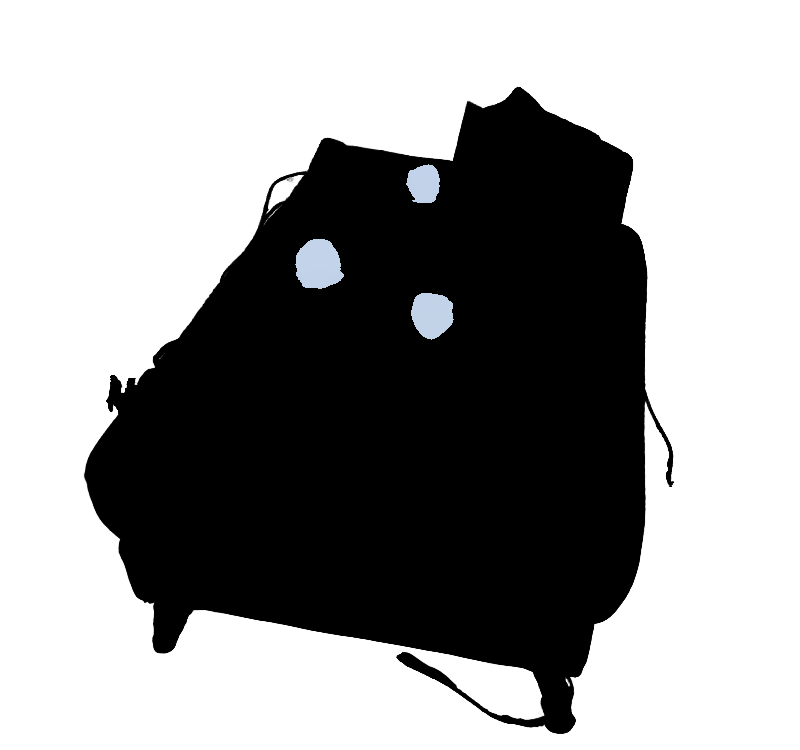} & \includegraphics[valign=m,width=.22\textwidth]{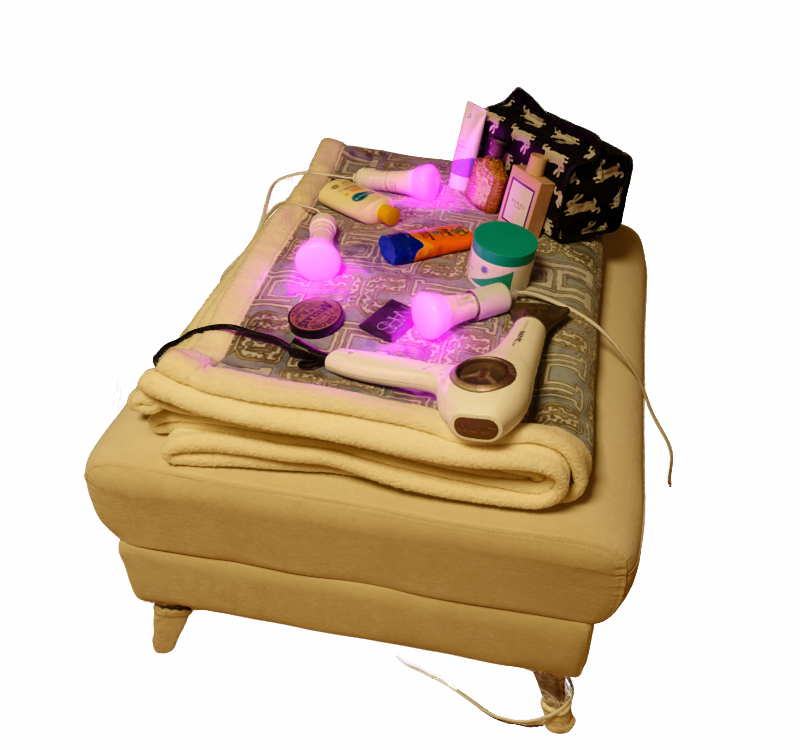}  & \includegraphics[valign=m,width=.22\textwidth]{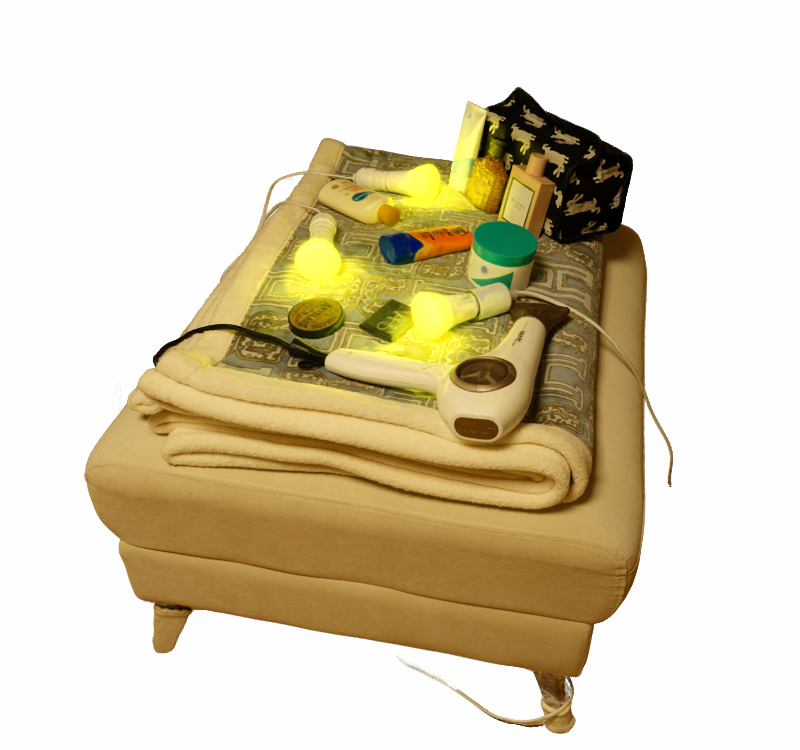}  \\
        \includegraphics[valign=m,width=.22\textwidth]{figures/appendix/Real/dolls_resize/input_crop_resize.jpg}    & \includegraphics[valign=m,width=.22\textwidth]{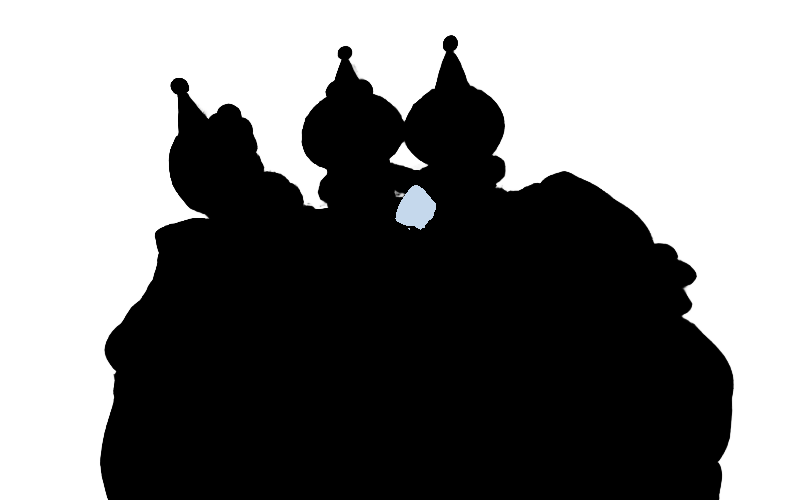}    & \includegraphics[valign=m,width=.22\textwidth]{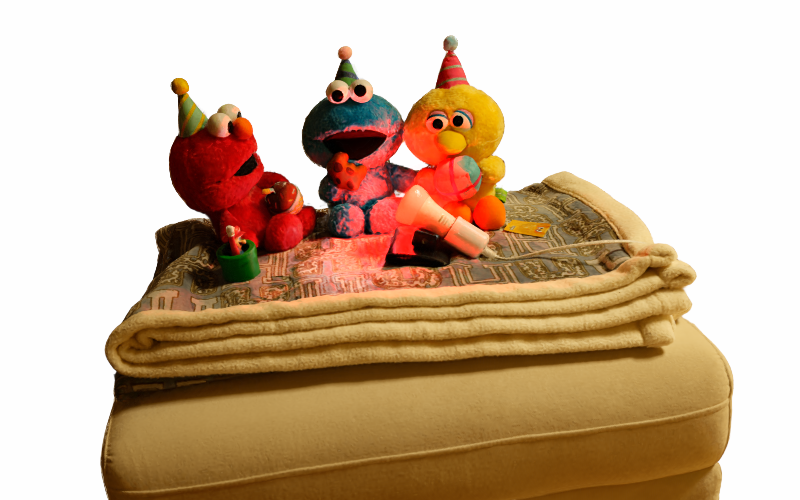}     & \includegraphics[valign=m,width=.22\textwidth]{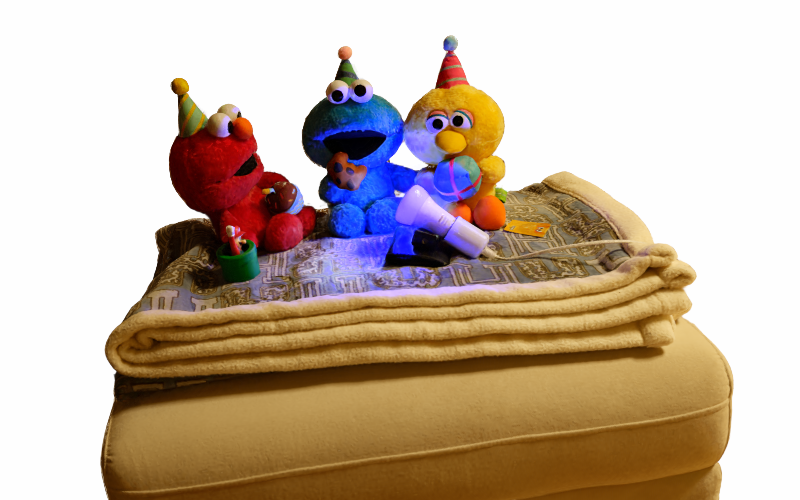}     \\
    \end{tabular}
    \caption{Identified emissive sources and edited results on real scenes.}
    \label{fig:real_em}
\end{figure*}

\begin{figure*}
    \centering
    \scriptsize
    \renewcommand{\arraystretch}{3} 
    \begin{tabular}{@{} *{7}{c@{\hspace{2pt}}} @{}}
        \multicolumn{1}{c}{Image}                                                                                      & \multicolumn{1}{c}{Edit}                                                                         & \multicolumn{1}{c}{G.T.}                                                                                          & \multicolumn{1}{c}{Edit}                                                                         & \multicolumn{1}{c}{G.T.}                                                                                          \\
        \includegraphics[valign=m,width=.195\textwidth]{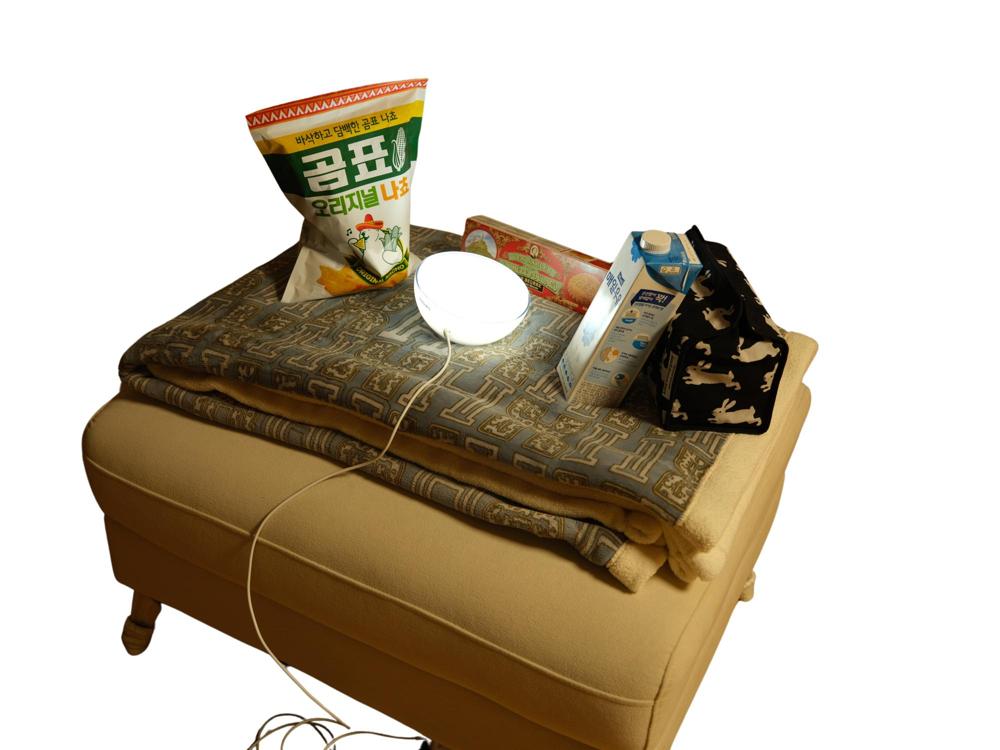}   & \includegraphics[valign=m,width=.195\textwidth]{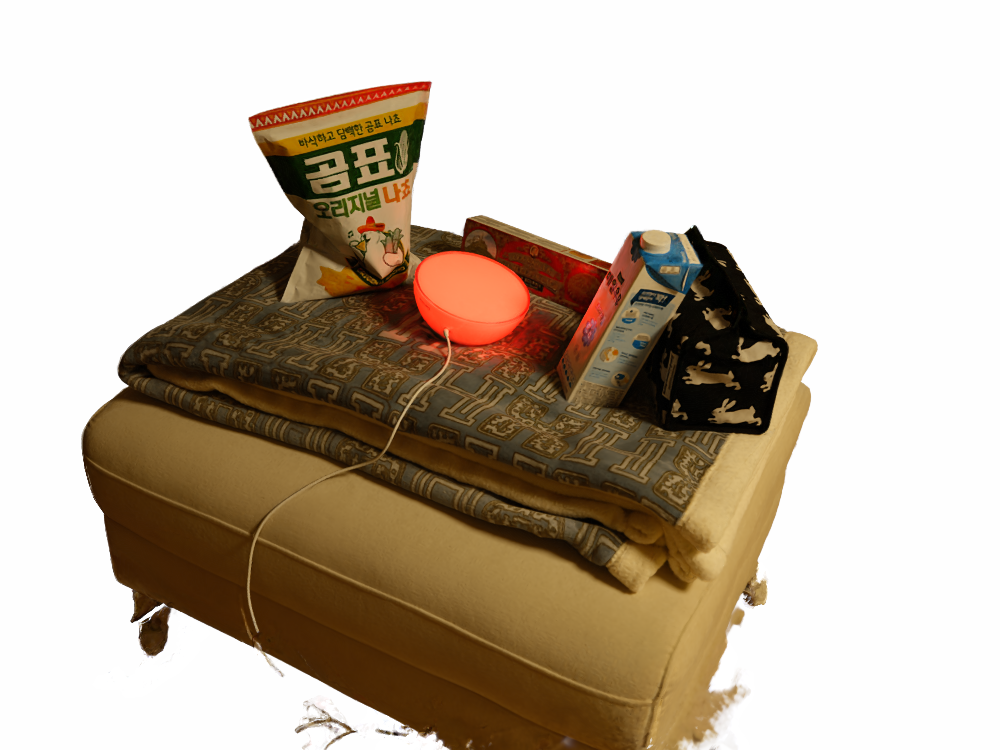}   & \includegraphics[valign=m,width=.195\textwidth]{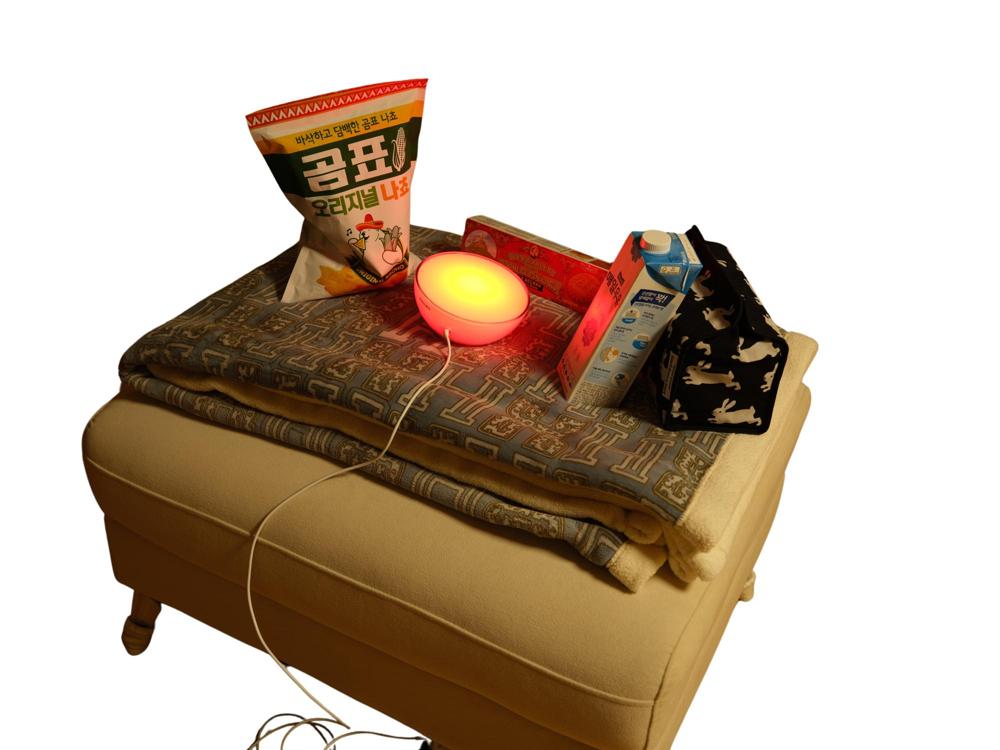}        & \includegraphics[valign=m,width=.195\textwidth]{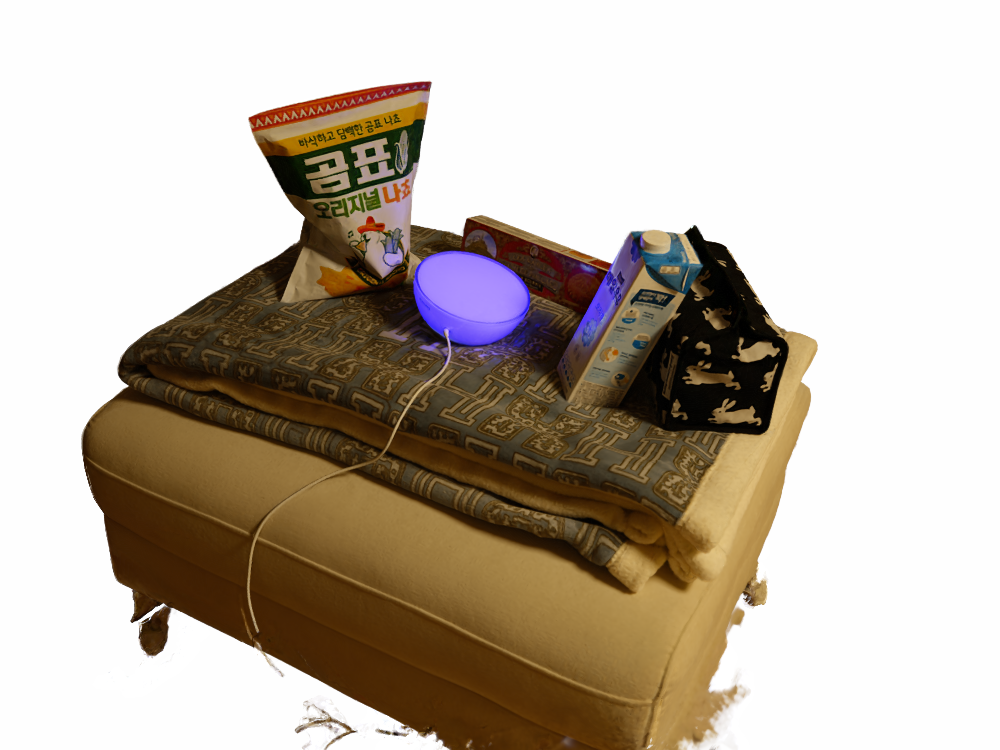}   & \includegraphics[valign=m,width=.195\textwidth]{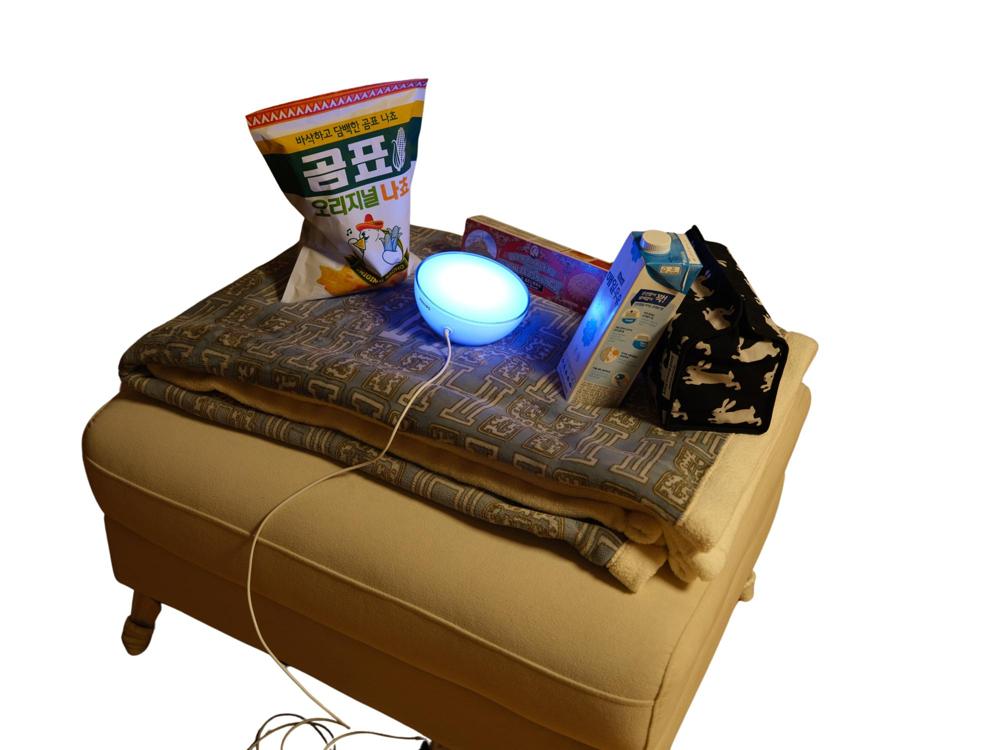}        \\
        \includegraphics[valign=m,width=.195\textwidth]{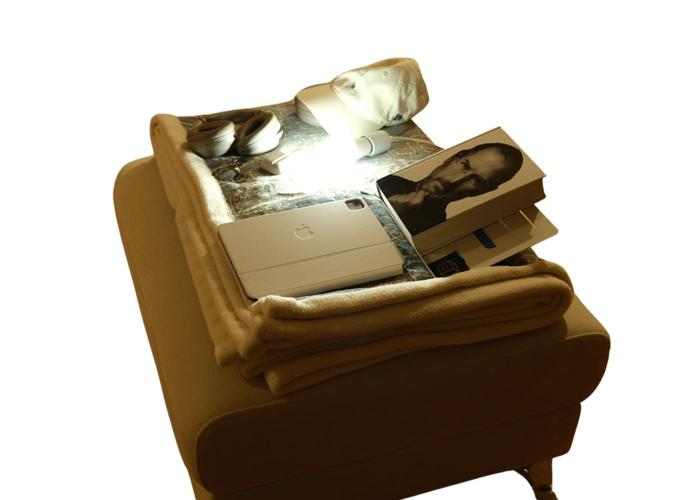}    & \includegraphics[valign=m,width=.195\textwidth]{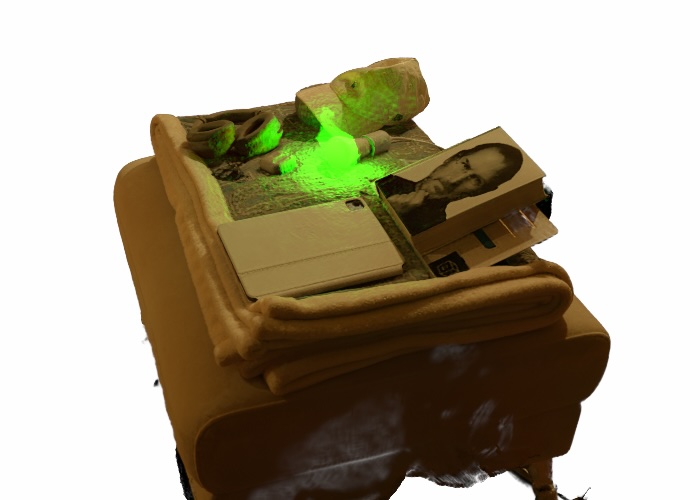}      & \includegraphics[valign=m,width=.195\textwidth]{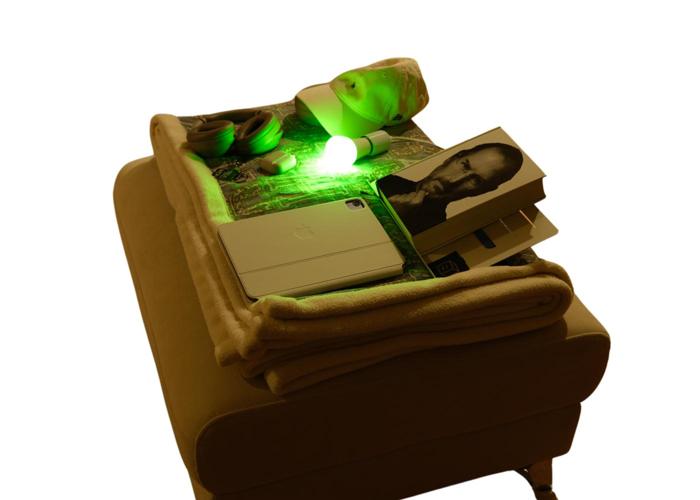}    & \includegraphics[valign=m,width=.195\textwidth]{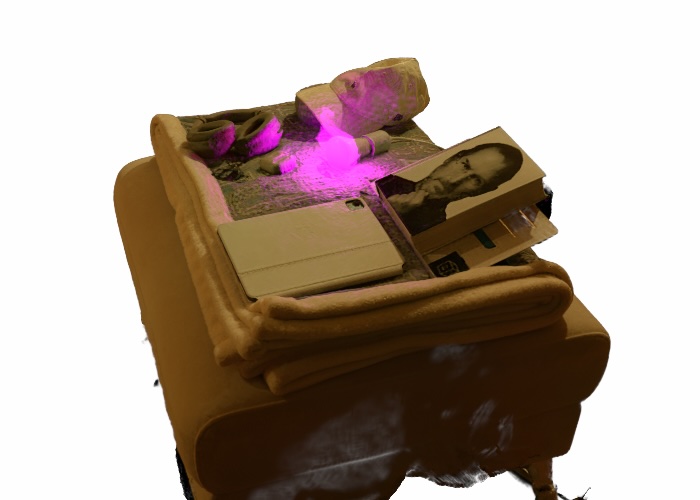}      & \includegraphics[valign=m,width=.195\textwidth]{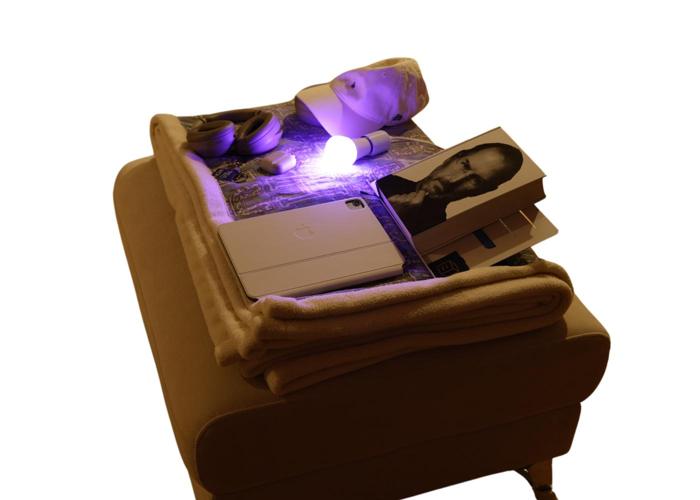}    \\
        \includegraphics[valign=m,width=.195\textwidth]{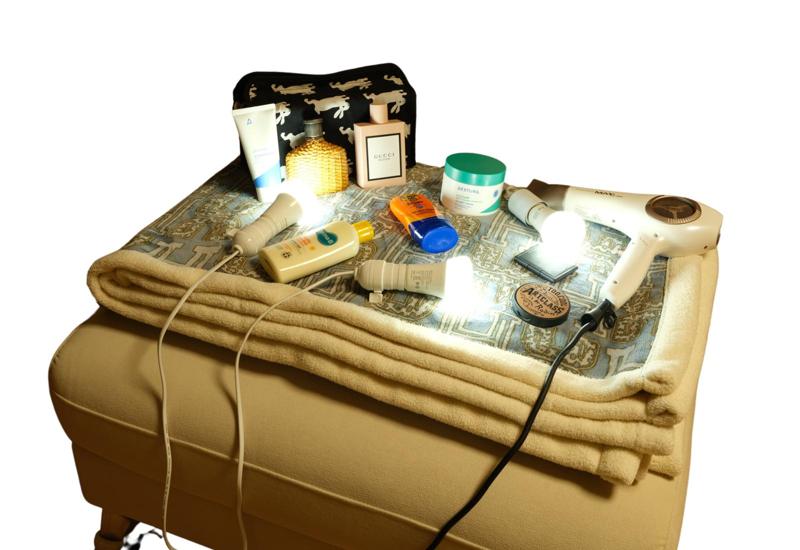} & \includegraphics[valign=m,width=.195\textwidth]{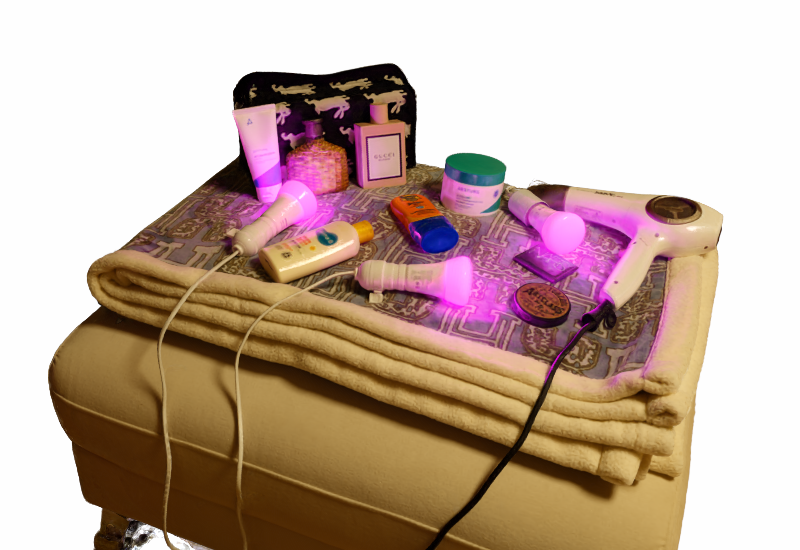} & \includegraphics[valign=m,width=.195\textwidth]{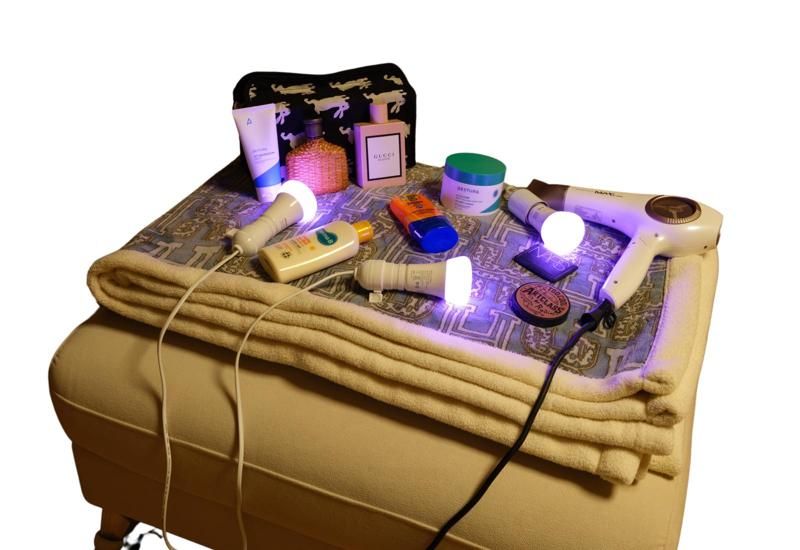} & \includegraphics[valign=m,width=.195\textwidth]{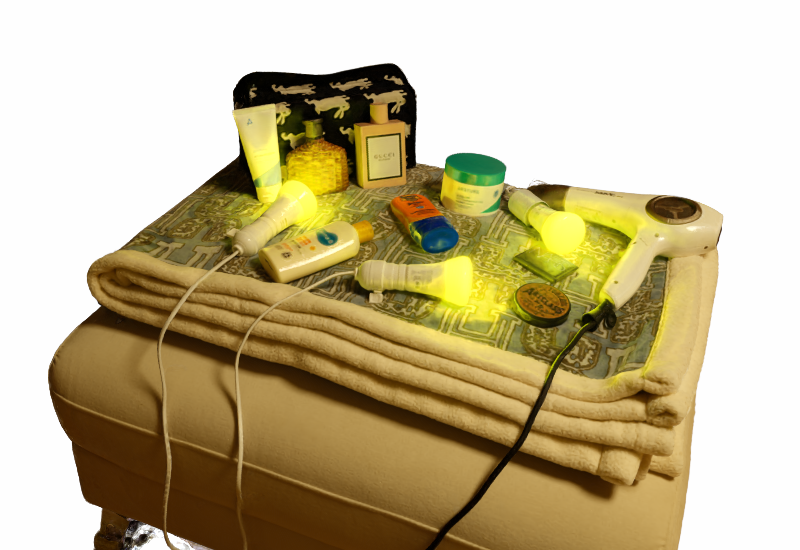} & \includegraphics[valign=m,width=.195\textwidth]{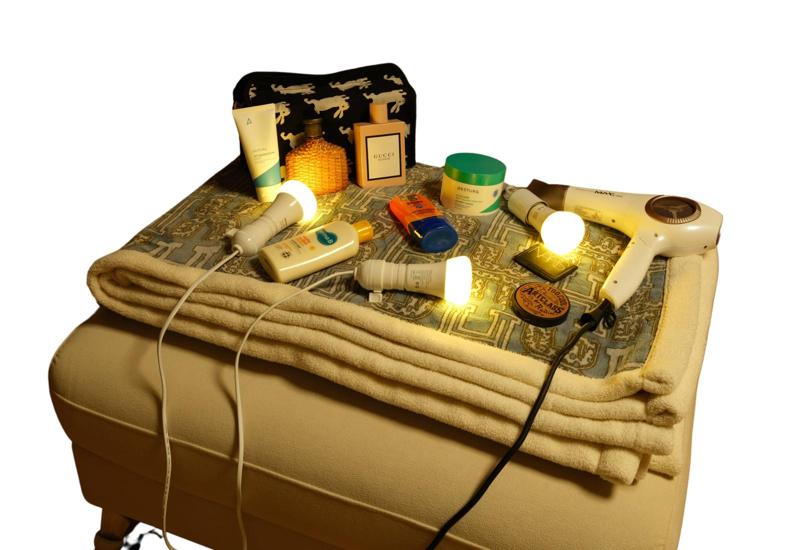} \\
        \includegraphics[valign=m,width=.195\textwidth]{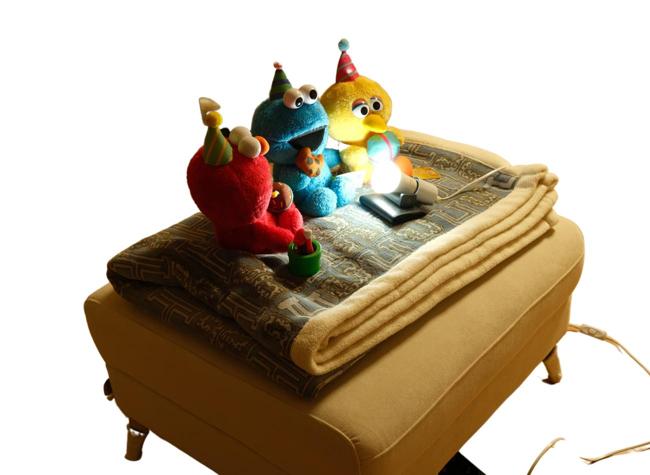}   & \includegraphics[valign=m,width=.195\textwidth]{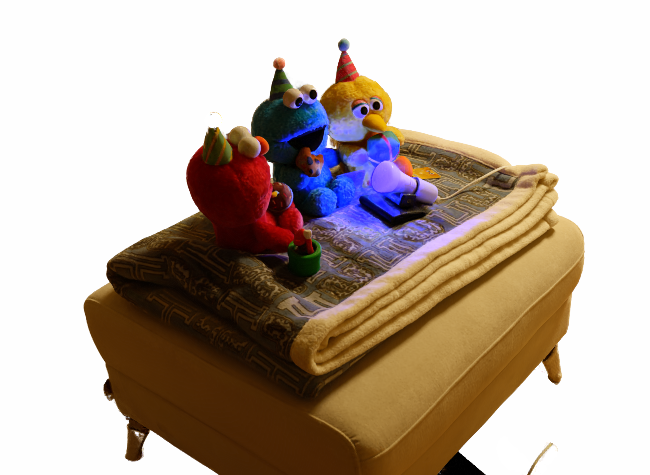}   & \includegraphics[valign=m,width=.195\textwidth]{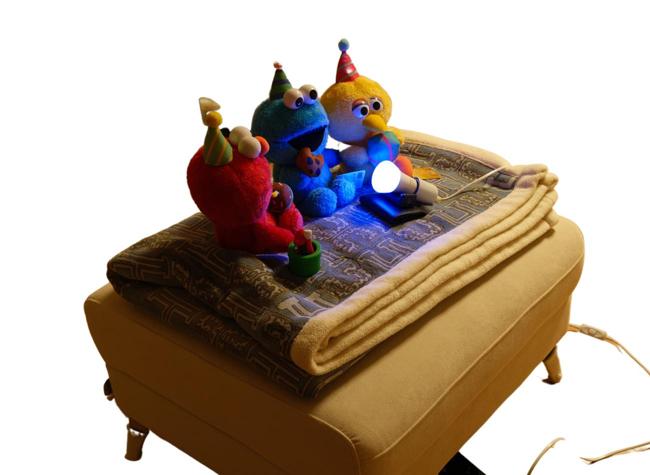}   & \includegraphics[valign=m,width=.195\textwidth]{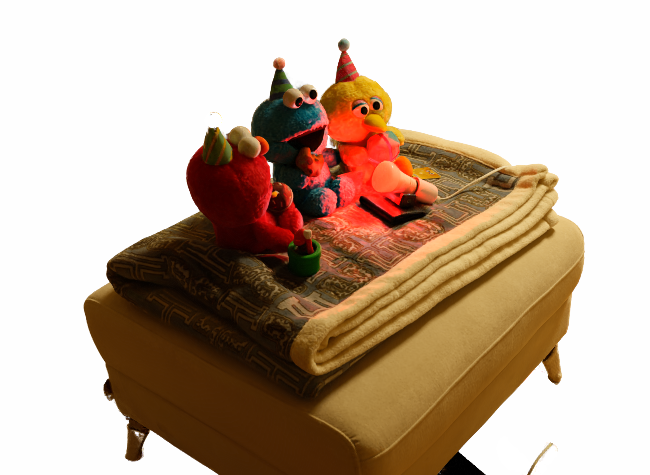}   & \includegraphics[valign=m,width=.195\textwidth]{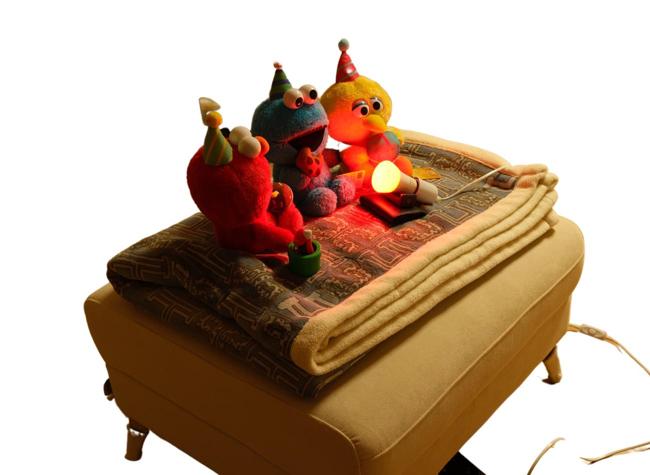}   \\
    \end{tabular}
    \caption{Qualitative results comparison with ground truth obtained using commercial smart bulbs.}
    \label{fig:real_em_gt}
\end{figure*}

\begin{figure*}
    \centering
    \footnotesize
    \renewcommand{\arraystretch}{0} 
    \begin{tabular}{@{} c@{\hspace{1pt}} *{3}{c@{\hspace{3.5pt}}}:*{3}{c@{\hspace{3.5pt}}} @{}}
                                                       & \multicolumn{1}{c}{TensoIR}                                                                               & \multicolumn{1}{c}{NeILF++}                                                                                    & \multicolumn{1}{c}{\modelname}                                                                    & \multicolumn{1}{c}{TensoIR}                                                                                & \multicolumn{1}{c}{NeILF++}                                                                                  & \multicolumn{1}{c}{\modelname}                                                                  \\
        \rotatebox[origin=c]{90}{\makecell{Emission}}  & \includegraphics[valign=m,width=.16\textwidth]{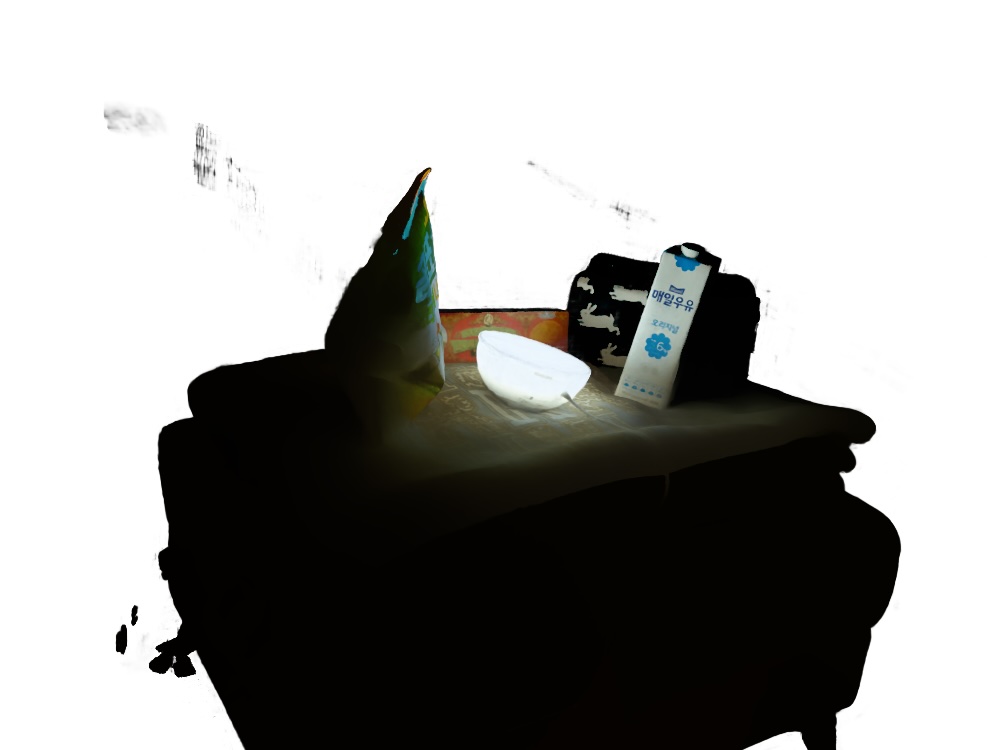}     & \includegraphics[valign=m,width=.16\textwidth]{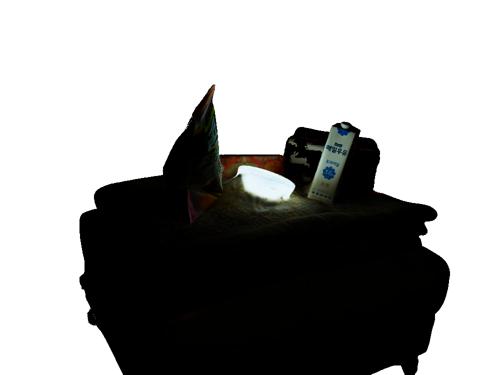}   & \includegraphics[valign=m,width=.16\textwidth]{figures/appendix/Real/snacks/emission.png}         & \includegraphics[valign=m,width=.16\textwidth]{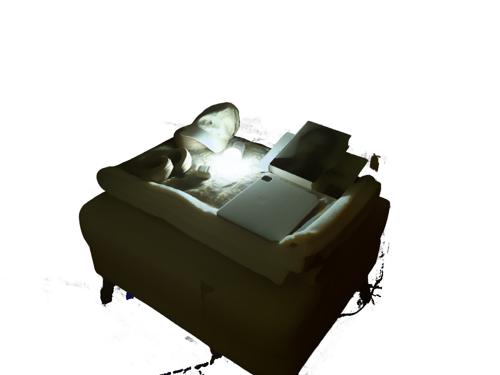} & \includegraphics[valign=m,width=.16\textwidth]{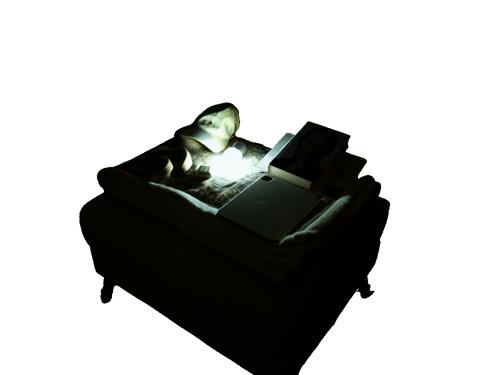}   & \includegraphics[valign=m,width=.16\textwidth]{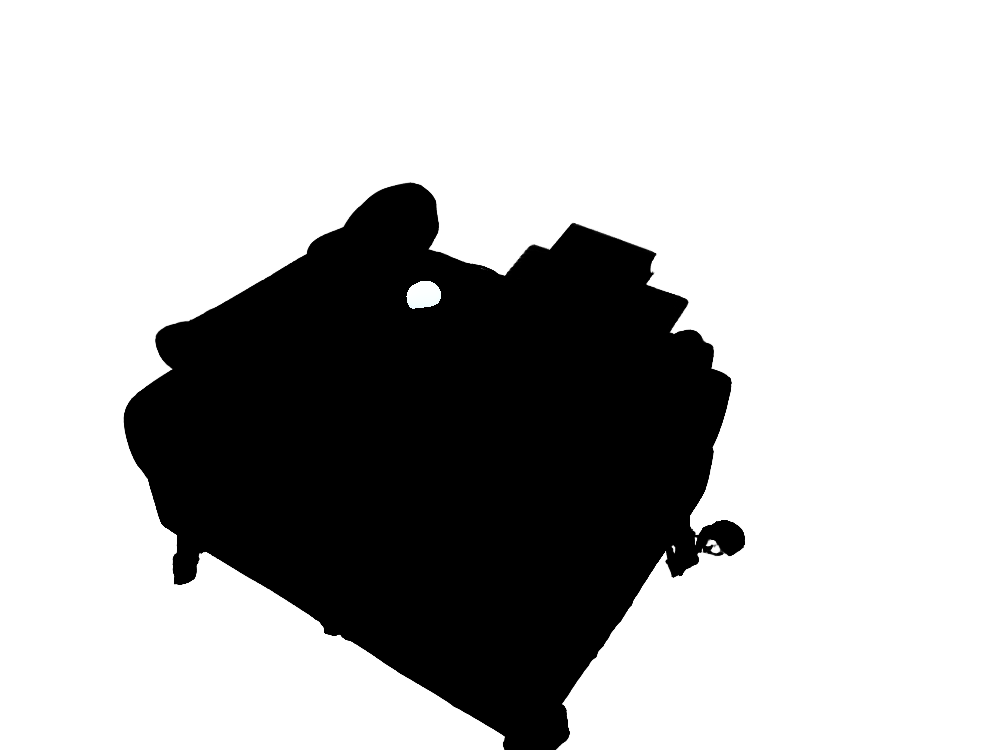}         \\
        \rotatebox[origin=c]{90}{\makecell{Albedo}}    & \includegraphics[valign=m,width=.16\textwidth]{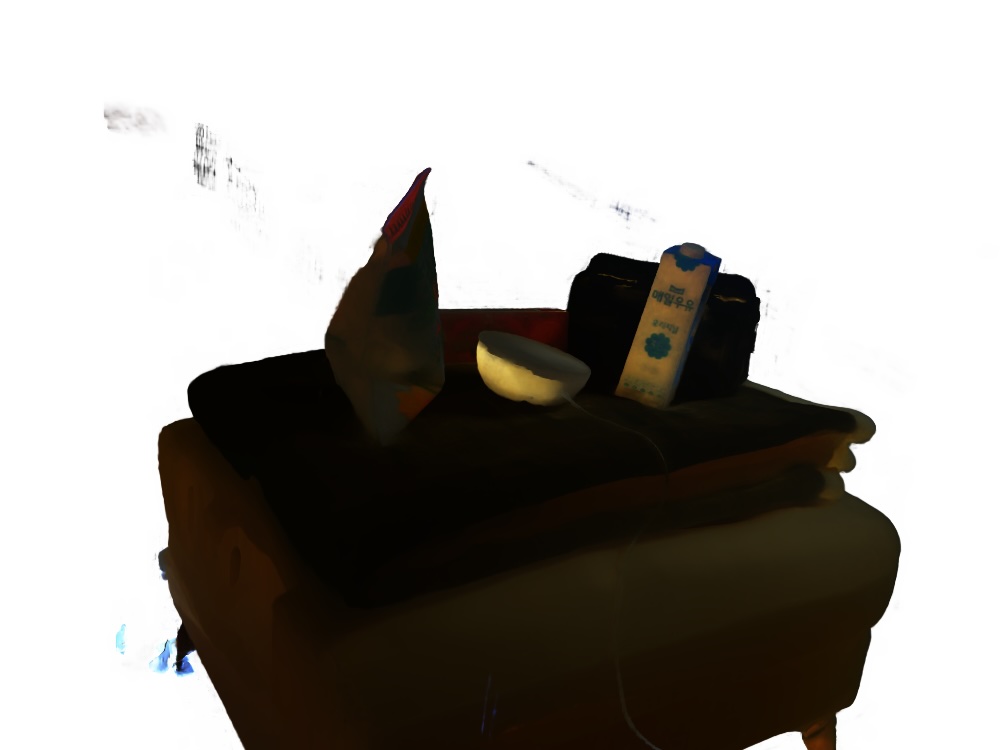}           & \includegraphics[valign=m,width=.16\textwidth]{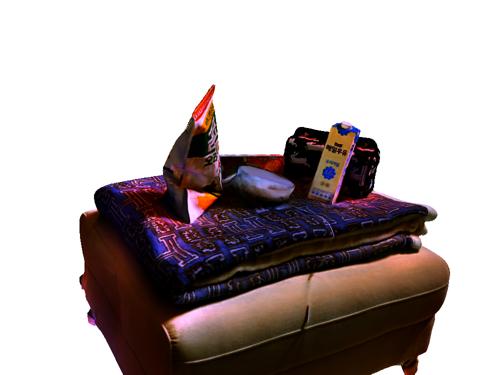} & \includegraphics[valign=m,width=.16\textwidth]{figures/appendix/Real/snacks/basecolor_resize.jpg} & \includegraphics[valign=m,width=.16\textwidth]{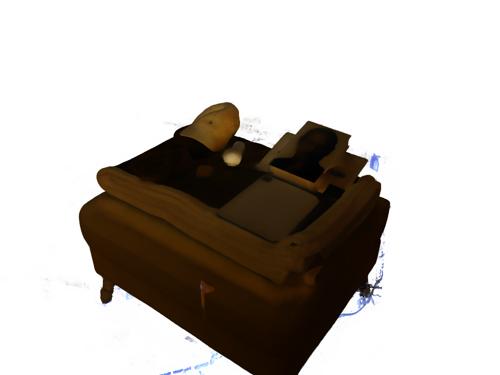}       & \includegraphics[valign=m,width=.16\textwidth]{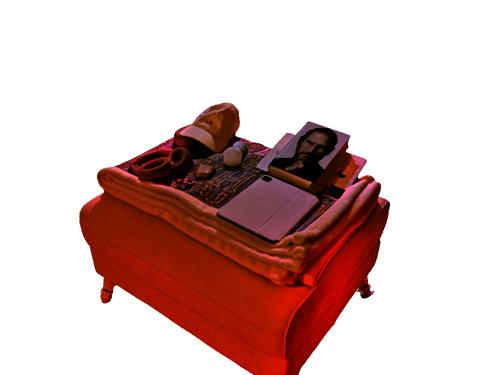} & \includegraphics[valign=m,width=.16\textwidth]{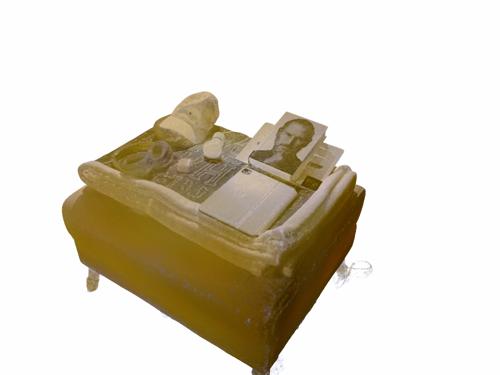} \\
        \rotatebox[origin=c]{90}{\makecell{Roughness}} & \includegraphics[valign=m,width=.16\textwidth]{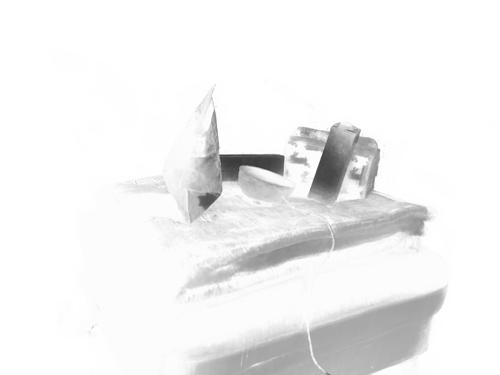} & \includegraphics[valign=m,width=.16\textwidth]{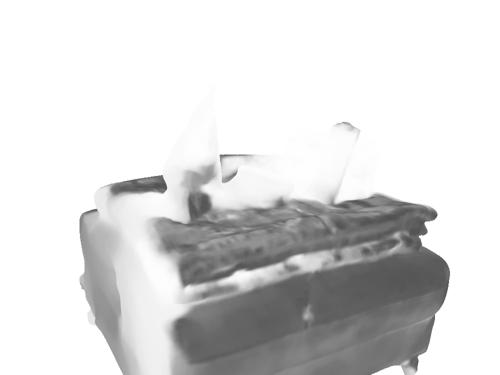}  & \includegraphics[valign=m,width=.16\textwidth]{figures/appendix/Real/snacks/roughness_resize.jpg} & \includegraphics[valign=m,width=.16\textwidth]{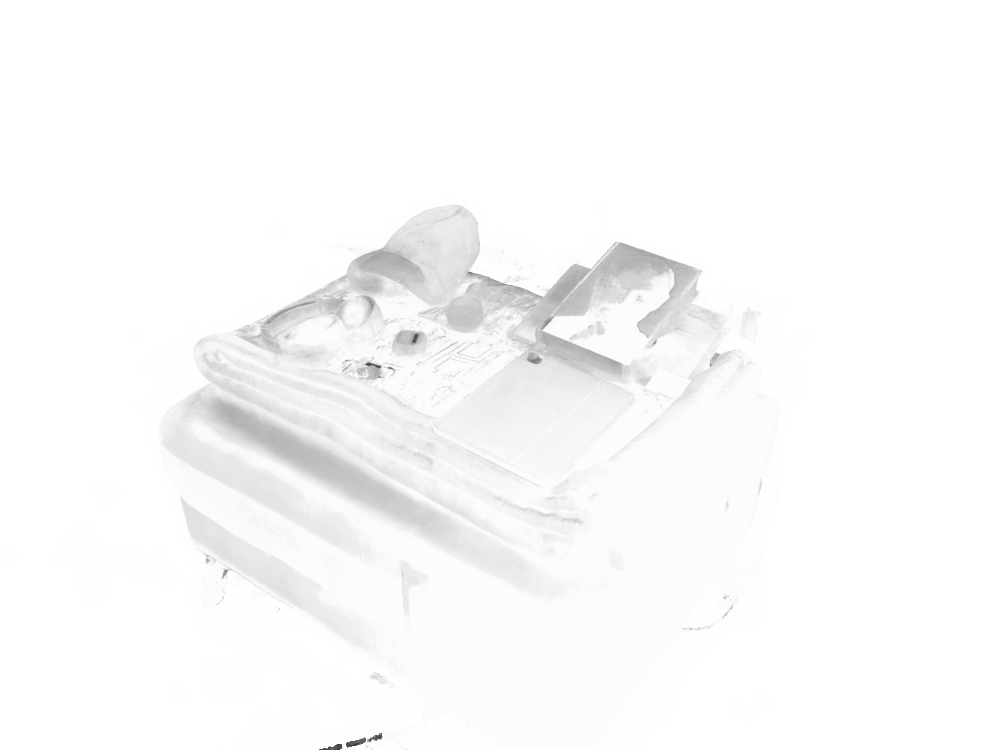}           & \includegraphics[valign=m,width=.16\textwidth]{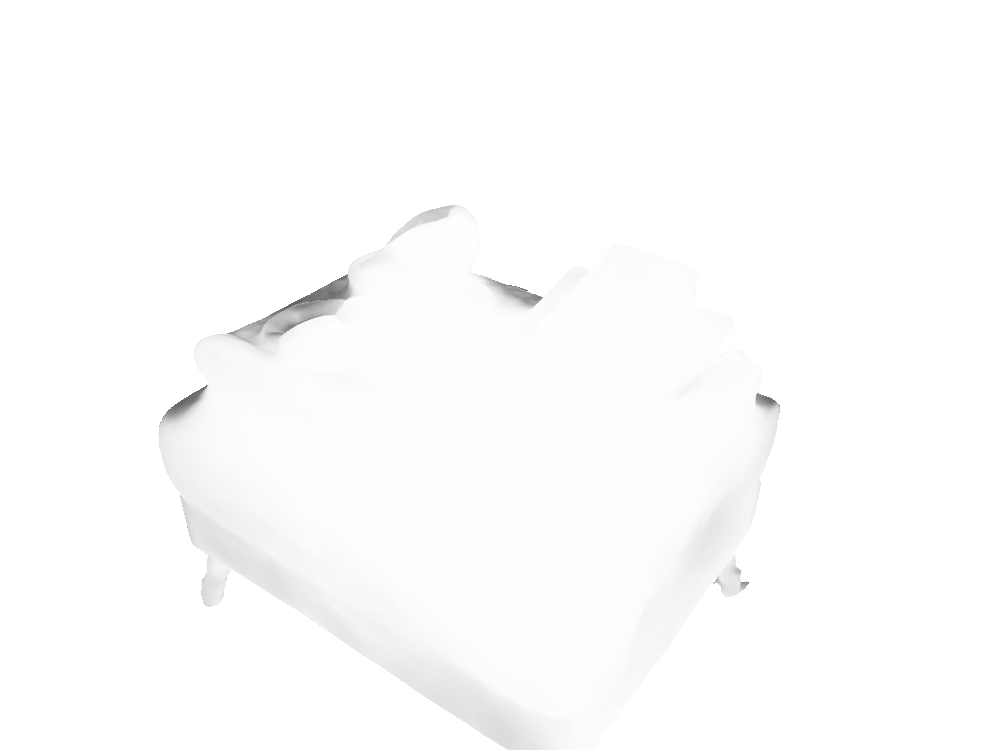}         & \includegraphics[valign=m,width=.16\textwidth]{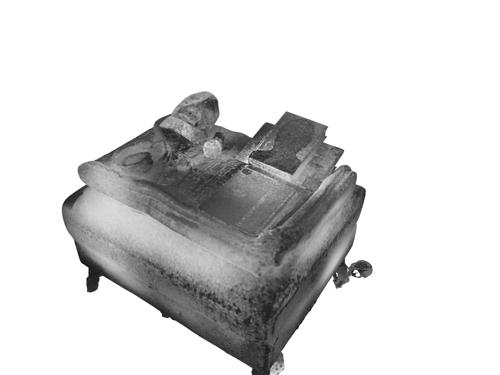} \\
    \end{tabular}
    \caption{
        Comparison of identified emissive sources and decomposed BRDF.
    }
    \label{fig:real_snack_jobs_comparison}
\end{figure*}

\begin{figure*}
    \centering
    \footnotesize
    \renewcommand{\arraystretch}{0} 
    \begin{tabular}{@{} c@{\hspace{1pt}} *{3}{c@{\hspace{3.5pt}}}:*{3}{c@{\hspace{3.5pt}}} @{}}
                                                       & \multicolumn{1}{c}{TensoIR}                                                                                    & \multicolumn{1}{c}{NeILF++}                                                                                      & \multicolumn{1}{c}{\modelname}                                                                      & \multicolumn{1}{c}{TensoIR}                                                                                 & \multicolumn{1}{c}{NeILF++}                                                                                   & \multicolumn{1}{c}{\modelname}                                                                   \\
        \rotatebox[origin=c]{90}{\makecell{Emission}}  & \includegraphics[valign=m,width=.16\textwidth]{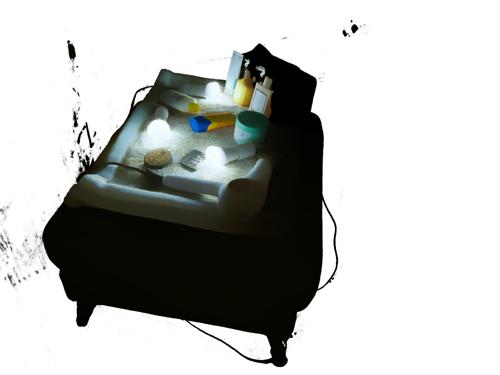} & \includegraphics[valign=m,width=.16\textwidth]{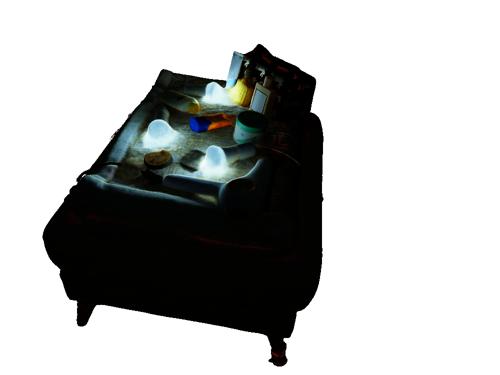}   & \includegraphics[valign=m,width=.16\textwidth]{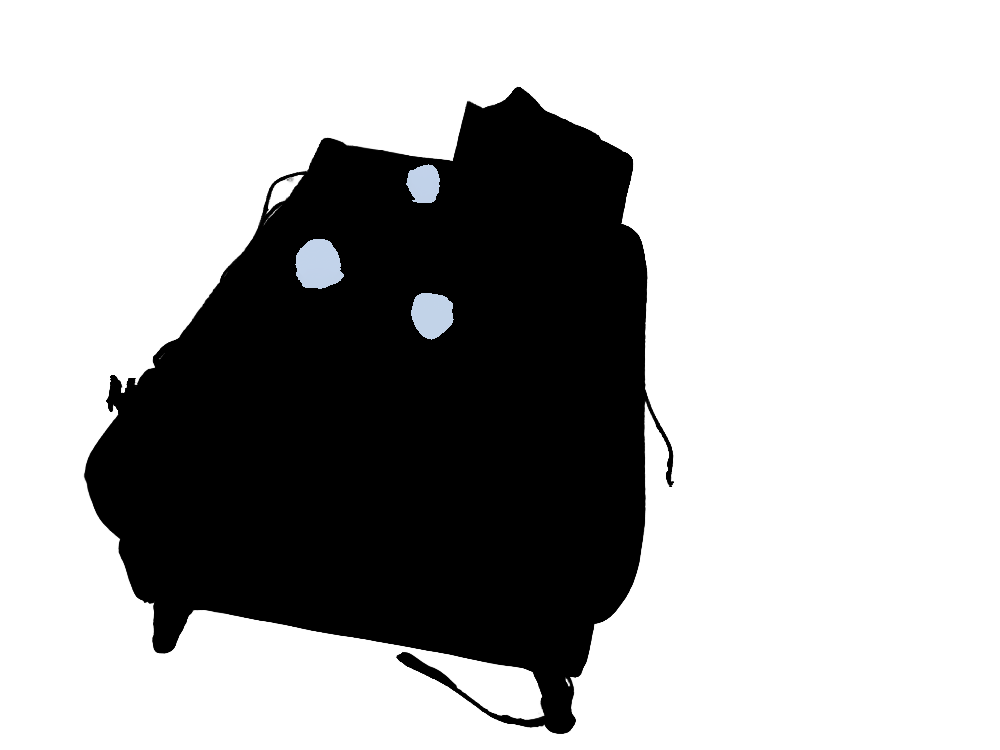}         & \includegraphics[valign=m,width=.16\textwidth]{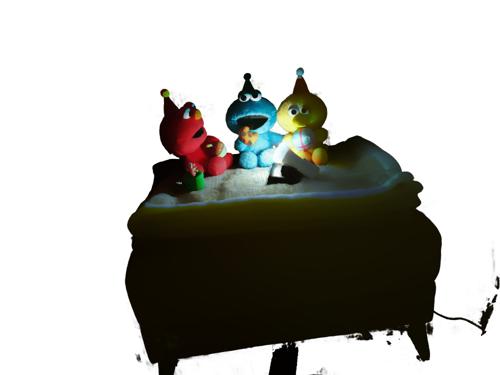} & \includegraphics[valign=m,width=.16\textwidth]{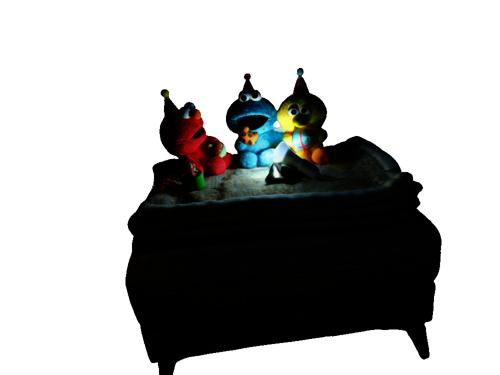}   & \includegraphics[valign=m,width=.16\textwidth]{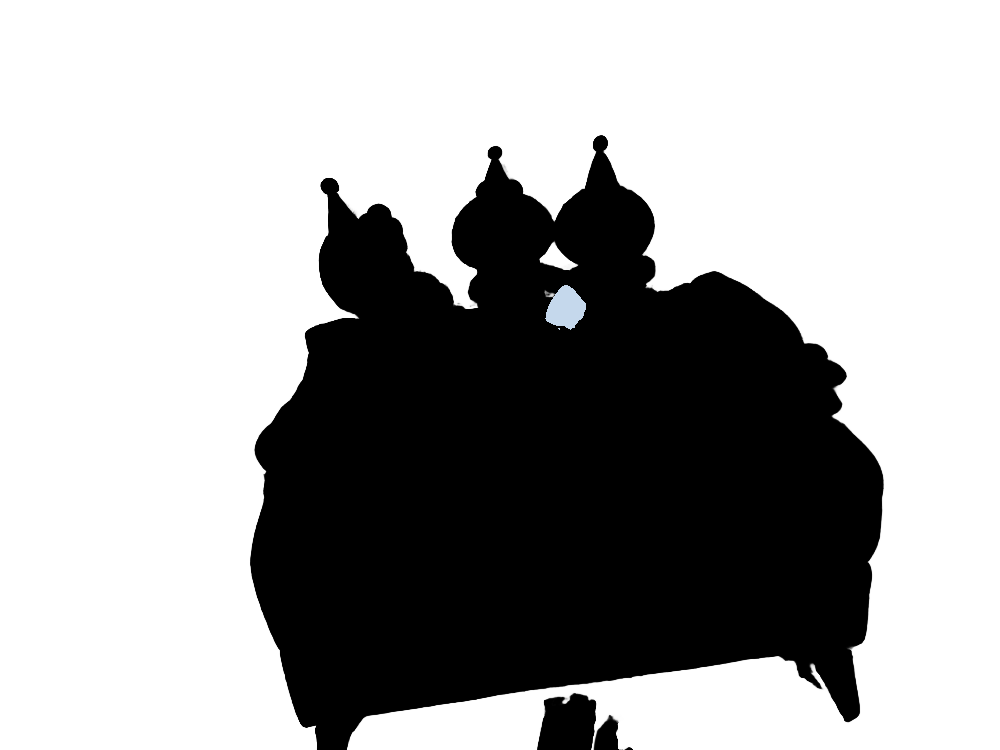}         \\
        \rotatebox[origin=c]{90}{\makecell{Albedo}}    & \includegraphics[valign=m,width=.16\textwidth]{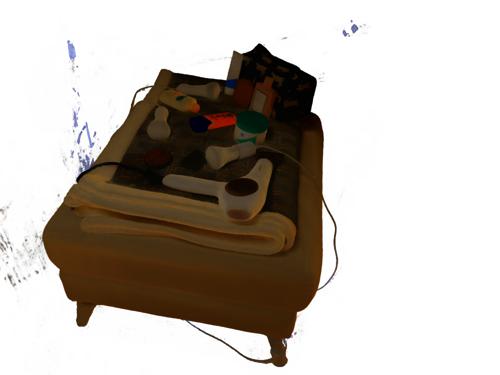}       & \includegraphics[valign=m,width=.16\textwidth]{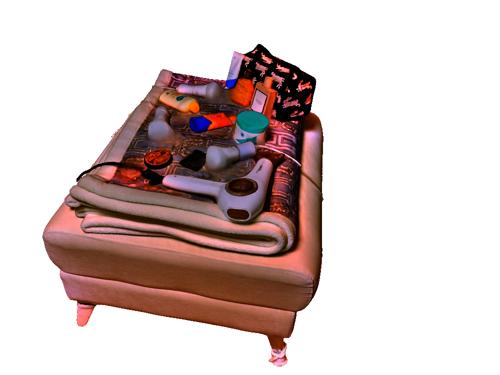} & \includegraphics[valign=m,width=.16\textwidth]{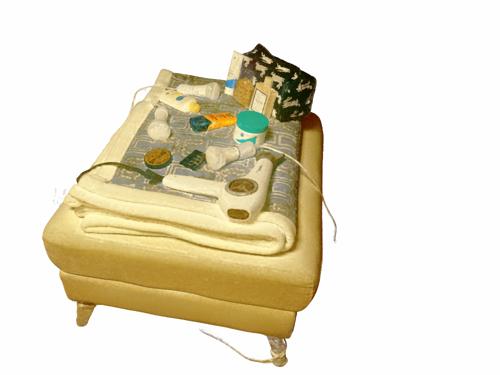} & \includegraphics[valign=m,width=.16\textwidth]{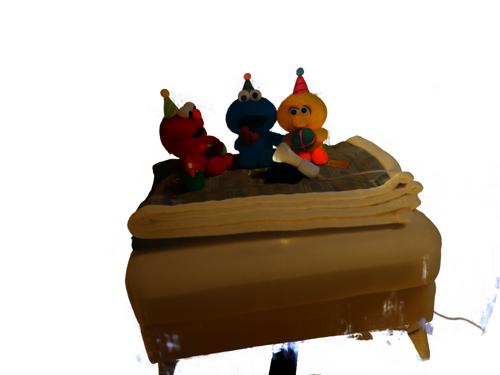}       & \includegraphics[valign=m,width=.16\textwidth]{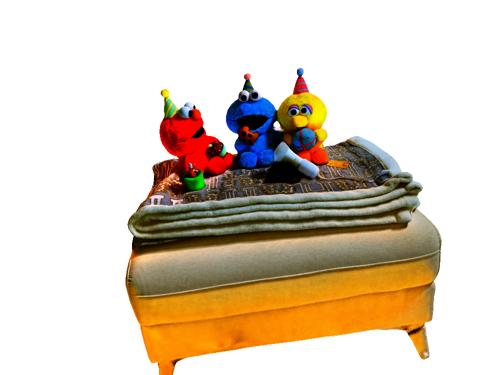} & \includegraphics[valign=m,width=.16\textwidth]{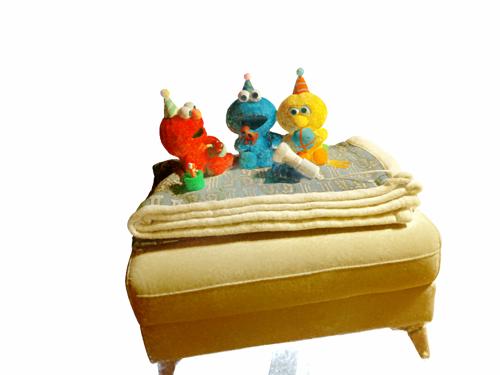} \\
        \rotatebox[origin=c]{90}{\makecell{Roughness}} & \includegraphics[valign=m,width=.16\textwidth]{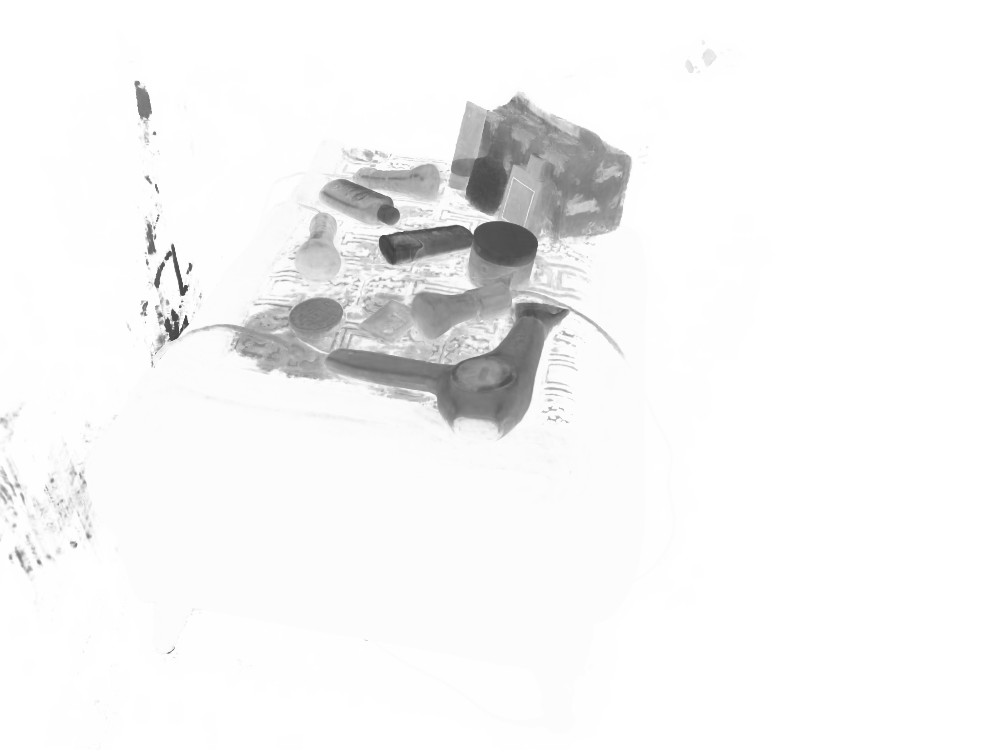}           & \includegraphics[valign=m,width=.16\textwidth]{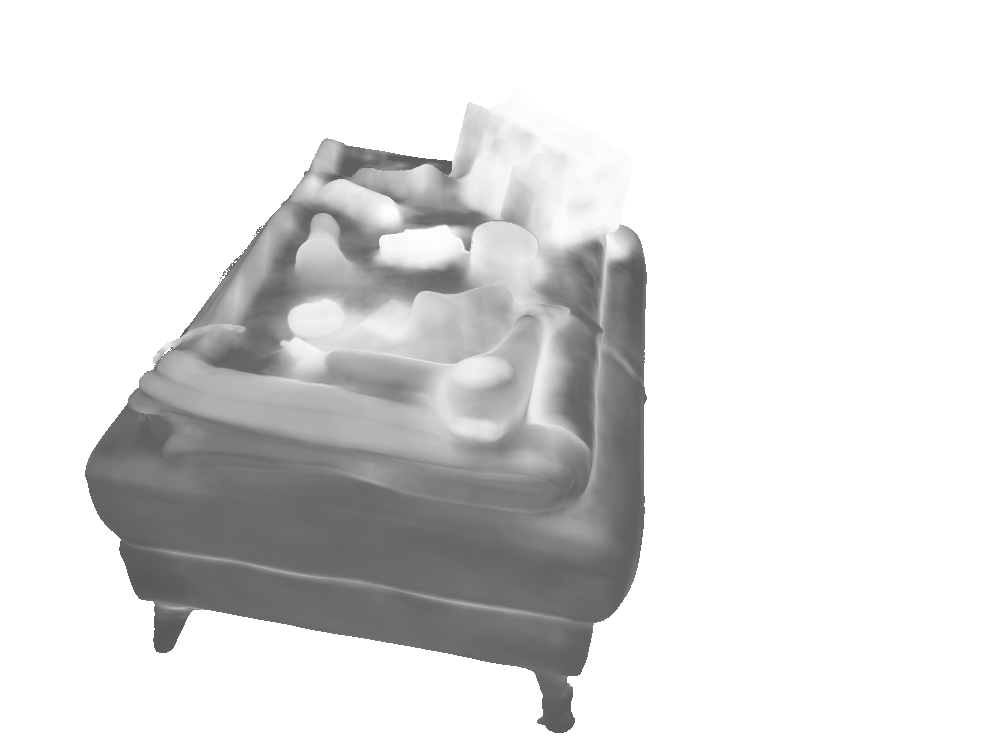}         & \includegraphics[valign=m,width=.16\textwidth]{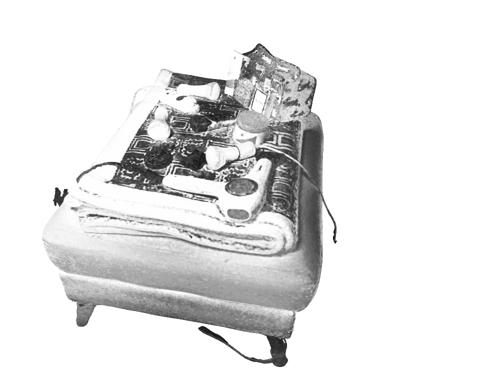} & \includegraphics[valign=m,width=.16\textwidth]{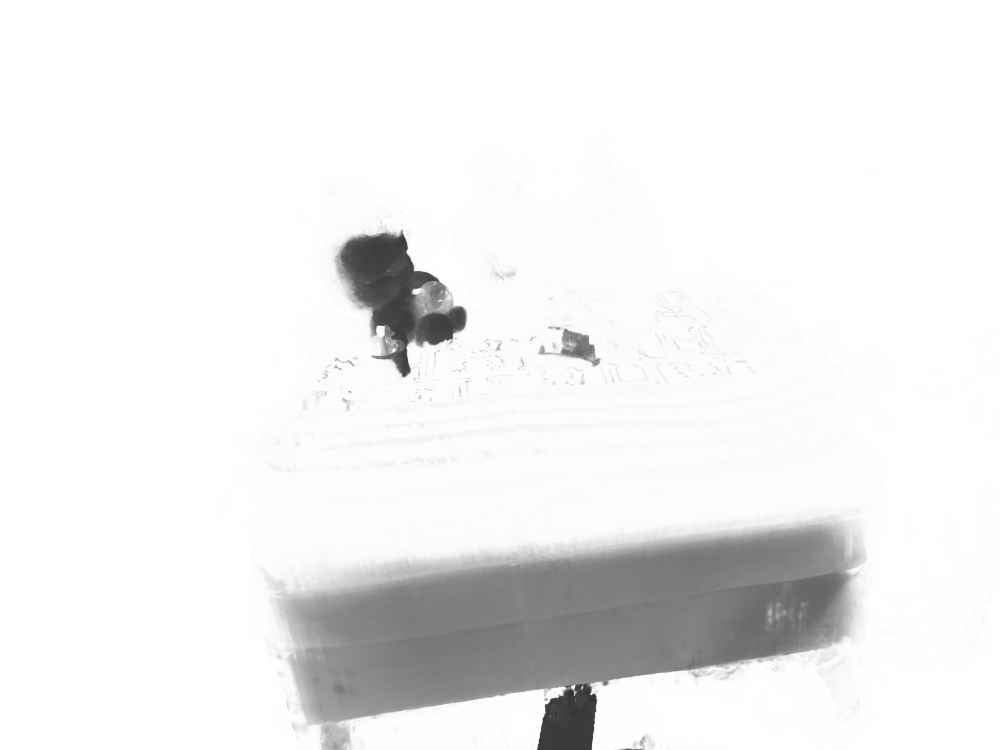}           & \includegraphics[valign=m,width=.16\textwidth]{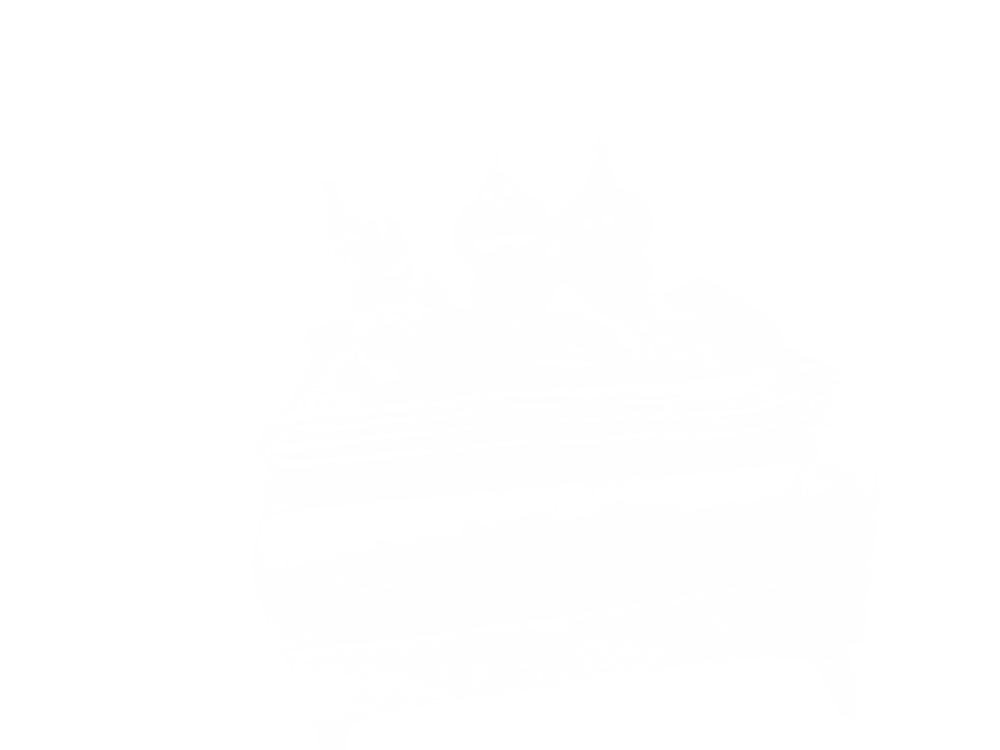}         & \includegraphics[valign=m,width=.16\textwidth]{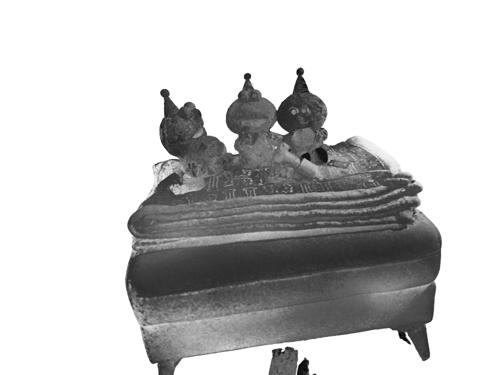} \\
    \end{tabular}
    \caption{
        Comparison of identified emissive sources and decomposed BRDF.
    }
    \label{fig:real_cosmetic_dolls_comparison}
\end{figure*}

\subsection{Baseline Implementation}

In our evaluation, we compared against two leading re-lighting methods, TensoIR~\cite{Jin2023TensoIR} and NeILF++~\cite{zhang2023neilf++}, known for their ability to operate without prior knowledge of scene components
Additionally, we made Twins, a method focused on emissive source reconstruction without relying on inverse rendering techniques.
For an in-depth analysis of scene editing capabilities, we include PaletteNeRF~\cite{kuang2023palettenerf}, which achieves scene modification through re-colorization, and NeRF-W~\cite{nerfw2020}, which adjusts scene illumination by interpolating between learned latent vectors.
For surface reconstruction evaluations on the DTU dataset, we selected state-of-the-art methods such as Voxurf~\cite{wu2022voxurf} and NeuS~\cite{wang2021neus}, alongside Neural-PBIR~\cite{sun2023neuralpbir}, which offers a joint reconstruction of surfaces, materials, and environment maps.
We utilized the official implementations provided by the authors for all baselines, with the exception of NeRF-W.
We used the official implementation codes provided by the authors for all baseline methods except for Twins and NeRF-W.
Twins employs a dual-model strategy for 'light-on' and 'light-off' conditions, using radiance differences for emissive source identification and scene illumination editing.
NeRF-W leverages two latent embeddings for similar purposes, focusing on intensity adjustments.
Both Twins and NeRF-W are based on the Voxurf architecture to ensure a fair comparison with \modelname.
For NeILF++, we omitted the use of prior scene information to align with methods that do not use geometry hints like object meshes or oriented point clouds.
Neural-PBIR was excluded from emissive source reconstruction experiments as the code is not publicly available yet.
Baseline performance data on the DTU dataset are borrowed directly from the Voxurf, NeuS, and Neural-PBIR papers.

\subsection{Real Scene}


We showcase the effectiveness of \modelname in identifying emissive sources in real-world scenes.
Camera poses are estimated using COLMAP~\cite{Schonberger_2016_CVPR}.
We use commercial smart light bulbs from Philips, which offer control over light colors.
Since precise control over the color of the smart bulbs is infeasible, we provide qualitative results for emissive source identification and scene editing in real scenes.
Fig.~\ref{fig:real_decomp} presents the decomposed scene components, such as normal, base color, roughness, metallic, and the environment map.
In Fig.~\ref{fig:real_em}, our method successfully identifies emissive sources, enabling scene illumination adjustments.
Fig.~\ref{fig:real_em_gt} presents qualitative results for comparison with ground truth data.
Although our model successfully identifies emissive sources, it encounters difficulties with complex reflections inside light bulbs, as indicated by the bright spots at the bulb centers in the ground truth edit images.
Despite these challenges, \modelname stands out as the first NeRF-based inverse rendering method to address the reconstruction of emissive sources, enabling scene illumination modifications through the identification of light sources within a scene.

\subsection{Reconstructed Scene Components}
\begin{figure*}[t]
    \centering
    \scriptsize
    \renewcommand{\arraystretch}{3} 
\begin{tabular}{@{} *{7}{c@{\hspace{2pt}}} @{}}
        \multicolumn{1}{c}{Image}                                                                          & \multicolumn{1}{c}{Emissoin}                                                                                 & \multicolumn{1}{c}{Normal}                                                                                  & \multicolumn{1}{c}{Base Color}                                                                                 & \multicolumn{1}{c}{Roughness}                                                                                  & \multicolumn{1}{c}{Metallic}                                                                                 \\
        \includegraphics[valign=m,width=.16\textwidth]{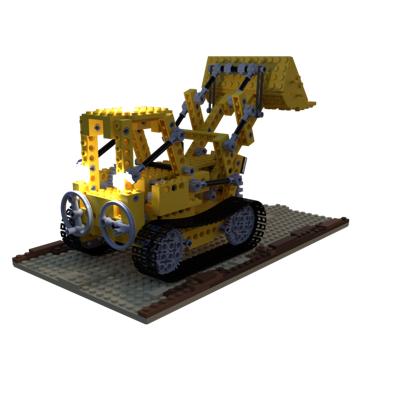}      & \includegraphics[valign=m,width=.16\textwidth]{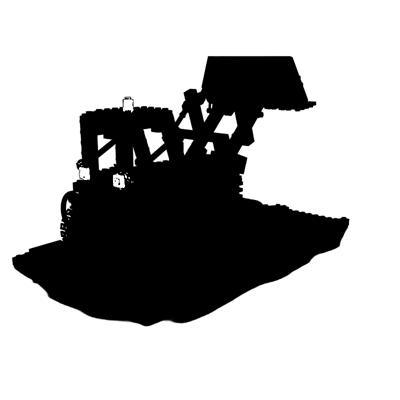}      & \includegraphics[valign=m,width=.16\textwidth]{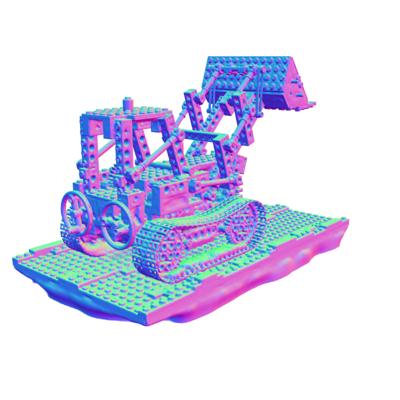}      & \includegraphics[valign=m,width=.16\textwidth]{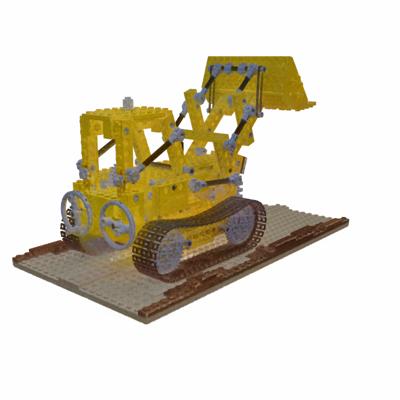}      & \includegraphics[valign=m,width=.16\textwidth]{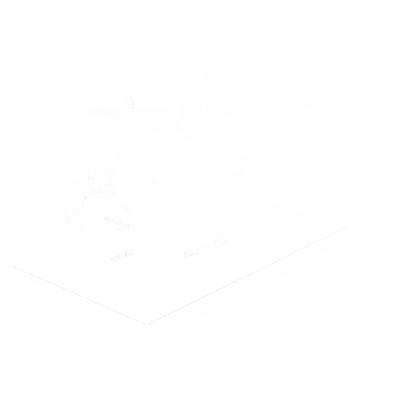}      & \includegraphics[valign=m,width=.16\textwidth]{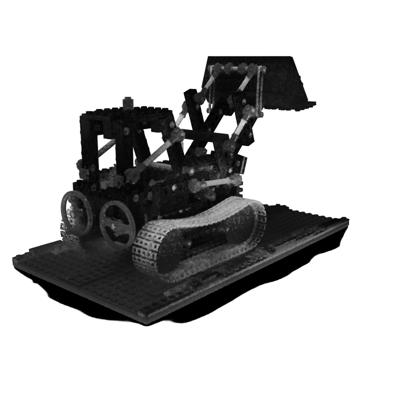}      \\
\end{tabular}
    \caption{Decomposed scene components on scenes with white-colored emissive sources.}
    \label{fig:brdf_w}
\end{figure*}

\begin{figure*}[t]
    \centering
    \scriptsize
    \renewcommand{\arraystretch}{3} 
    \begin{tabular}{@{} *{7}{c@{\hspace{2pt}}} @{}}
        \multicolumn{1}{c}{Image}                                                                          & \multicolumn{1}{c}{Emissoin}                                                                                 & \multicolumn{1}{c}{Normal}                                                                                  & \multicolumn{1}{c}{Base Color}                                                                                 & \multicolumn{1}{c}{Roughness}                                                                                  & \multicolumn{1}{c}{Metallic}                                                                                 \\
        \includegraphics[valign=m,width=.16\textwidth]{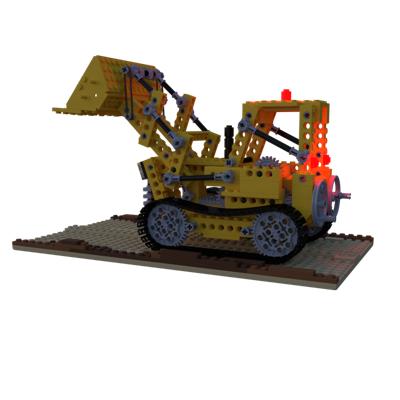}      & \includegraphics[valign=m,width=.16\textwidth]{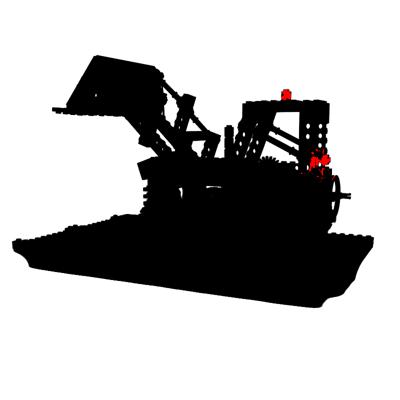}      & \includegraphics[valign=m,width=.16\textwidth]{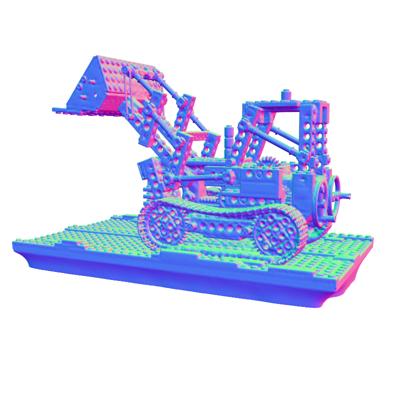}      & \includegraphics[valign=m,width=.16\textwidth]{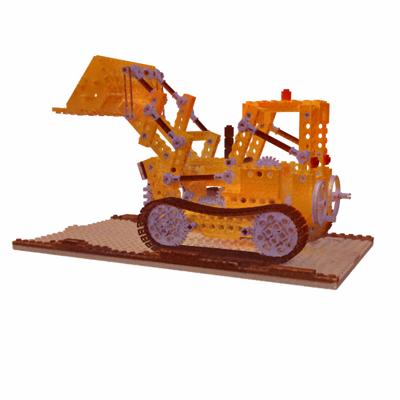}      & \includegraphics[valign=m,width=.16\textwidth]{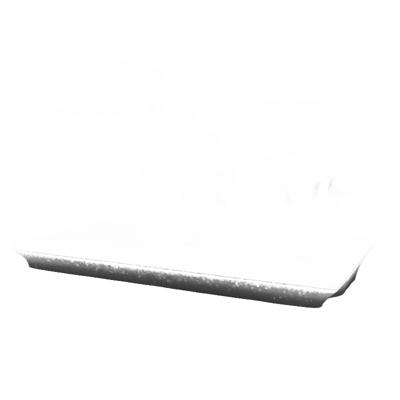}      & \includegraphics[valign=m,width=.16\textwidth]{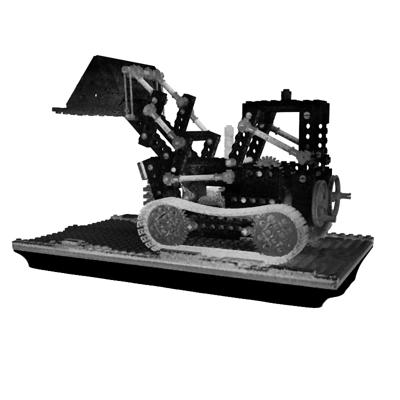}      \\
\end{tabular}
    \caption{Decomposed scene components on scenes with vivid-colored emissive sources.}
    \label{fig:brdf_c}
\end{figure*}

\begin{figure*}
    \centering
    \footnotesize
    \renewcommand{\arraystretch}{0} 
    \begin{tabular}{@{} c@{\hspace{1pt}} *{3}{c@{\hspace{3.5pt}}}:*{3}{c@{\hspace{3.5pt}}} @{}}
                                                       & \multicolumn{1}{c}{TensoIR}                                                                                  & \multicolumn{1}{c}{NeILF++}                                                                                    & \multicolumn{1}{c}{\modelname}                                                                            & \multicolumn{1}{c}{TensoIR}                                                                                  & \multicolumn{1}{c}{NeILF++}                                                                                    & \multicolumn{1}{c}{\modelname}                                                                            \\
        \rotatebox[origin=c]{90}{\makecell{Emission}}  & \includegraphics[valign=m,width=.15\textwidth]{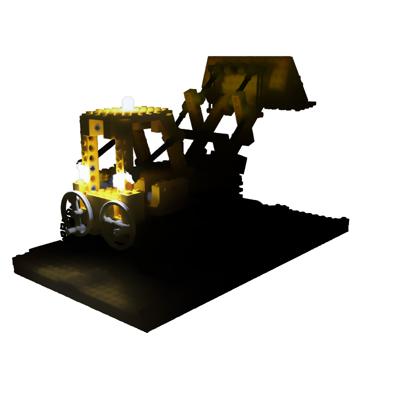} & \includegraphics[valign=m,width=.15\textwidth]{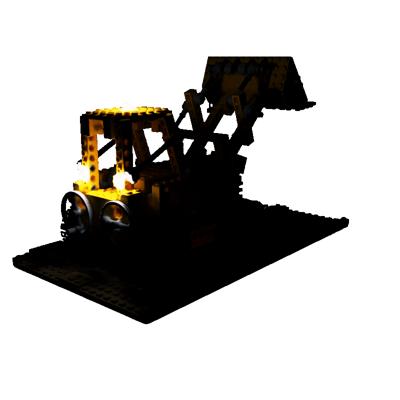}   & \includegraphics[valign=m,width=.15\textwidth]{figures/appendix/BRDF_resize/lego_w/emission_masked.jpg}   & \includegraphics[valign=m,width=.15\textwidth]{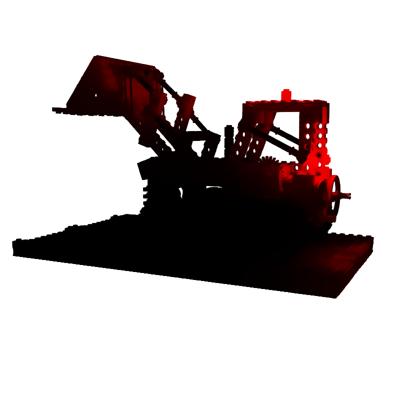} & \includegraphics[valign=m,width=.15\textwidth]{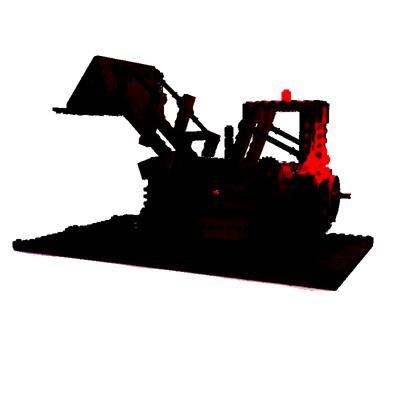}   & \includegraphics[valign=m,width=.15\textwidth]{figures/appendix/BRDF_resize/lego_c/emission_masked.jpg}   \\
\end{tabular}
    \caption{
        Comparison of identified emissive sources and decomposed BRDF on the Lego scene. Left: Lego white. Right: Lego vivid.
    }
    \label{fig:lego_comparison}
\end{figure*}

\begin{figure*}
    \centering
    \footnotesize
    \renewcommand{\arraystretch}{0} 
    \begin{tabular}{@{} c@{\hspace{1pt}} *{3}{c@{\hspace{3.5pt}}}:*{3}{c@{\hspace{3.5pt}}} @{}}
                                                       & \multicolumn{1}{c}{TensoIR}                                                                                  & \multicolumn{1}{c}{NeILF++}                                                                                    & \multicolumn{1}{c}{\modelname}                                                                            & \multicolumn{1}{c}{TensoIR}                                                                                  & \multicolumn{1}{c}{NeILF++}                                                                                    & \multicolumn{1}{c}{\modelname}                                                                            \\
        \rotatebox[origin=c]{90}{\makecell{Emission}}  & \includegraphics[valign=m,width=.15\textwidth]{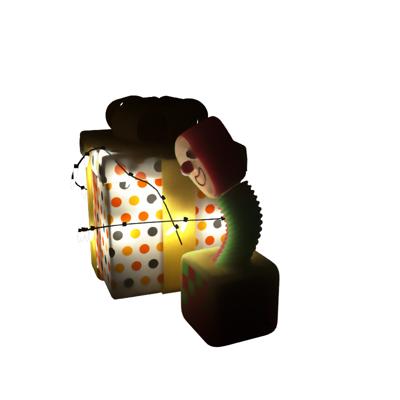} & \includegraphics[valign=m,width=.15\textwidth]{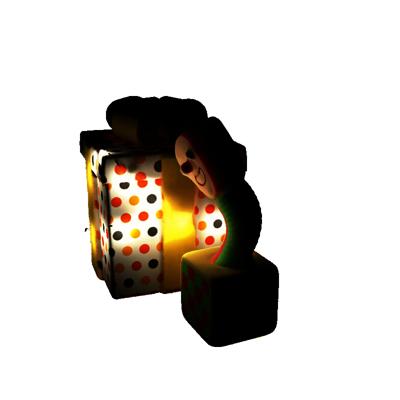}   & \includegraphics[valign=m,width=.15\textwidth]{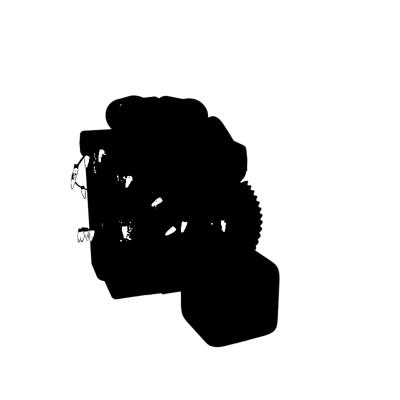}   & \includegraphics[valign=m,width=.15\textwidth]{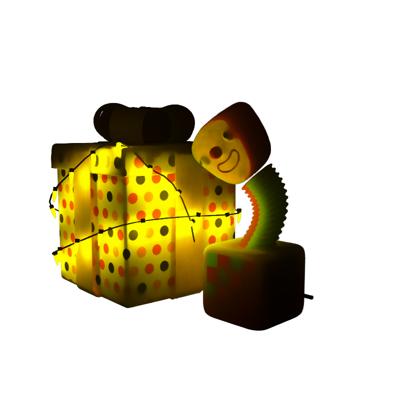} & \includegraphics[valign=m,width=.15\textwidth]{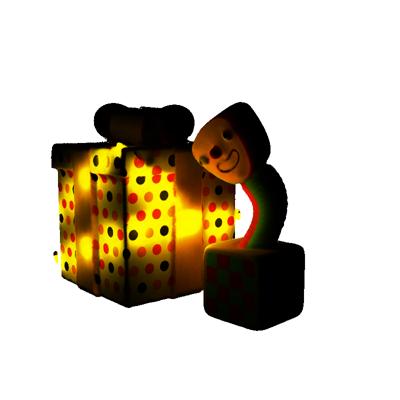}   & \includegraphics[valign=m,width=.15\textwidth]{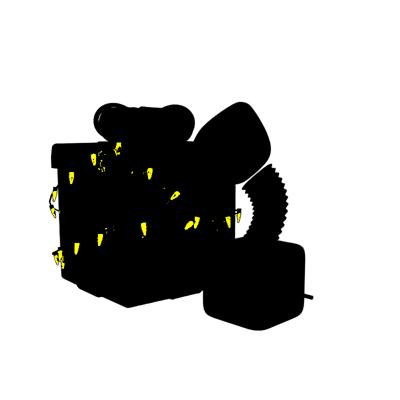}   \\
\end{tabular}
    \caption{
        Comparison of identified emissive sources and decomposed BRDF on the Gift scene. Left: Gift white. Right: Gift vivid.
    }
    \label{fig:gift_comparison}
\end{figure*}

\begin{figure*}
    \centering
    \footnotesize
    \renewcommand{\arraystretch}{0} 
    \begin{tabular}{@{} c@{\hspace{1pt}} *{3}{c@{\hspace{3.5pt}}}:*{3}{c@{\hspace{3.5pt}}} @{}}
                                                       & \multicolumn{1}{c}{TensoIR}                                                                                  & \multicolumn{1}{c}{NeILF++}                                                                                    & \multicolumn{1}{c}{\modelname}                                                                            & \multicolumn{1}{c}{TensoIR}                                                                                  & \multicolumn{1}{c}{NeILF++}                                                                                    & \multicolumn{1}{c}{\modelname}                                                                            \\
        \rotatebox[origin=c]{90}{\makecell{Emission}}  & \includegraphics[valign=m,width=.15\textwidth]{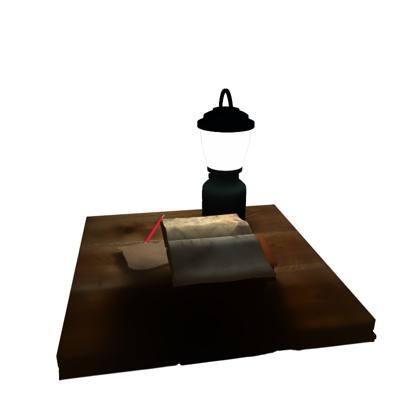} & \includegraphics[valign=m,width=.15\textwidth]{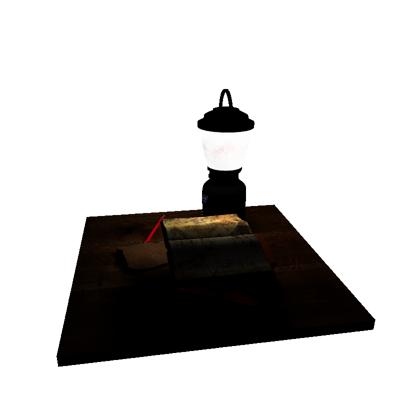}   & \includegraphics[valign=m,width=.15\textwidth]{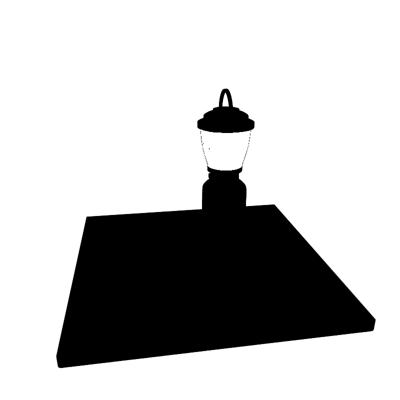}   & \includegraphics[valign=m,width=.15\textwidth]{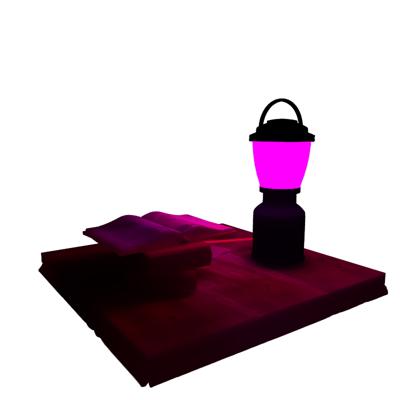} & \includegraphics[valign=m,width=.15\textwidth]{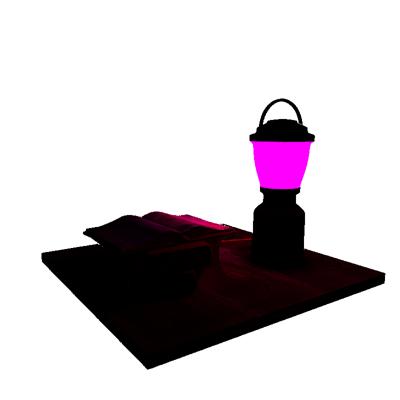}   & \includegraphics[valign=m,width=.15\textwidth]{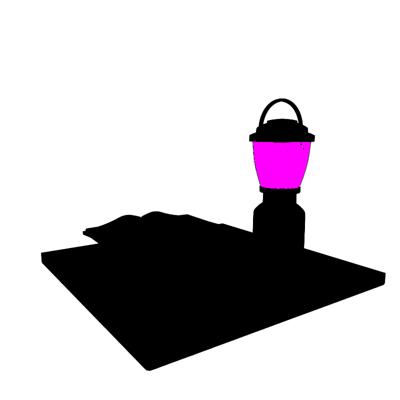}   \\
\end{tabular}
    \caption{
        Comparison of identified emissive sources and decomposed BRDF on the Book scene. Left: Book white. Right: Book vivid.
    }
    \label{fig:book_comparison}
\end{figure*}

\begin{figure*}
    \centering
    \footnotesize
    \renewcommand{\arraystretch}{0} 
    \begin{tabular}{@{} c@{\hspace{1pt}} *{3}{c@{\hspace{3.5pt}}}:*{3}{c@{\hspace{3.5pt}}} @{}}
                                                       & \multicolumn{1}{c}{TensoIR}                                                                                  & \multicolumn{1}{c}{NeILF++}                                                                                    & \multicolumn{1}{c}{\modelname}                                                                            & \multicolumn{1}{c}{TensoIR}                                                                                  & \multicolumn{1}{c}{NeILF++}                                                                                    & \multicolumn{1}{c}{\modelname}                                                                            \\
        \rotatebox[origin=c]{90}{\makecell{Emission}}  & \includegraphics[valign=m,width=.15\textwidth]{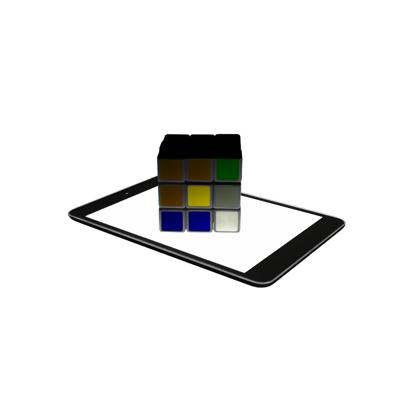} & \includegraphics[valign=m,width=.15\textwidth]{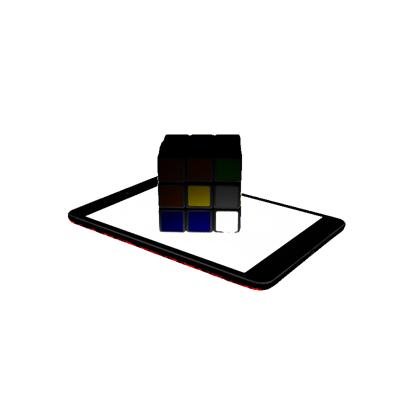}   & \includegraphics[valign=m,width=.15\textwidth]{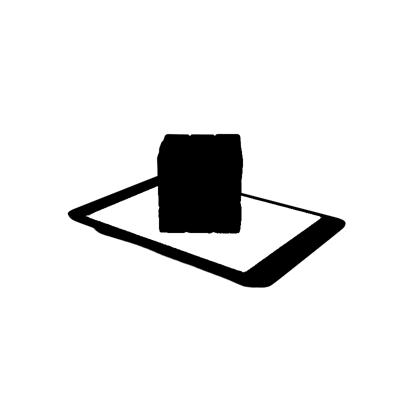}   & \includegraphics[valign=m,width=.15\textwidth]{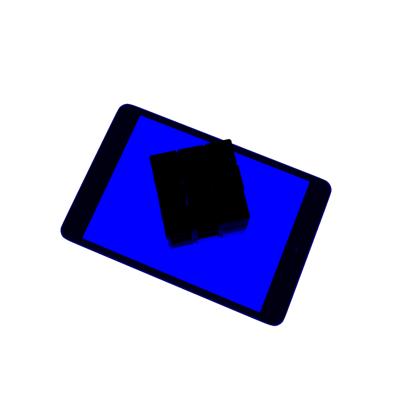} & \includegraphics[valign=m,width=.15\textwidth]{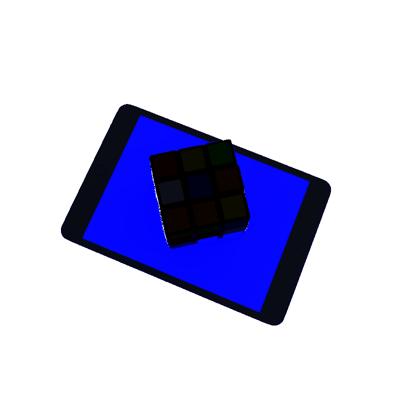}   & \includegraphics[valign=m,width=.15\textwidth]{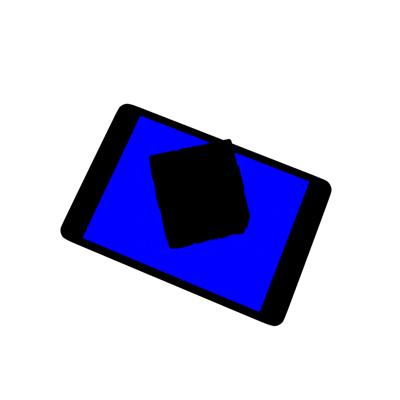}   \\
\end{tabular}
    \caption{
        Comparison of identified emissive sources and decomposed BRDF on the Cube scene. Left: Cube white. Right: Cube vivid.
    }
    \label{fig:cube_comparison}
\end{figure*}

\begin{figure*}
    \centering
    \footnotesize
    \renewcommand{\arraystretch}{0} 
    \begin{tabular}{@{} c@{\hspace{1pt}} *{3}{c@{\hspace{3.5pt}}}:*{3}{c@{\hspace{3.5pt}}} @{}}
                                                       & \multicolumn{1}{c}{TensoIR}                                                                                       & \multicolumn{1}{c}{NeILF++}                                                                                         & \multicolumn{1}{c}{\modelname}                                                                                 & \multicolumn{1}{c}{TensoIR}                                                                                       & \multicolumn{1}{c}{NeILF++}                                                                                         & \multicolumn{1}{c}{\modelname}                                                                                 \\
        \rotatebox[origin=c]{90}{\makecell{Emission}}  & \includegraphics[valign=m,width=.15\textwidth]{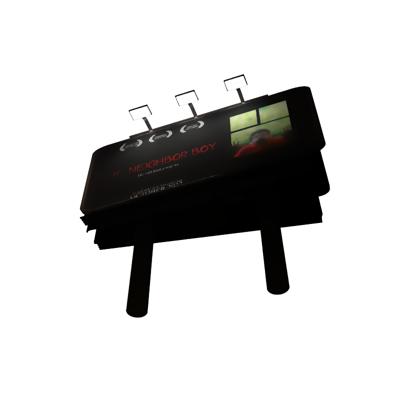} & \includegraphics[valign=m,width=.15\textwidth]{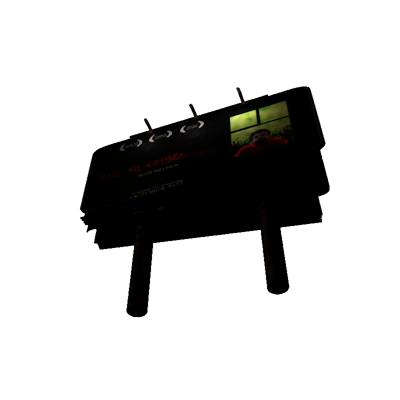}   & \includegraphics[valign=m,width=.15\textwidth]{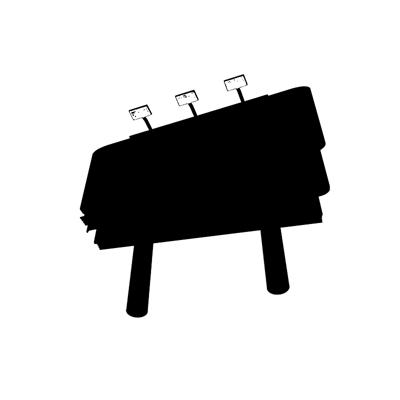}   & \includegraphics[valign=m,width=.15\textwidth]{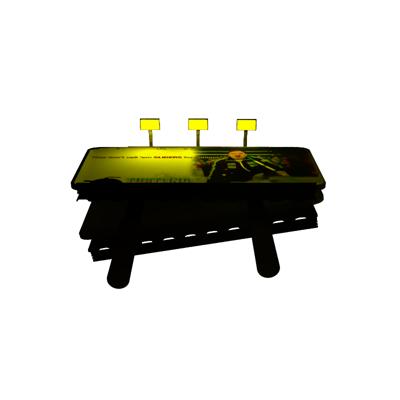} & \includegraphics[valign=m,width=.15\textwidth]{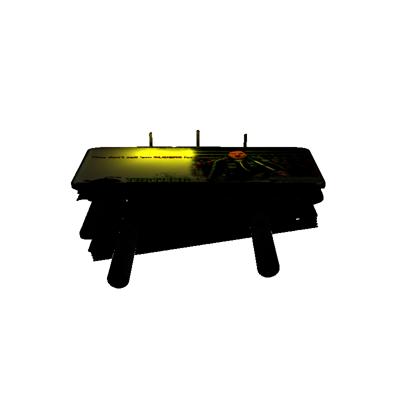}   & \includegraphics[valign=m,width=.15\textwidth]{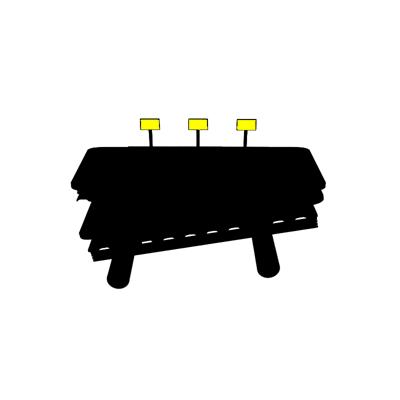}   \\
\end{tabular}
    \caption{
        Comparison of identified emissive sources and decomposed BRDF on the Billboard scene. Left: Billboard white. Right: Billboard vivid.
    }
    \label{fig:billboard_comparison}
\end{figure*}

\begin{figure*}[t]
    \centering
    \scriptsize
    \renewcommand{\arraystretch}{3} 
    \begin{tabular}{@{} *{7}{c@{\hspace{2pt}}} @{}}
        \multicolumn{1}{c}{Image}                                                                & \multicolumn{1}{c}{Normal}                                                                   & \multicolumn{1}{c}{Base Color}                                                                  & \multicolumn{1}{c}{Roughness}                                                                   & \multicolumn{1}{c}{Metallic}                                                                   & \multicolumn{1}{c}{Env. Map}                                                               \\
        \includegraphics[valign=m,width=.15\textwidth]{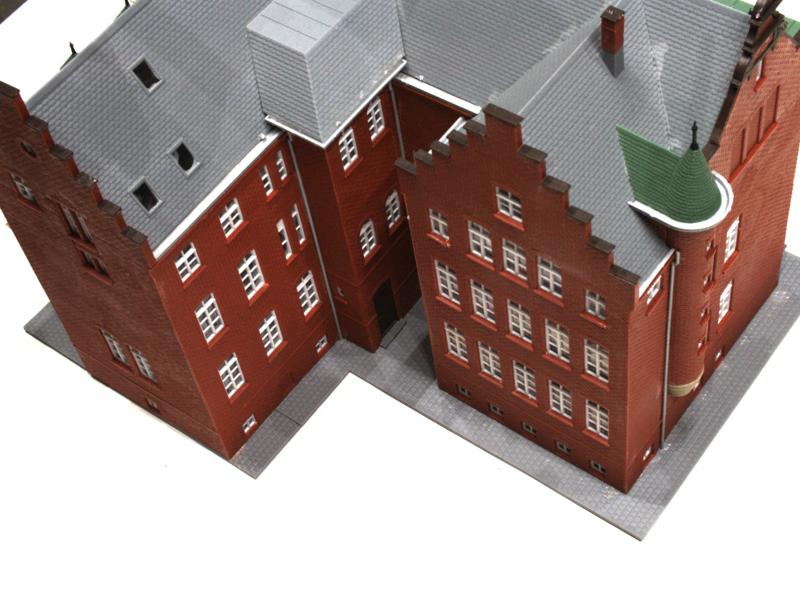} & \includegraphics[valign=m,width=.15\textwidth]{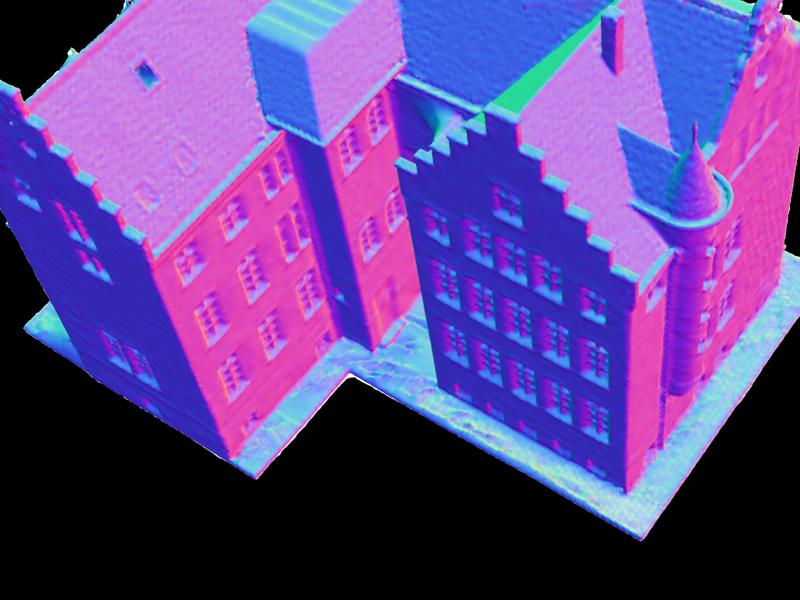} & \includegraphics[valign=m,width=.15\textwidth]{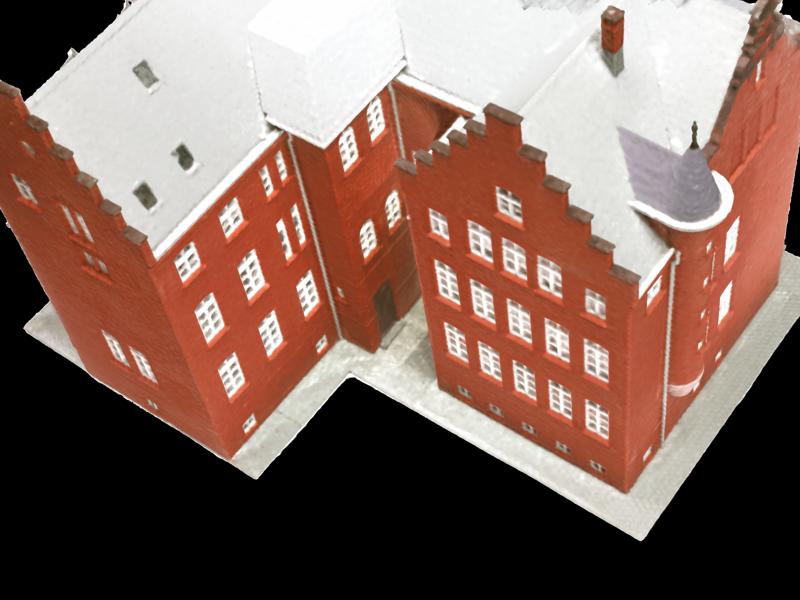} & \includegraphics[valign=m,width=.15\textwidth]{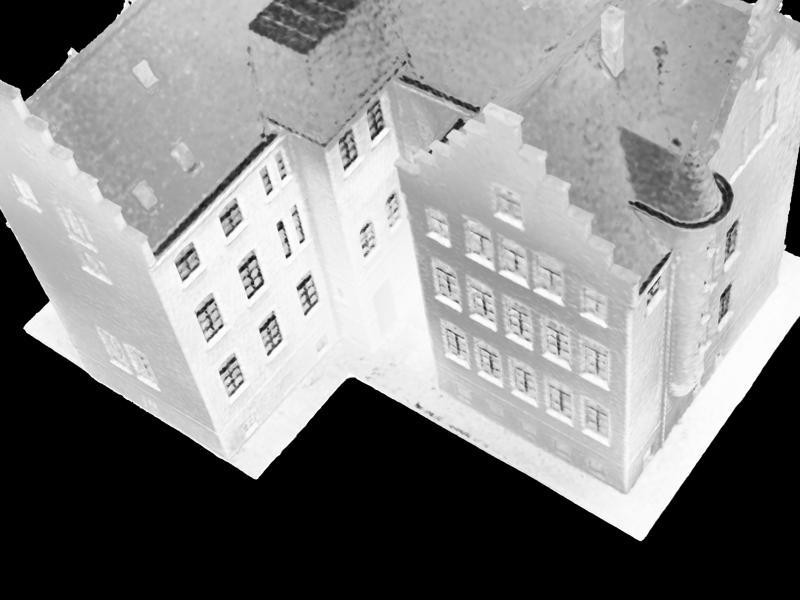} & \includegraphics[valign=m,width=.15\textwidth]{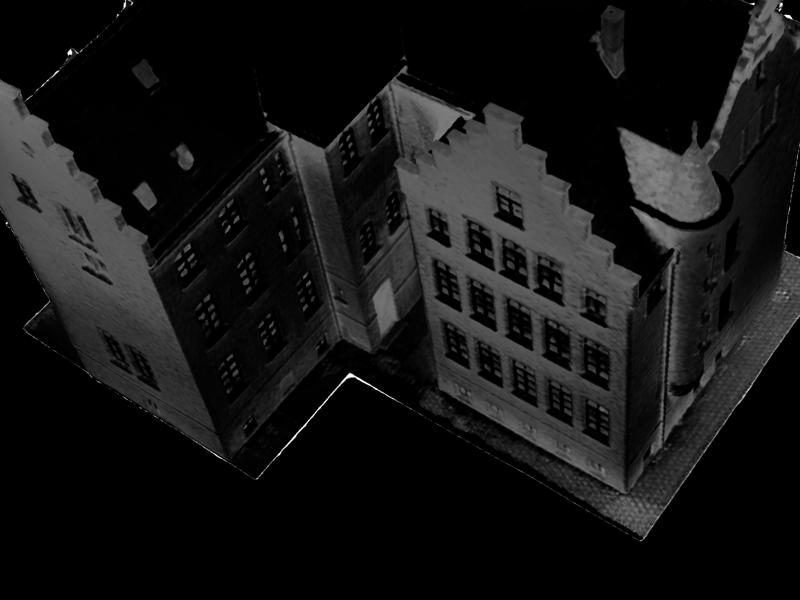} & \includegraphics[valign=m,width=.225\textwidth]{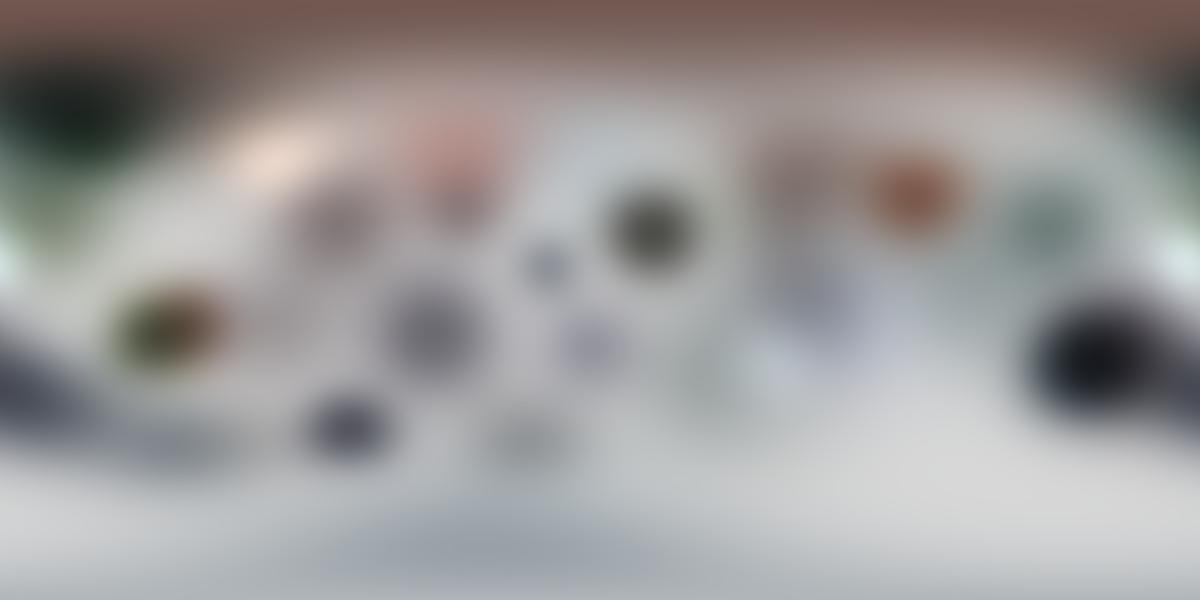} \\
        \includegraphics[valign=m,width=.15\textwidth]{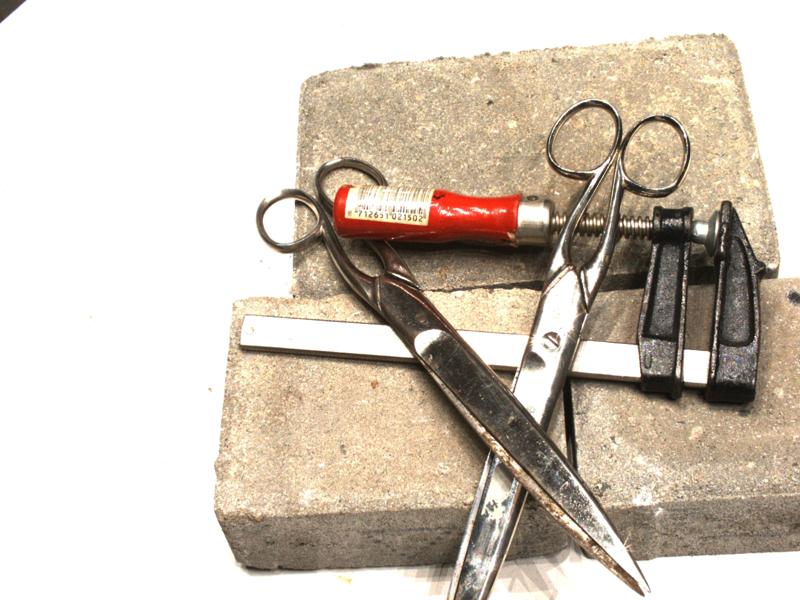} & \includegraphics[valign=m,width=.15\textwidth]{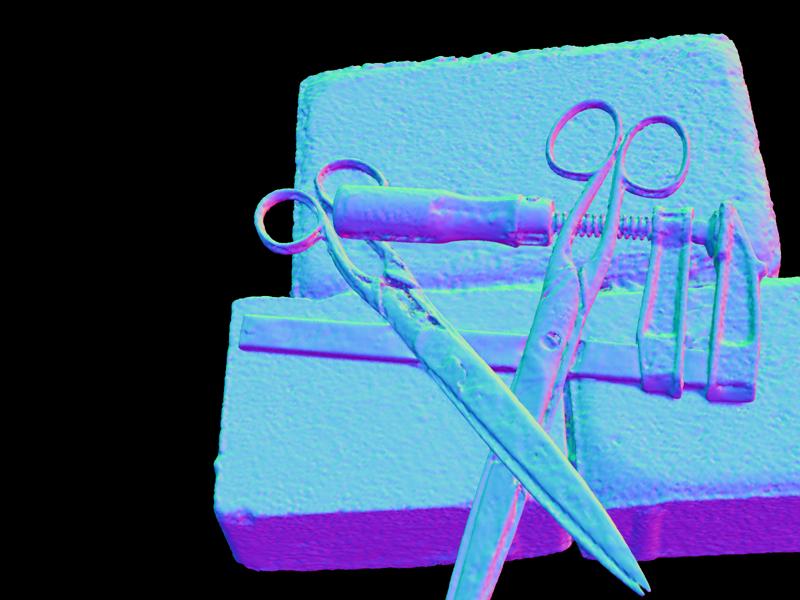}    & \includegraphics[valign=m,width=.15\textwidth]{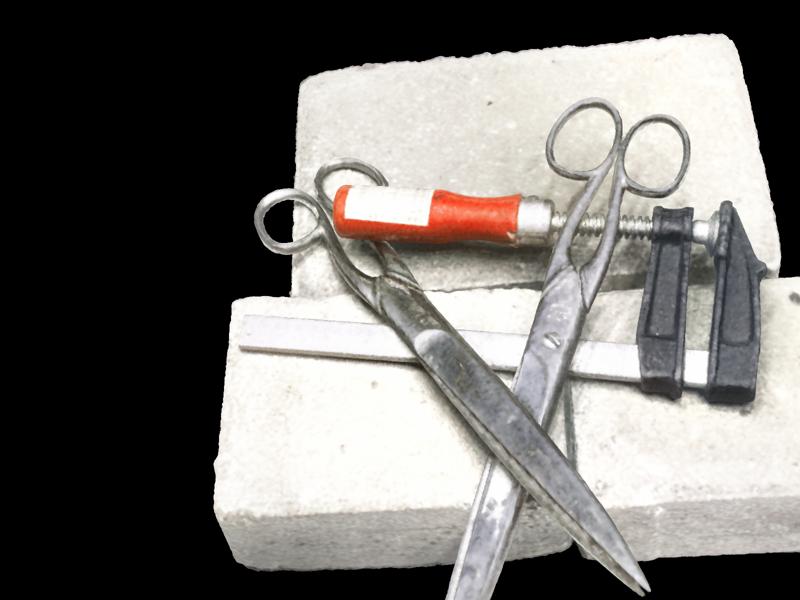}    & \includegraphics[valign=m,width=.15\textwidth]{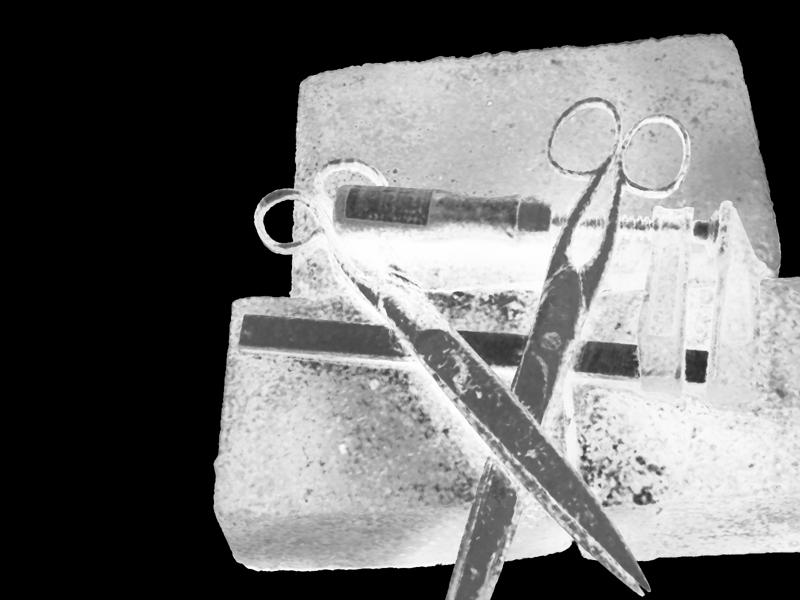}    & \includegraphics[valign=m,width=.15\textwidth]{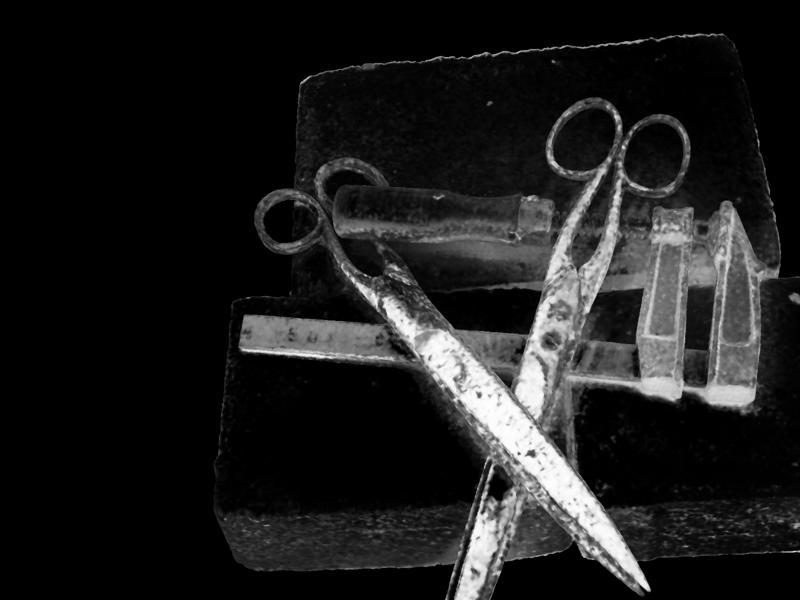}    & \includegraphics[valign=m,width=.225\textwidth]{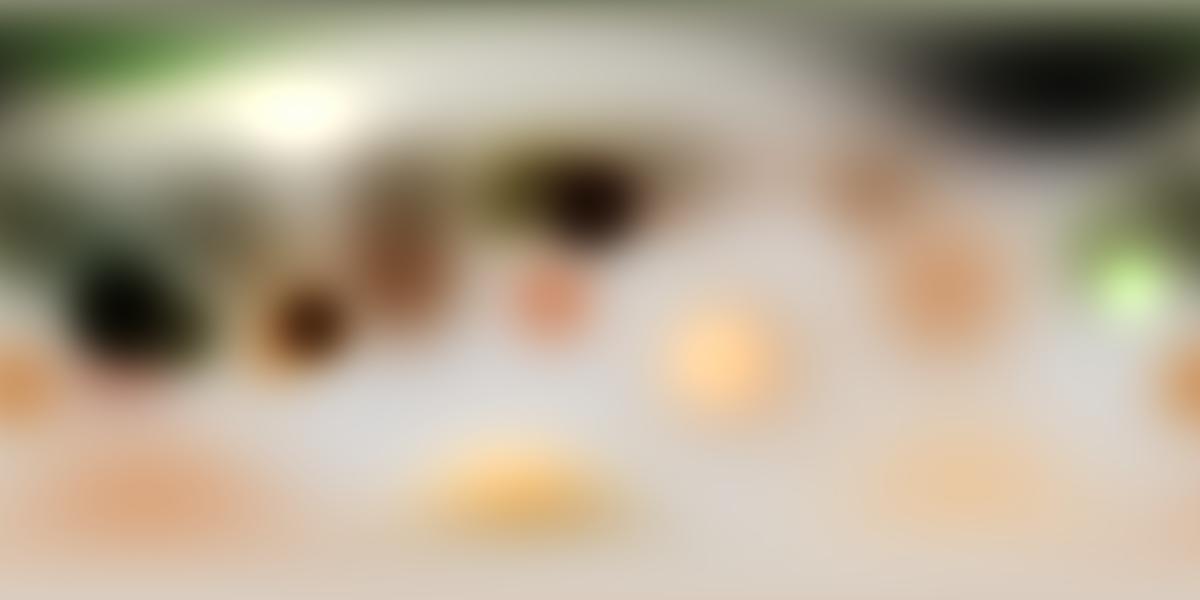} \\
\end{tabular}
    \caption{Decomposed scene components on DTU scenes without emissive sources.}
    \label{fig:dtu1}
\end{figure*}

\begin{figure*}[t]
    \centering
    \scriptsize
    \renewcommand{\arraystretch}{3} 
    \begin{tabular}{@{} *{7}{c@{\hspace{2pt}}} @{}}
        \multicolumn{1}{c}{Image}                                                                 & \multicolumn{1}{c}{Normal}                                                                    & \multicolumn{1}{c}{Base Color}                                                                   & \multicolumn{1}{c}{Roughness}                                                                    & \multicolumn{1}{c}{Metallic}                                                                    & \multicolumn{1}{c}{Env. Map}                                                                \\
        \includegraphics[valign=m,width=.15\textwidth]{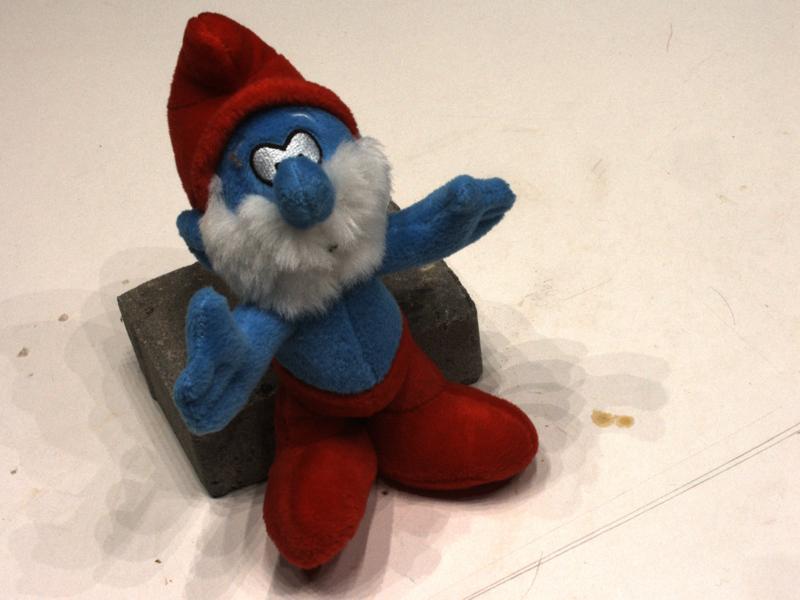}  & \includegraphics[valign=m,width=.15\textwidth]{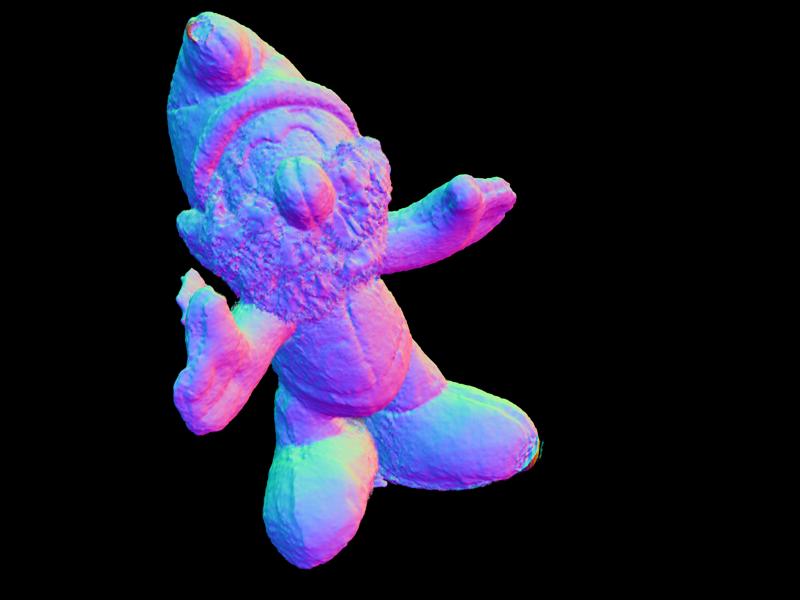}     & \includegraphics[valign=m,width=.15\textwidth]{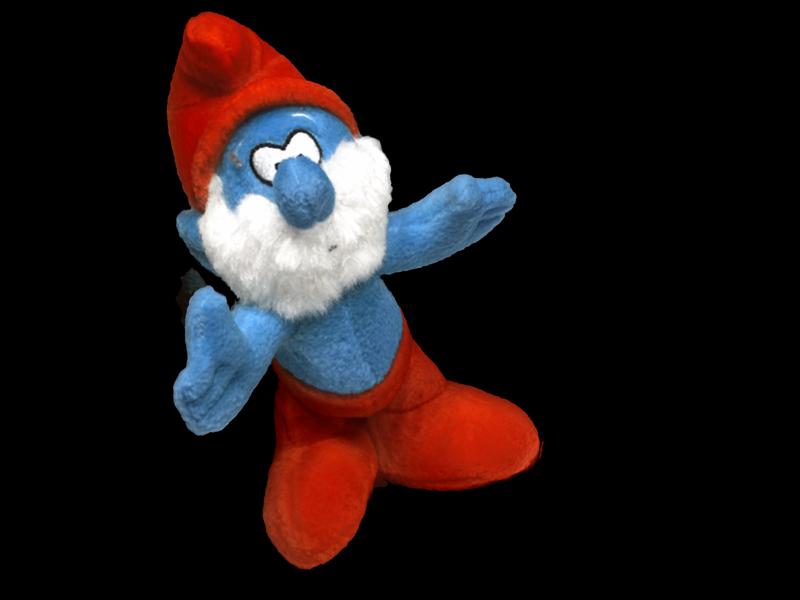}     & \includegraphics[valign=m,width=.15\textwidth]{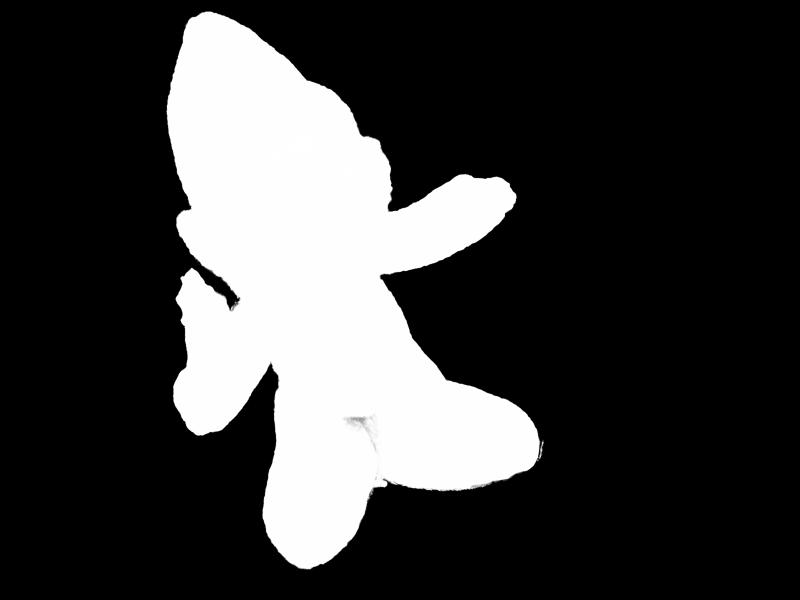}     & \includegraphics[valign=m,width=.15\textwidth]{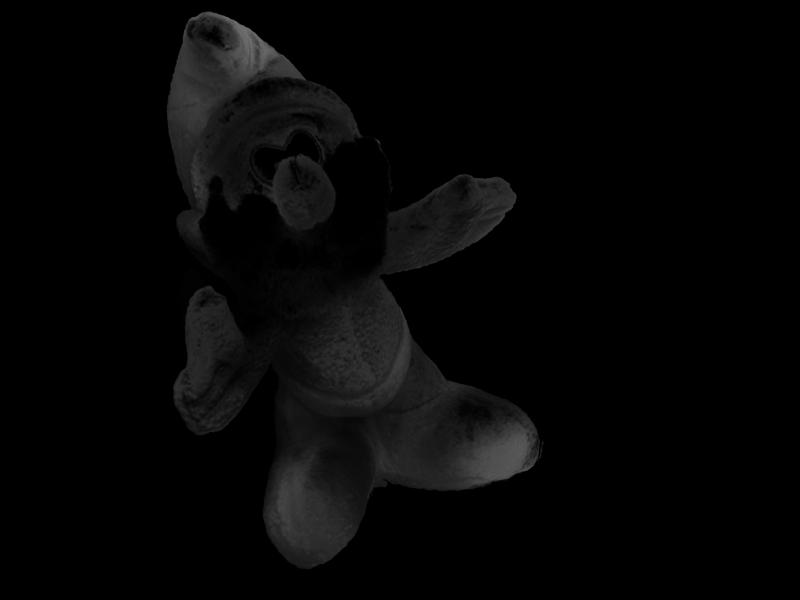}     & \includegraphics[valign=m,width=.225\textwidth]{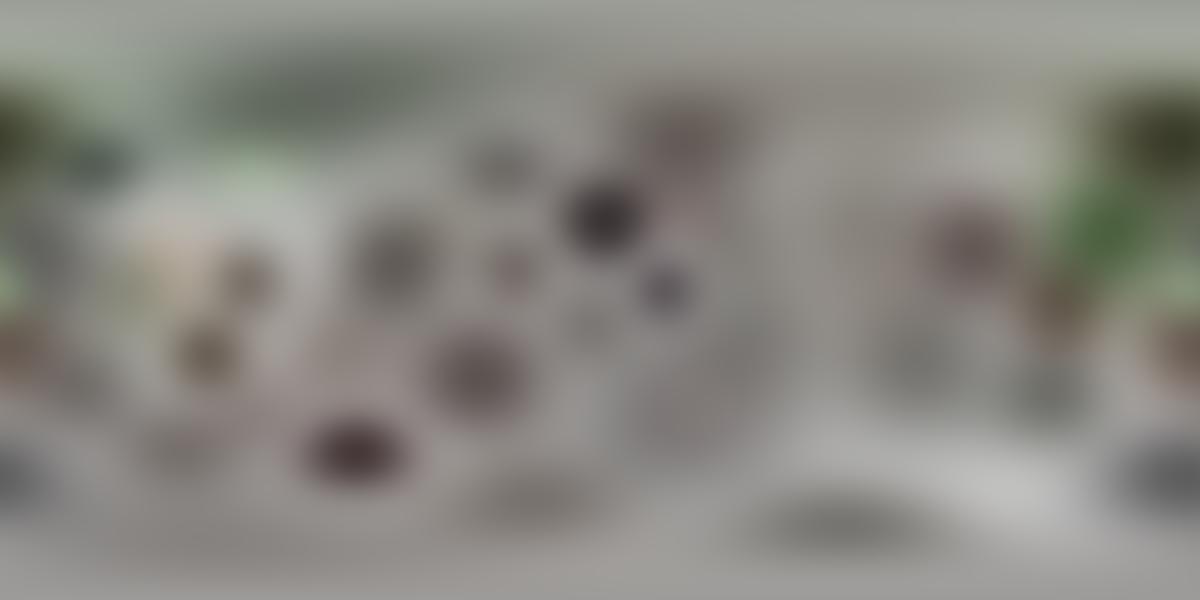}  \\
        \includegraphics[valign=m,width=.15\textwidth]{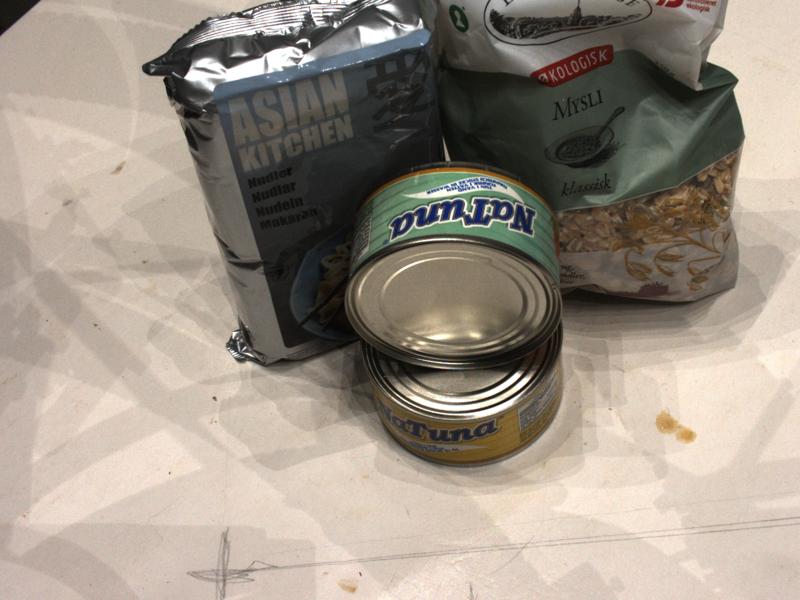}  & \includegraphics[valign=m,width=.15\textwidth]{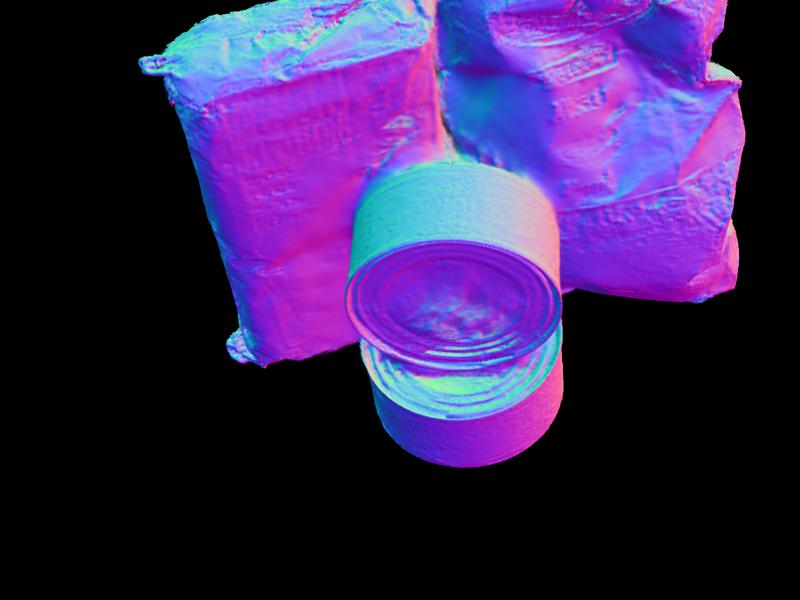}    & \includegraphics[valign=m,width=.15\textwidth]{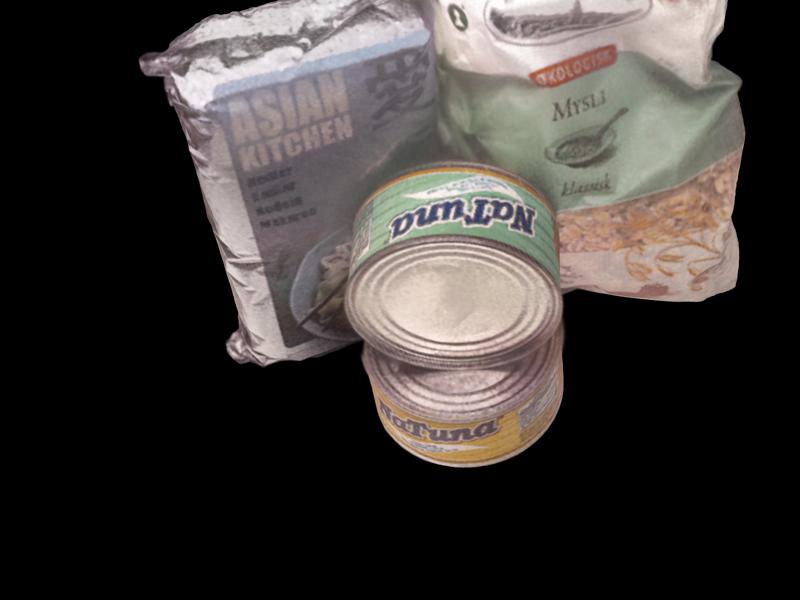}     & \includegraphics[valign=m,width=.15\textwidth]{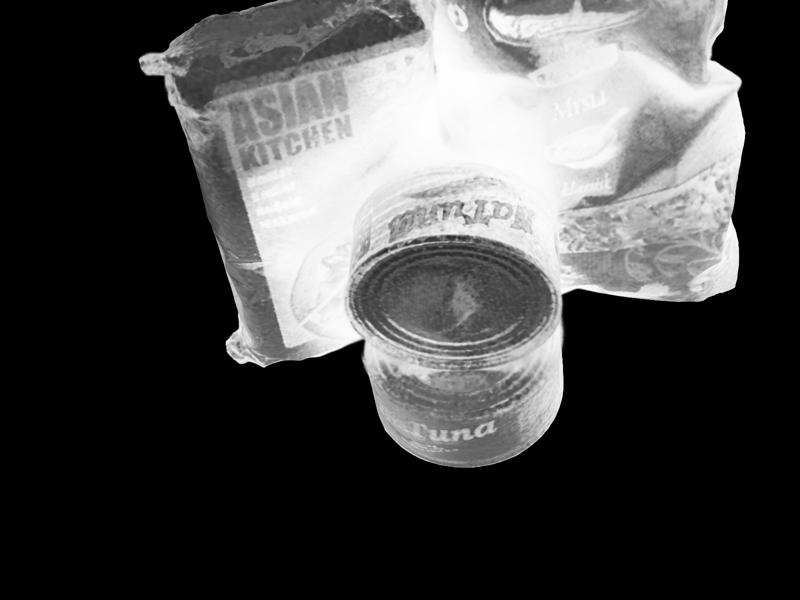}     & \includegraphics[valign=m,width=.15\textwidth]{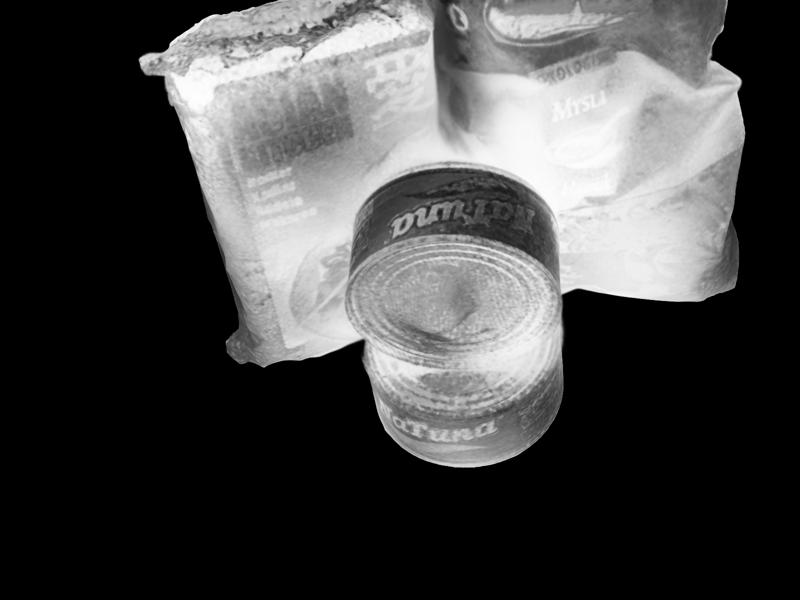}     & \includegraphics[valign=m,width=.225\textwidth]{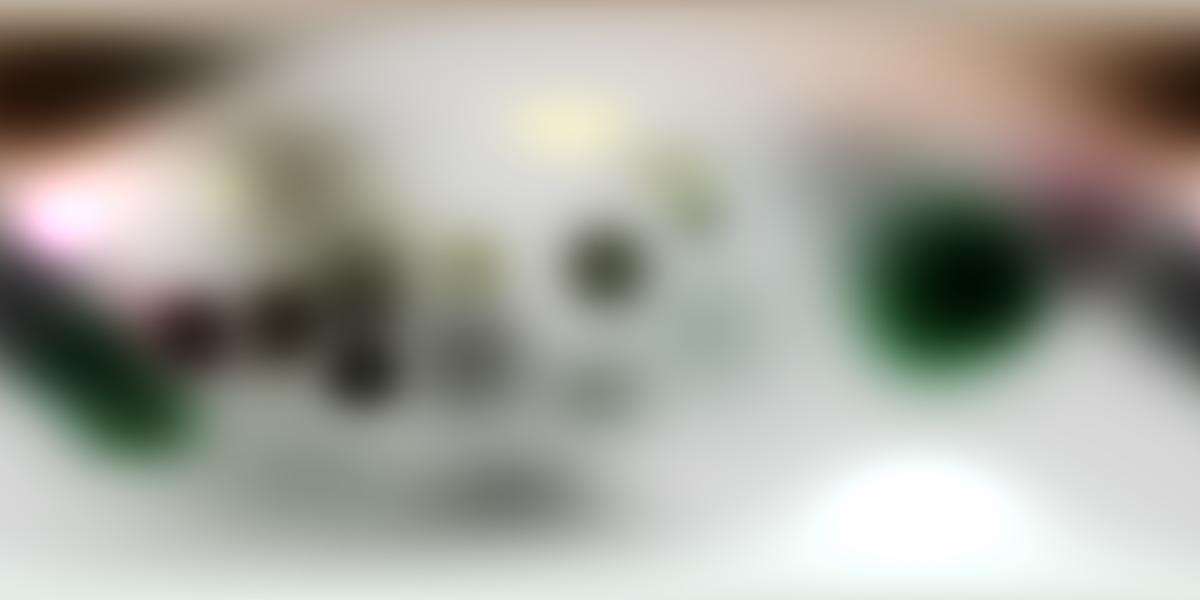}  \\
\end{tabular}
    \caption{Decomposed scene components on DTU scenes without emissive sources.}
    \label{fig:dtu2}
\end{figure*}

We present the reconstructed components of our synthetic scenes, including emissions, surface normals, and BRDF, in Fig.~\ref{fig:brdf_w} and~\ref{fig:brdf_c}.
We also provide the comparison of the reconstructed emission and BRDF performance among TensoIR, NeILF++, and \modelname in Fig.~\ref{fig:real_snack_jobs_comparison} and~\ref{fig:real_cosmetic_dolls_comparison} for real scenes and Fig.~\ref{fig:lego_comparison} to~\ref{fig:billboard_comparison} for synthetic scenes.
TensoIR and NeILF++ encounter difficulties, as does \modelname, in capturing precise roughness, often resulting in shadows being baked into the albedo.
This issue is exacerbated by a relatively dark environment map, in contrast to previous works, and is compounded by strong emissions and shadows.
Nevertheless, while BRDF results are comparable, \modelname distinguishes itself in its primary goal: the accurate reconstruction of emissive sources
We also provide the reconstructed scene components on DTU dataset in Fig.~\ref{fig:dtu1} and~\ref{fig:dtu2}.

\begin{table*}[b]
    \centering
    \scriptsize
    \setlength{\tabcolsep}{3pt}
    \begin{tabular}{@{}c*{12}{>{\centering\arraybackslash}p{13pt}}:*{12}{>{\centering\arraybackslash}p{13pt}}}
        \toprule
                        & \multicolumn{12}{c|}{White colored} & \multicolumn{12}{c}{Vivid colored}                                                                                                                                                                                                                                                                                                                                                                                 \\
                        & \multicolumn{2}{c}{Lego}            & \multicolumn{2}{c}{Gift}           & \multicolumn{2}{c}{Book} & \multicolumn{2}{c}{Cube} & \multicolumn{2}{c}{Billboard} & \multicolumn{2}{c|}{Balls} & \multicolumn{2}{c}{Lego} & \multicolumn{2}{c}{Gift} & \multicolumn{2}{c}{Book} & \multicolumn{2}{c}{Cube} & \multicolumn{2}{c}{Billboard} & \multicolumn{2}{c}{Balls}                                                                                      \\
        \cmidrule(lr){2-3} \cmidrule(lr){4-5} \cmidrule(lr){6-7} \cmidrule(lr){8-9} \cmidrule(lr){10-11} \cmidrule(lr){12-13} \cmidrule(lr){14-15} \cmidrule(lr){16-17}\cmidrule(lr){18-19}\cmidrule(lr){20-21}\cmidrule(lr){22-23}\cmidrule(lr){24-25}
                        & IoU                                 & MSE                                & IoU                      & MSE                      & IoU                           & MSE                        & IoU                      & MSE                      & IoU                      & MSE                      & IoU                           & MSE                       & IoU  & MSE  & IoU  & MSE  & IoU  & MSE  & IoU  & MSE   & IoU  & MSE  & IoU  & MSE  \\
        w/o progressive & 0.09                                & 18.87                              & 0.05                     & 5.93                     & 0.38                          & 2.84                       & 0.82                     & 30.82                    & 0.14                     & 1.00                     & 0.93                          & 0.04                      & 0.09 & 6.71 & 0.05 & 3.89 & 0.37 & 1.69 & 0.84 & 10.60 & 0.14 & 0.64 & 0.94 & 0.02 \\
        w/o sg          & 0.79                                & 8.33                               & 0.50                     & 5.32                     & 0.35                          & 2.91                       & 0.96                     & 21.28                    & 0.72                     & 0.80                     & 0.95                          & 0.04                      & 0.16 & 6.43 & 0.35 & 3.60 & 0.35 & 1.87 & 0.93 & 8.65  & 0.89 & 0.25 & 0.92 & 0.03 \\
        \modelname      & 0.81                                & 8.38                               & 0.60                     & 3.49                     & 0.96                          & 1.19                       & 0.97                     & 17.87                    & 0.84                     & 0.46                     & 0.95                          & 0.04                      & 0.51 & 5.48 & 0.59 & 2.50 & 0.96 & 0.51 & 0.97 & 7.94  & 0.88 & 0.26 & 0.94 & 0.03 \\
        \bottomrule
    \end{tabular}
    \caption{
        Per-scene metrics on emissive source reconstruction tasks.
        The IoU measures the source area identification (a higher value is better), and the MSE quantifies the difference between reconstructed images and HDR ground truth images (a lower value is better).
    }
    \label{tab:perscene}
    \vspace{-10pt}
\end{table*}

\subsection{Illumination Decomposition}
We present additional results of decomposed illumination in Fig.~\ref{fig:illum_decomp_1}.
These visualizations offer insights into the effectiveness of \modelname in factorizing the scene illumination.
The off image, for instance, is generated by merging direct and indirect illumination from the environment map, as shown in the first row.
The second row illustrates the decomposition of emission effects, including both the emission and its reflection.
Light-on images are created by adding the light-off and the emission effects images.

\subsection{Scene Editing w\&w/o Radiance Fine-tuning}
\begin{figure*}[t]
    \centering
    \scriptsize
    \renewcommand{\arraystretch}{3} 
    \begin{tabular}{@{} *{4}{c@{\hspace{2pt}}}: *{4}{c@{\hspace{2pt}}}@{}}
        \multicolumn{1}{c}{Image}                                                                              & \multicolumn{1}{c}{Intensity Edit}                                                                                 & \multicolumn{1}{c}{Color Edit}                                                                                          & \multicolumn{1}{c}{Intensity \& Color Edit}                                                                              & \multicolumn{1}{c}{Image}                                                                            & \multicolumn{1}{c}{Intensity Edit}                                                                                 & \multicolumn{1}{c}{Color Edit}                                                                                     & \multicolumn{1}{c}{Intensity \& Color Edit}                                                                         \\
        \includegraphics[valign=m,width=.12\linewidth]{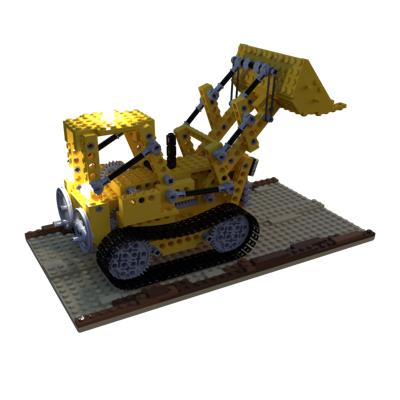}      & \includegraphics[valign=m,width=.12\linewidth]{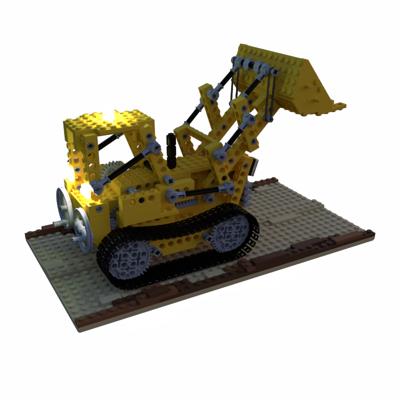} & \includegraphics[valign=m,width=.12\linewidth]{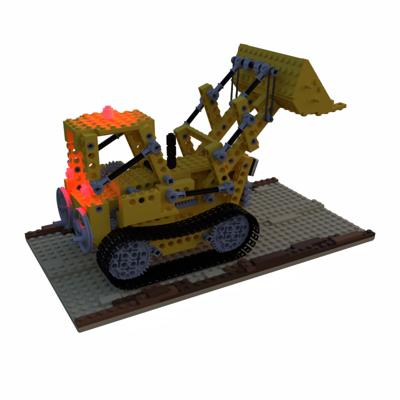}      & \includegraphics[valign=m,width=.12\linewidth]{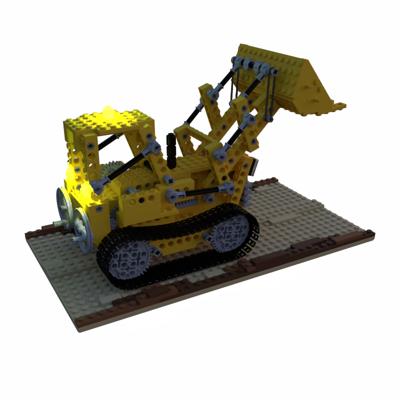}      & \includegraphics[valign=m,width=.12\linewidth]{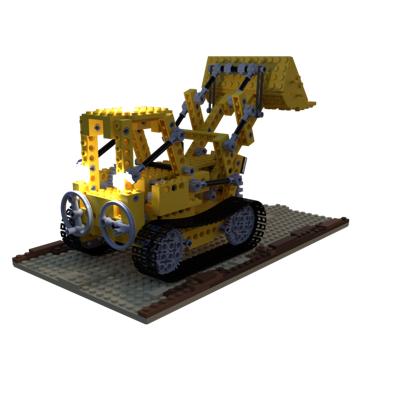}      & \includegraphics[valign=m,width=.12\linewidth]{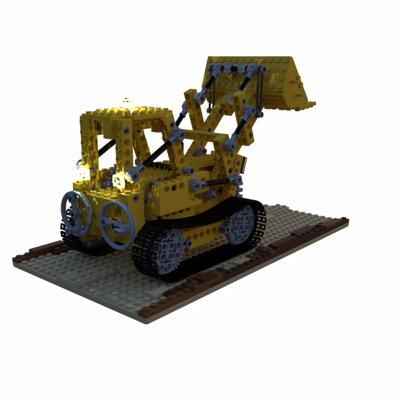}      & \includegraphics[valign=m,width=.12\linewidth]{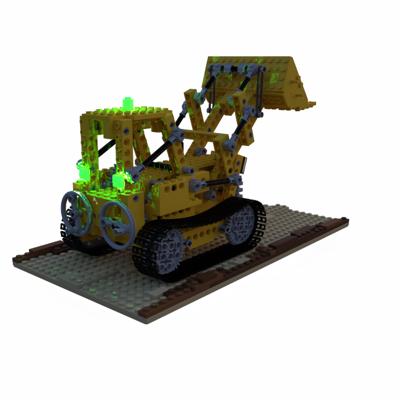}      & \includegraphics[valign=m,width=.12\linewidth]{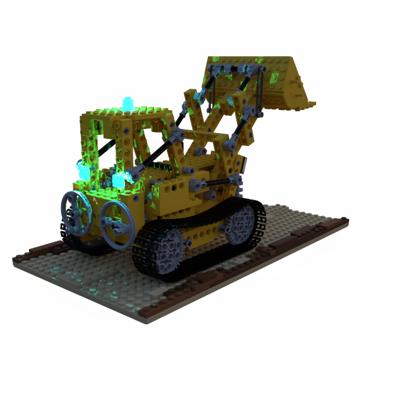}      \\
\end{tabular}
    \caption{Re-lighting scenes containing white emissive sources. Left: through fine-tuning radiance fields, Right: computing direct illumination from reconstructed emissive sources.}
    \label{fig:finetune_w}
\end{figure*}

\begin{figure*}[t]
    \centering
    \scriptsize
    \renewcommand{\arraystretch}{3} 
    \begin{tabular}{@{} *{4}{c@{\hspace{2pt}}}:*{4}{c@{\hspace{2pt}}} @{}}
        \multicolumn{1}{c}{Image}                                                                              & \multicolumn{1}{c}{Intensity Edit}                                                                                      & \multicolumn{1}{c}{Color Edit}                                                                                          & \multicolumn{1}{c}{Intensity \& Color Edit}                                                                              & \multicolumn{1}{c}{Image}                                                                            & \multicolumn{1}{c}{Intensity Edit}                                                                                 & \multicolumn{1}{c}{Color Edit}                                                                                     & \multicolumn{1}{c}{Intensity \& Color Edit}                                                                         \\
        \includegraphics[valign=m,width=.12\linewidth]{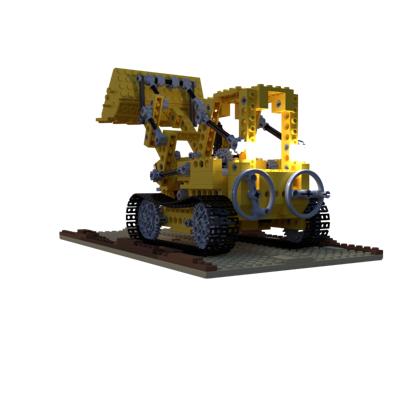}     & \includegraphics[valign=m,width=.12\linewidth]{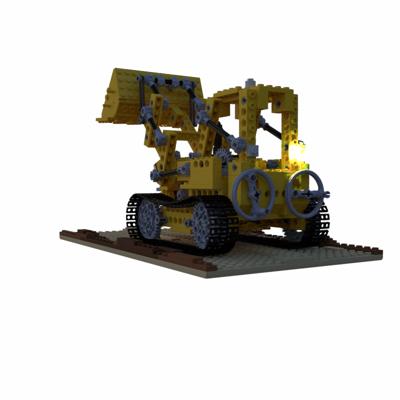}      & \includegraphics[valign=m,width=.12\linewidth]{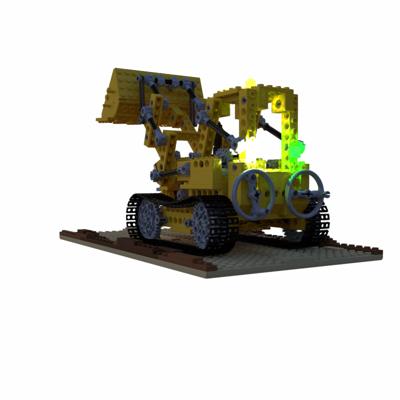}      & \includegraphics[valign=m,width=.12\linewidth]{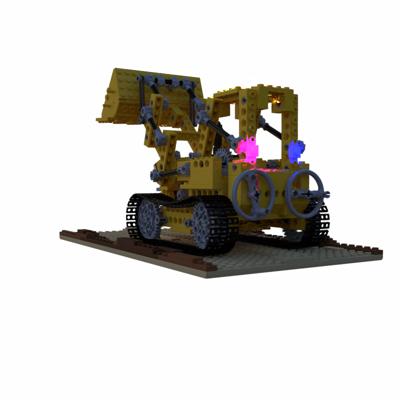}      & \includegraphics[valign=m,width=.12\linewidth]{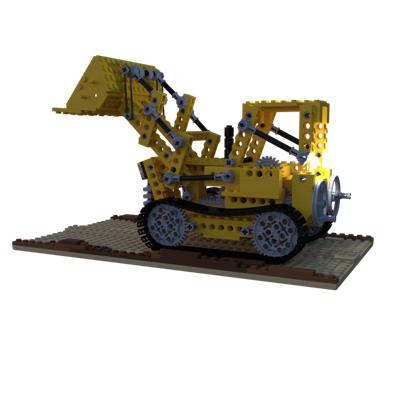}      & \includegraphics[valign=m,width=.12\linewidth]{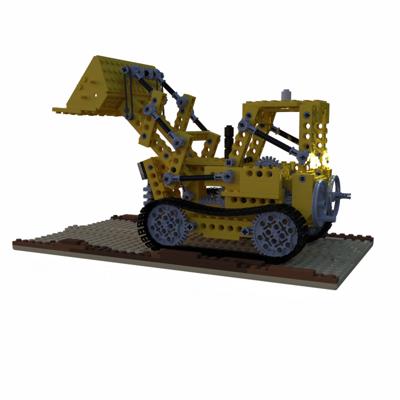}      & \includegraphics[valign=m,width=.12\linewidth]{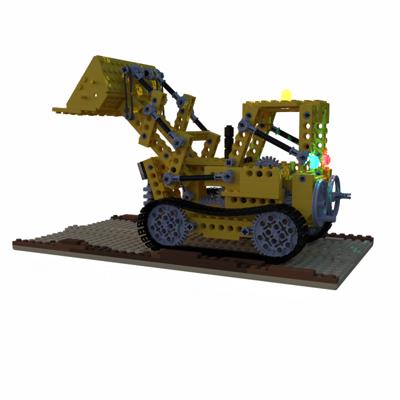}      & \includegraphics[valign=m,width=.12\linewidth]{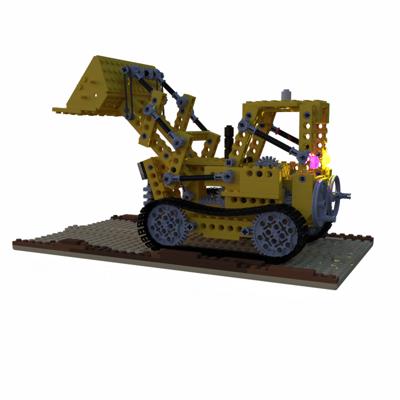}      \\
\end{tabular}
    \caption{Individual emissive sources control. Left: through fine-tuning radiance fields, Right: computing direct illumination from reconstructed emissive sources.}
    \label{fig:finetune_i}
\end{figure*}

Fig.~\ref{fig:finetune_w} and~\ref{fig:finetune_i} present additional scene editing examples, illustrating various scenarios including intensity and color edits, as well as their combination.
As discussed in the conclusion section of the main paper, scene illumination can be adjusted without fine-tuning radiance fields, using alternative methods.
Results on the right side of Fig.~\ref{fig:finetune_w} to~\ref{fig:finetune_i} are rendered by calculating only direct illumination from emissive sources for re-lighting, a technique commonly used in prior research~\cite{Jin2023TensoIR,modelingindirect2022,nerfactor2021}, bypassing the fine-tuning of trained networks.
This approach is particularly effective for scenes with vividly colored emissive sources, as shown in Fig.~\ref{fig:direct_c}.
To evaluate the effectiveness of direct illumination in scene editing, we provide quantitative results for each scenario in Tab.~\ref{tab:white_finetune} and~\ref{tab:white_direct}.
Quantitative comparisons for scenes with vivid-colored emissive sources are detailed in Tab.~\ref{tab:vivid_finetune} and~\ref{tab:vivid_direct}.

\subsection{Analysis of Learnable Tone-mapper}

We eliminate the constraint on the range of radiance values to address the unbounded nature of emissive sources and their reflections.
Instead of the commonly used sigmoid activation function in NeRF-based methods~\cite{nerf2020,nerv2021,tensorf2022,modelingindirect2022,media2021} for radiance prediction, we employ the softplus activation, extending the radiance range from $[0,1]$ to $[0,\infty]$.

\begin{figure}[t]
    \vspace{-10pt}
    \centering
    \scriptsize
    \renewcommand{\arraystretch}{3} 
    \begin{tabular}{@{} *{4}{c@{\hspace{2pt}}} @{}}
        \multicolumn{1}{c}{Image}                                                                             & \multicolumn{1}{c}{Normal}                                                                                             & \multicolumn{1}{c}{Rendered}                                                                                                   \\
        \includegraphics[width=.33\linewidth]{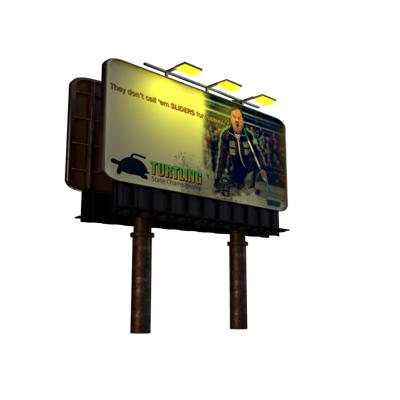} & \includegraphics[width=.33\linewidth]{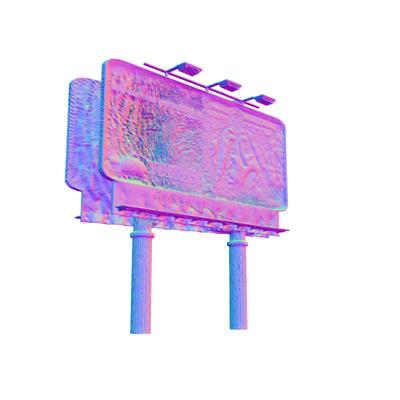} & \includegraphics[width=.33\linewidth]{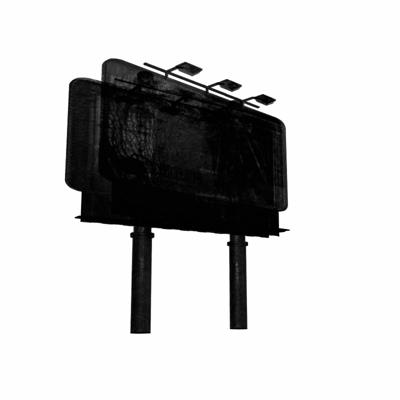} \\
        \includegraphics[width=.33\linewidth]{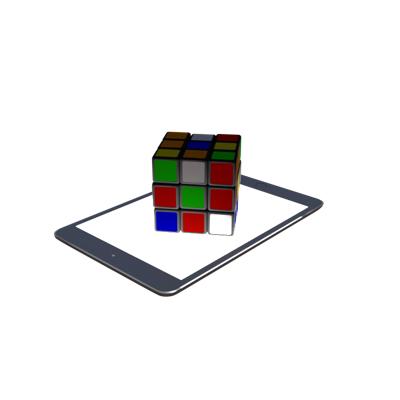}       & \includegraphics[width=.33\linewidth]{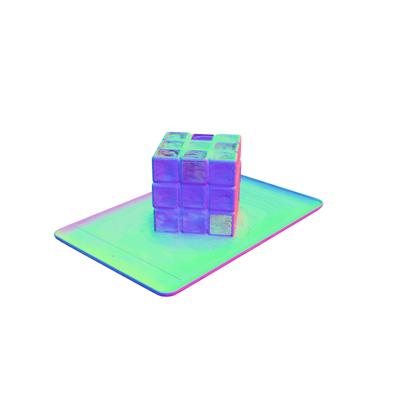}      & \includegraphics[width=.33\linewidth]{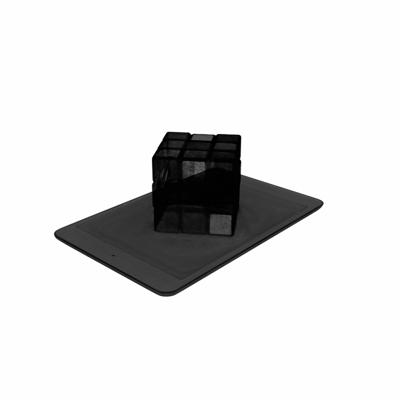}      \\
        \includegraphics[width=.33\linewidth]{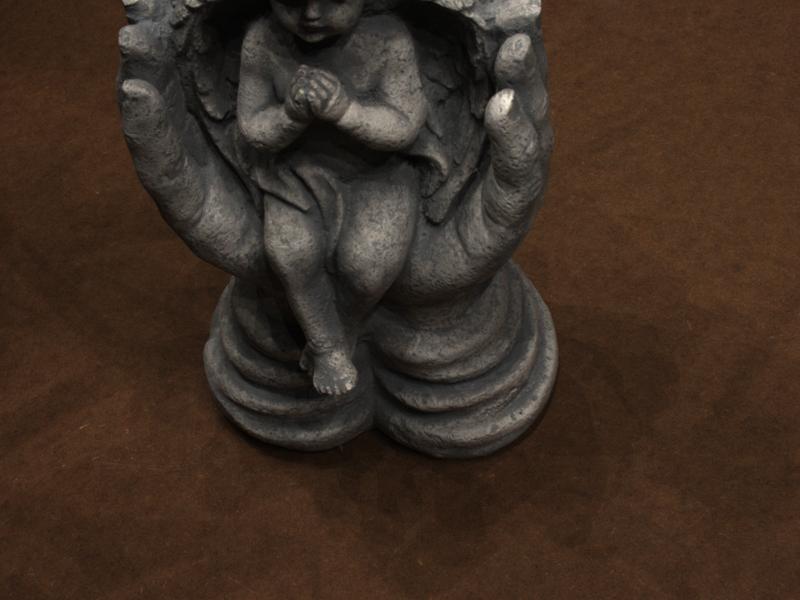}    & \includegraphics[width=.33\linewidth]{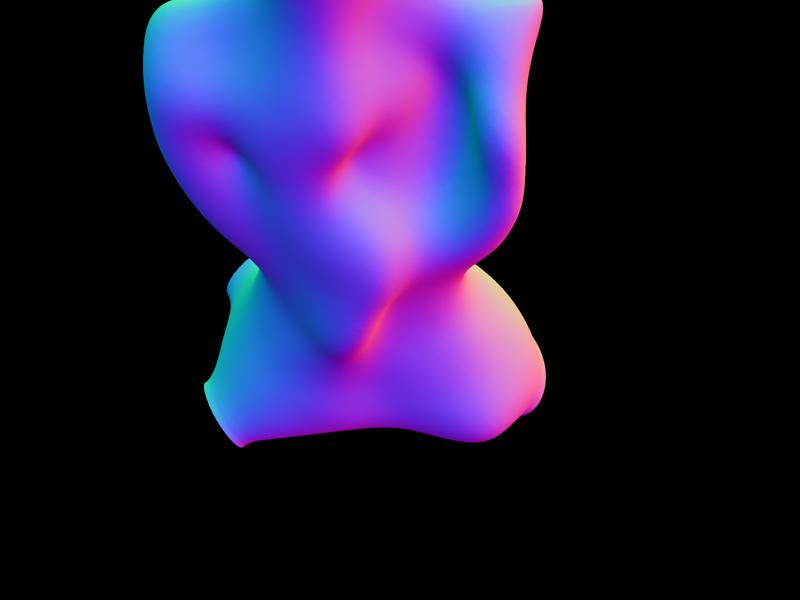}                    & \includegraphics[width=.33\linewidth]{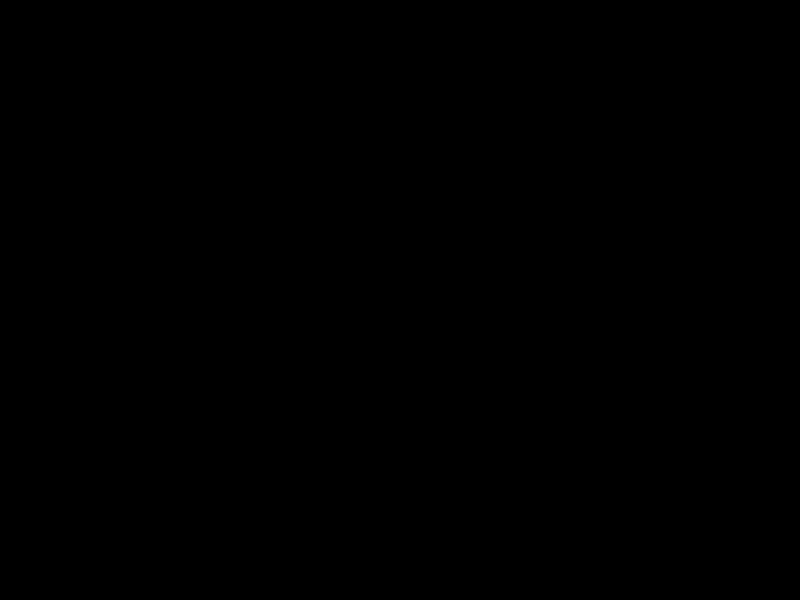}                        \\
        \includegraphics[width=.33\linewidth]{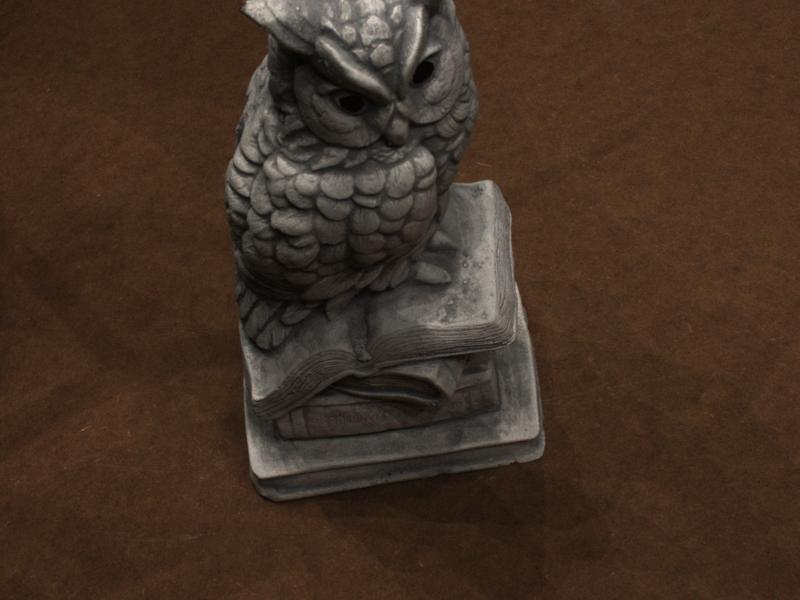}    & \includegraphics[width=.33\linewidth]{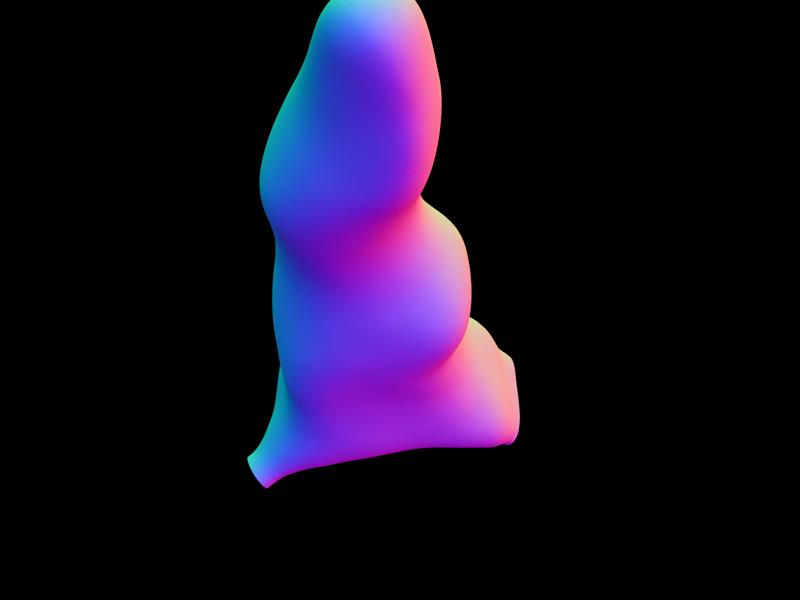}                    & \includegraphics[width=.33\linewidth]{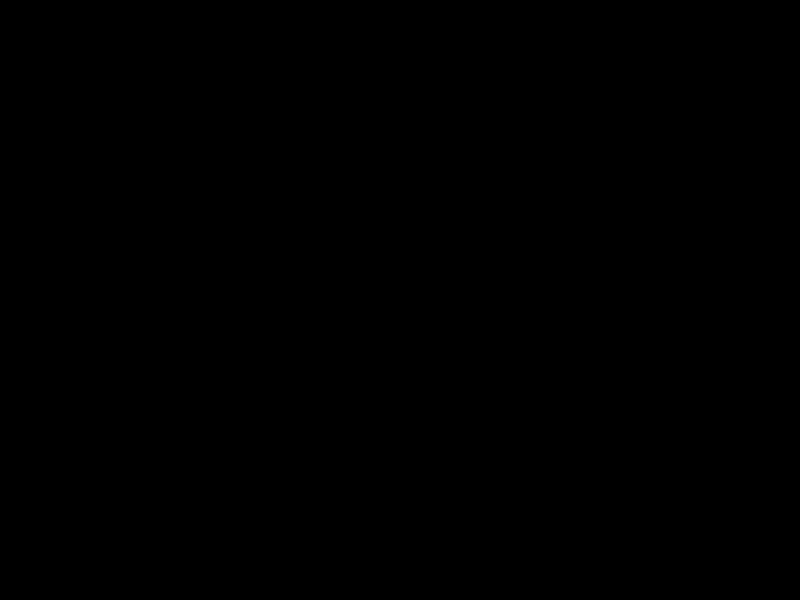}                        \\
    \end{tabular}
    \caption{Erroneously reconstructed surfaces and rendered linear images when using softplus activation for radiances without utilizing the tone-mapper $m_\theta$. Gamma correction is applied to linear images for easy comparison.}
    \label{fig:no_tonemap}
    \vspace{-10pt}
\end{figure}

However, this modification may lead to inaccurate surface reconstructions, as highlighted in the main paper.
Fig.~\ref{fig:no_tonemap} shows instances where surfaces become semi-transparen, lose structural details, and the rendered images significantly deviate from the ground truth, making the accurate reconstruction of emissive sources infeasible.

\begin{figure}[t]
    \centering
    \scriptsize
    \renewcommand{\arraystretch}{3} 
    \begin{tabular}{@{} *{5}{c@{\hspace{2pt}}} c @{}}
        Image                                                                                                                         & \multicolumn{1}{c}{$\lambda_\tau=0.001$}                                                                                              & \multicolumn{1}{c}{$\lambda_\tau=0.01$}                                                                                               & \multicolumn{1}{c}{$\lambda_\tau=0.1$}                                                                                                & \multicolumn{1}{c}{$\lambda_\tau=1.0$}                                                                                                                                                 \\
        \multirow{2}{*}{\includegraphics[valign=m,width=.19\linewidth]{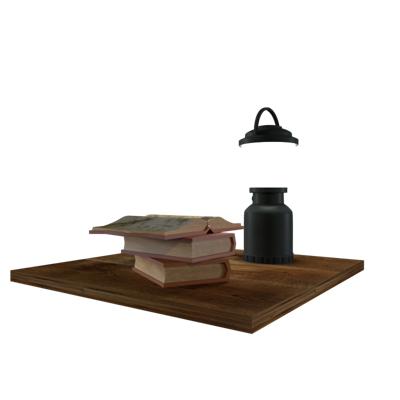}} & \includegraphics[valign=m,width=.19\linewidth]{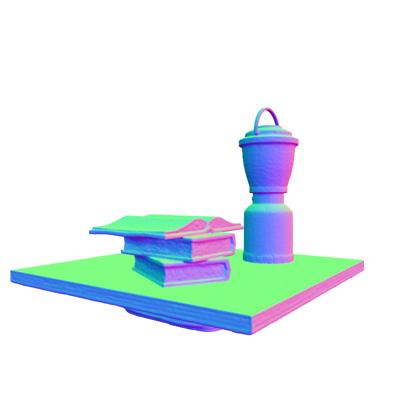}        & \includegraphics[valign=m,width=.19\linewidth]{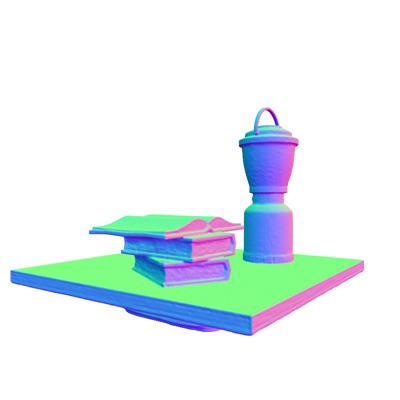}        & \includegraphics[valign=m,width=.19\linewidth]{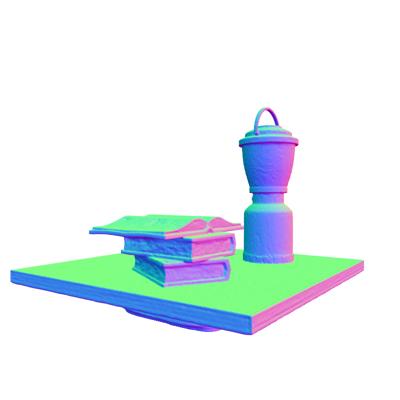}        & \includegraphics[valign=m,width=.19\linewidth]{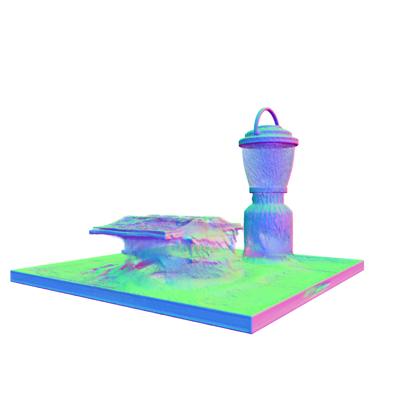}        & \rotatebox[origin=c]{-90}{\makecell{Normal}}   \\
\end{tabular}
    \caption{Reconstructed surface normals and rendered linear images with varying $\lambda_\tau$ values. Gamma correction is applied to linear images for easy comparison.}
    \label{fig:tonemap}
\end{figure}

To address this issue, we introduce a learnable tone-mapper $m_\theta$, taking positionally encoded HDR linear color as input and produce LDR sRGB colors outputs.
Fig.~\ref{fig:tonemap} reveals that this tone-mapper helps in obtaining accurate surface normals and rendering photo-realistic images.
Nonetheless, a trade-off exists between the quality of surface normals and rendered images, when using the learnable tone-mapper.
For example, a low $\lambda_\tau$ value, which indicates a heavier reliance on the tone-mapper in the rendering loss, may improve surface details but linear color values deviate significantly from expectations.
This discrepancy occurs as the correlation between predicted linear colors and actual image pixel colors weakens with lower $\lambda_\tau$ values.
Conversely, a higher $\lambda_\tau$ compromises surface reconstruction quality.
Thus, setting $\lambda_\tau$ requires careful consideration of the balance between surface detail and color accuracy.

\begin{figure}[b]
    \vspace{-10pt}
    \centering
    \footnotesize
    \renewcommand{\arraystretch}{1.3} 
    \begin{tabular}{@{} *{3}{c@{\hspace{4pt}}} @{}}
        \multicolumn{1}{c}{High $\lambda_\tau$}                                                      & \multicolumn{1}{c}{Low $\lambda_\tau$}                                      & \multicolumn{1}{c}{Pseudo G.T.}                                                     \\
        \includegraphics[width=.28\linewidth]{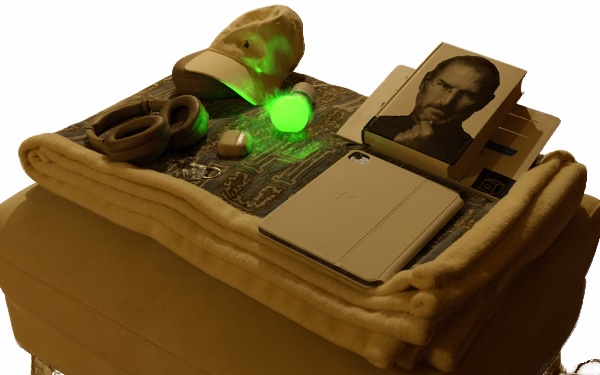} & \includegraphics[width=.28\linewidth]{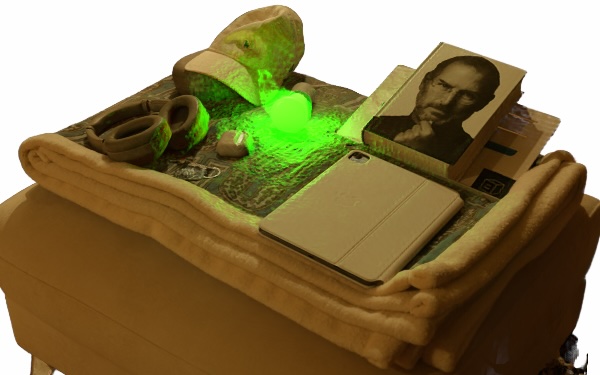} & \includegraphics[width=.28\linewidth]{figures/rebuttal/nvc_gt_crop_crop_resize.jpg} \\
    \end{tabular}
    \vspace{-10pt}
    \caption{
        Scene edit results on jobs scene. Lower $\lambda_\tau$ results in stronger emission.
        The middle image is rendered with direct light for proving enhanced emission strength.
    }
    \label{fig:tonemap_real}
\end{figure}

Interestingly, the choice of $\lambda_\tau$ also impacts the reconstruction of emissive sources in real scenes.
A high $\lambda_\tau$ tends to result in lower intensity of reconstructed emissive sources.
Re-lighting experiments in Fig.~\ref{fig:tonemap_real} show illumination effects confined to a narrow area compared to ground truth data.
We suspect the camera may edit images for low contrast and apply color grading, particularly in HDR scenes.
We used the Fuji 100s camera.
A high $\lambda_\tau$ in the rendering loss could be problematic, as it aims to align gamma-corrected linear values with manipulated colors.
Based on this insight, we slightly reduced $\lambda_\tau$ by 0.1 to enhance emission intensity (1.4 vs. 37.2) and expand reflections in re-lighting scenarios.

\subsection{Near-zero IoU Results of Baselines}
State-of-the-art re-lighting methods struggle with ambiguities surrounding emissive sources, often failing to accurately identify them.
These methods typically cannot differentiate between reflections and emissions, leading to most regions being misclassified as emissive sources.
This challenge is reflected in Tab.~\ref{tab:threhold}, where baseline methods exhibit near-zero IoU performance across various scenes.
Despite extensive trinary grid searches with an interval of 0.01 for thresholding values to report the peak performance of baselines, \modelname consistently outperforms them.
Additionally, our method's efficacy in classifying rays into the uncertain group for emissive source identification highlights its superiority in this task.
This is further supported by additional results obtained using thresholding techniques applied to the baselines.

\begin{table*}[t]
    \centering
    \scriptsize
    \setlength{\tabcolsep}{10pt}
    \begin{tabular}{@{}c*{6}{c}|*{6}{c}}
        \toprule
                         & \multicolumn{6}{c|}{White colored} & \multicolumn{6}{c}{Vivid colored}                                                                                                                                                                                                                                                                                                                                                     \\
                         & \multicolumn{1}{c}{Lego}           & \multicolumn{1}{c}{Gift}          & \multicolumn{1}{c}{Book}        & \multicolumn{1}{c}{Cube}        & \multicolumn{1}{c}{Billboard}   & \multicolumn{1}{c|}{Balls}      & \multicolumn{1}{c}{Lego}        & \multicolumn{1}{c}{Gift}        & \multicolumn{1}{c}{Book}        & \multicolumn{1}{c}{Cube}        & \multicolumn{1}{c}{Billboard}   & \multicolumn{1}{c}{Balls}       \\
        \midrule
        NeILF++          & \textcolor{teal}{\textbf{0.00}}    & \textcolor{teal}{\textbf{0.01}}   & \textcolor{teal}{\textbf{0.04}} & \textcolor{teal}{\textbf{0.39}} & 0.00                            & \textcolor{teal}{\textbf{0.07}} & \textcolor{teal}{\textbf{0.00}} & \textcolor{teal}{\textbf{0.01}} & \textcolor{teal}{\textbf{0.04}} & \textcolor{teal}{\textbf{0.39}} & 0.00                            & \textcolor{teal}{\textbf{0.07}} \\
        TensoIR          & \textcolor{teal}{\textbf{0.00}}    & \textcolor{teal}{\textbf{0.01}}   & \textcolor{teal}{\textbf{0.04}} & 0.37                            & \textcolor{teal}{\textbf{0.01}} & \textcolor{teal}{\textbf{0.07}} & \textcolor{teal}{\textbf{0.00}} & \textcolor{teal}{\textbf{0.01}} & \textcolor{teal}{\textbf{0.04}} & 0.37                            & \textcolor{teal}{\textbf{0.01}} & \textcolor{teal}{\textbf{0.07}} \\
        \modelname       & \textcolor{blue}{\textbf{0.81}}    & \textcolor{blue}{\textbf{0.60}}   & \textcolor{blue}{\textbf{0.96}} & \textcolor{blue}{\textbf{0.97}} & \textcolor{blue}{\textbf{0.84}} & \textcolor{blue}{\textbf{0.95}} & \textcolor{blue}{\textbf{0.51}} & \textcolor{blue}{\textbf{0.59}} & \textcolor{blue}{\textbf{0.96}} & \textcolor{blue}{\textbf{0.97}} & \textcolor{blue}{\textbf{0.88}} & \textcolor{blue}{\textbf{0.94}} \\
        \midrule
        NeILF++ ($*$)    & 0.43                               & 0.07                              & \textcolor{teal}{\textbf{0.95}} & 0.93                            & 0.01                            & 0.91                            & 0.30                            & 0.09                            & \textcolor{teal}{\textbf{0.95}} & 0.94                            & 0.02                            & 0.92                            \\
        TensoIR ($*$)    & \textcolor{teal}{\textbf{0.71}}    & \textcolor{teal}{\textbf{0.15}}   & \textcolor{teal}{\textbf{0.95}} & \textcolor{teal}{\textbf{0.95}} & \textcolor{teal}{\textbf{0.76}} & \textcolor{teal}{\textbf{0.95}} & \textcolor{teal}{\textbf{0.33}} & \textcolor{teal}{\textbf{0.15}} & \textcolor{teal}{\textbf{0.95}} & \textcolor{teal}{\textbf{0.96}} & \textcolor{teal}{\textbf{0.77}} & \textcolor{blue}{\textbf{0.95}} \\
        \modelname ($*$) & \textcolor{blue}{\textbf{0.81}}    & \textcolor{blue}{\textbf{0.60}}   & \textcolor{blue}{\textbf{0.98}} & \textcolor{blue}{\textbf{0.98}} & \textcolor{blue}{\textbf{0.94}} & \textcolor{blue}{\textbf{0.96}} & \textcolor{blue}{\textbf{0.51}} & \textcolor{blue}{\textbf{0.61}} & \textcolor{blue}{\textbf{0.98}} & \textcolor{blue}{\textbf{0.97}} & \textcolor{blue}{\textbf{0.93}} & \textcolor{teal}{\textbf{0.94}} \\
        \bottomrule
    \end{tabular}
    \caption{
        Results of emissive source identification.
        The IoU measures the source area identification (a higher value is better).
        The asterisk (*) denotes that thresholding is applied to reconstructed emission strengths.
    }
    \label{tab:threhold}
\end{table*}

\subsection{Failure Cases in Scene Editing}
\begin{figure}[t]

    \centering
    \scriptsize
    \renewcommand{\arraystretch}{3} 
    \begin{tabular}{@{} *{3}{c@{\hspace{2pt}}} @{}}
        \multicolumn{1}{c}{Image}                                                               & \multicolumn{1}{c}{Edit}                                                                                  & \multicolumn{1}{c}{G.T}                                                                   \\
        \vspace{-10pt}
        \includegraphics[width=.33\linewidth]{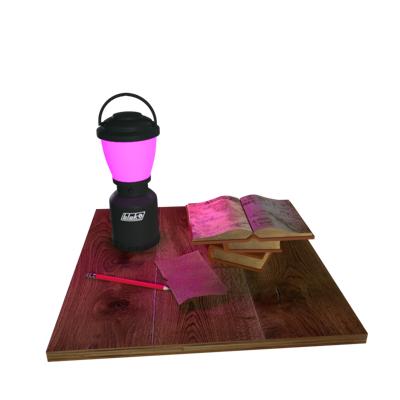} & \includegraphics[width=.33\linewidth]{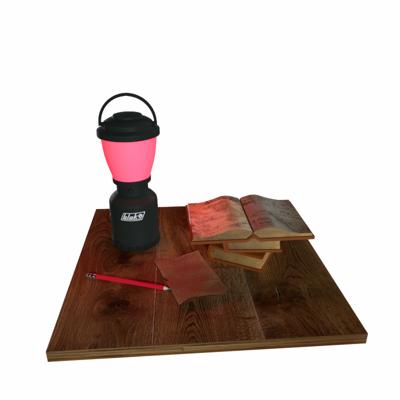}  & \includegraphics[width=.33\linewidth]{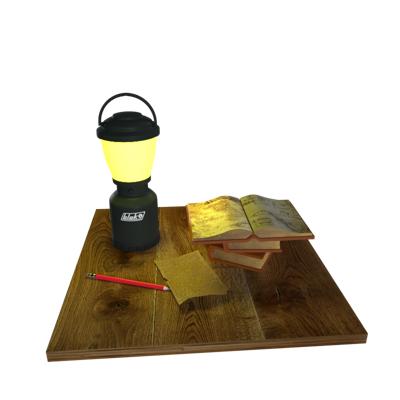}  \\
        \vspace{-10pt}
        \includegraphics[width=.33\linewidth]{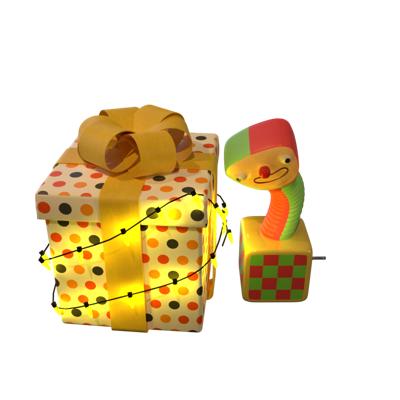} & \includegraphics[width=.33\linewidth]{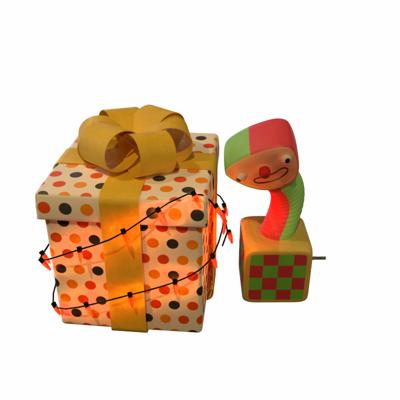}  & \includegraphics[width=.33\linewidth]{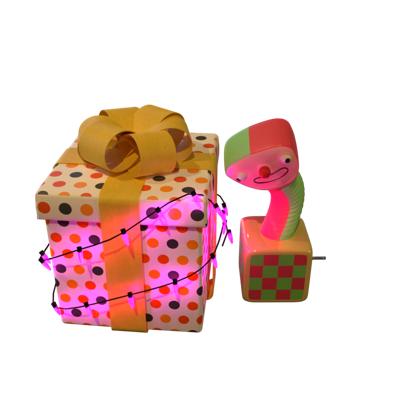}  \\
        \includegraphics[width=.33\linewidth]{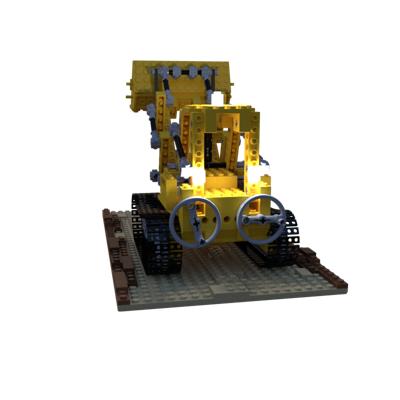} & \includegraphics[width=.33\linewidth]{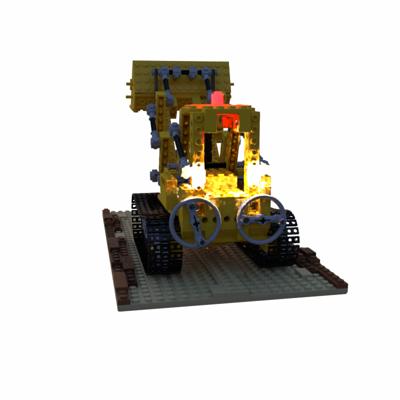} & \includegraphics[width=.33\linewidth]{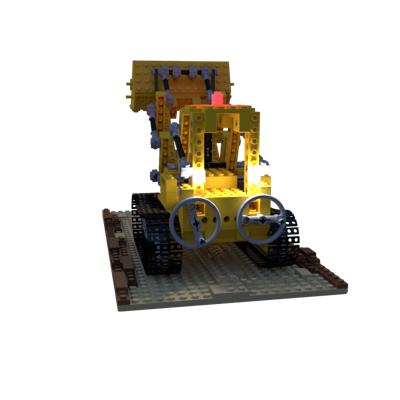} \\
    \end{tabular}
    \vspace{-20pt}
    \caption{Failure cases for editing scene illumination using the radiance fine-tuning method.}
    \label{fig:fail}
    \vspace{-10pt}
\end{figure}

We also present failure cases in scene editing, discussing the limitations of the radiance fine-tuning method for re-lighting in \S4.5 of the main paper.
While \modelname effectively reconstructs and manipulates emissive sources, the radiance fine-tuning method for re-lighting has its limitations.
These are depicted in Fig.~\ref{fig:fail}, where we note that LTS learning-based radiance fine-tuning may be constrained to color adjustments within the training spectrum.
In other words, using the LTS loss to transfer radiance within light transport segments may be weak in representing new colors that traverse unobserved light paths during training.
For example, it can shift colors from yellow to green but not to blue.
Additionally, the network's inherent smoothness capability may introduce illumination inaccuracies.
In the last row in Fig.~\ref{fig:fail}, changing only the top emissive source to red inadvertently affects the bulldozer's lower ceiling.

\begin{figure}[t]
    \centering
    \scriptsize
    \renewcommand{\arraystretch}{3} 
    \begin{tabular}{@{} *{4}{c@{\hspace{2pt}}} @{}}
        \multicolumn{1}{c}{Image}                                                                            & \multicolumn{1}{c}{Intensity Edit}                                                                                 & \multicolumn{1}{c}{Color Edit}                                                                                     & \multicolumn{1}{c}{Intensity \& Color Edit}                                                                         \\
        \includegraphics[valign=m,width=.24\linewidth]{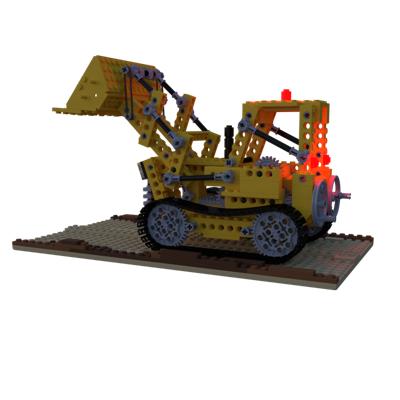}      & \includegraphics[valign=m,width=.24\linewidth]{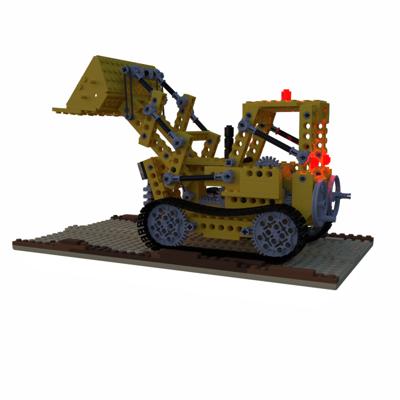}      & \includegraphics[valign=m,width=.24\linewidth]{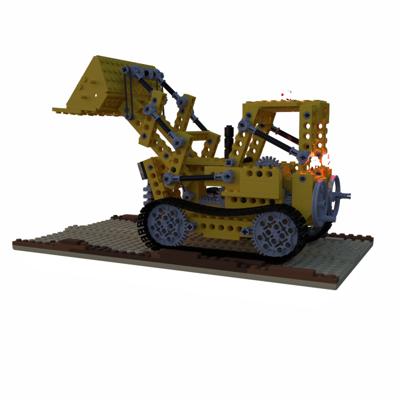}      & \includegraphics[valign=m,width=.24\linewidth]{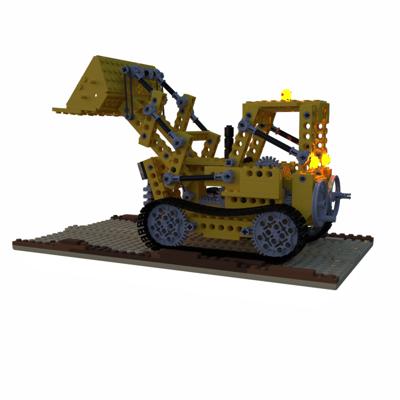}      \\
        \includegraphics[valign=m,width=.24\linewidth]{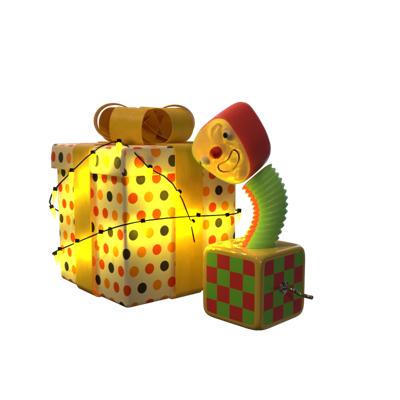}      & \includegraphics[valign=m,width=.24\linewidth]{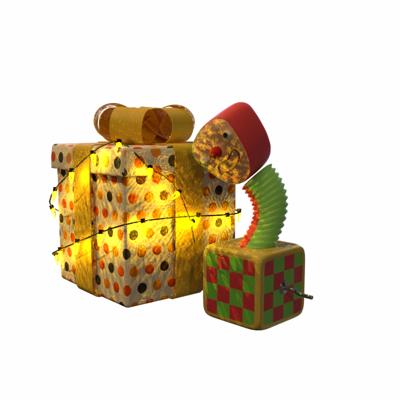}      & \includegraphics[valign=m,width=.24\linewidth]{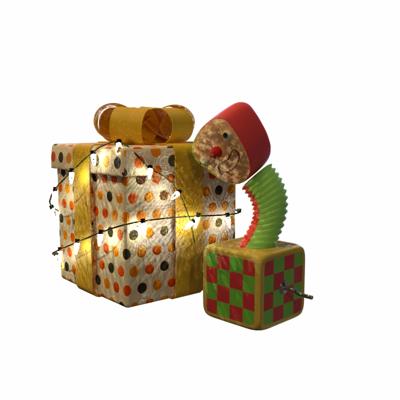}      & \includegraphics[valign=m,width=.24\linewidth]{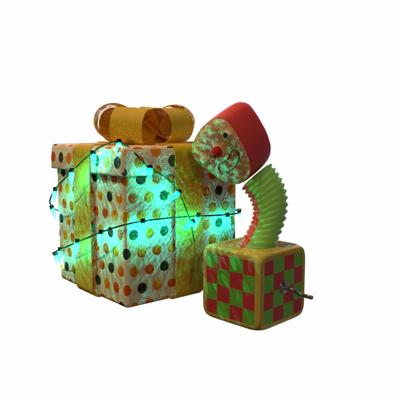}      \\
        \includegraphics[valign=m,width=.24\linewidth]{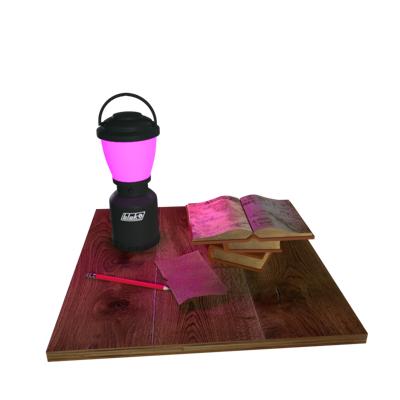}      & \includegraphics[valign=m,width=.24\linewidth]{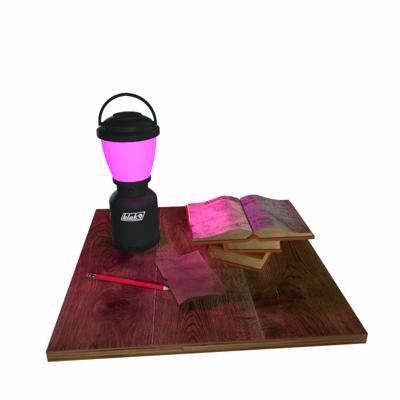}      & \includegraphics[valign=m,width=.24\linewidth]{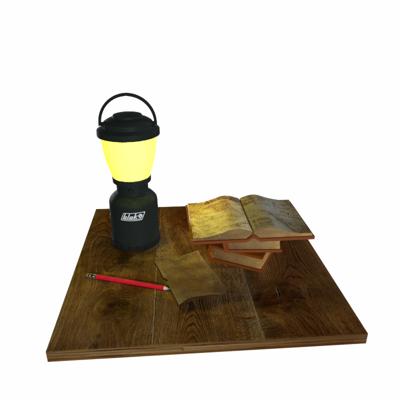}      & \includegraphics[valign=m,width=.24\linewidth]{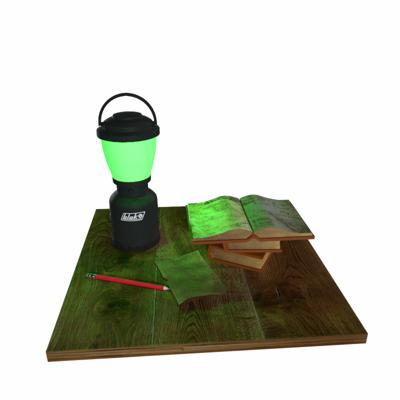}      \\
\end{tabular}
    \caption{Re-lighting scenes containing vivid-colored emissive sources by computing direct illumination from reconstructed emissive sources.}
    \label{fig:direct_c}
\end{figure}

Exploring alternative rendering approaches could address these issues.
We showcase scene editing results by computing direct illumination from emissive sources in Fig.~\ref{fig:direct_c}, enabling changes to any colors.
Reconstructing emissive sources using \modelname, then extracting emission texture maps to use rendering engines like Blender~\cite{blender} or Mitsuba~\cite{Mitsuba3} is also promising.
Howver, the texture map extraction in NeRF-like methods often faces severe UV atlas fragmentation.
Recent methods like Nuvo~\cite{srinivasan2023nuvo} offer some hope for feasible emission texture editing.
We consider these avenues for future exploration



\begin{figure*}[t]
    \centering
    \scriptsize
    \renewcommand{\arraystretch}{3} 
    \setlength{\tabcolsep}{10pt}
    \begin{minipage}{\linewidth}
        \begin{tabular}{@{} >{\centering\arraybackslash}c : >{\centering\arraybackslash}c@{}}
            \includegraphics[width=.48\linewidth]{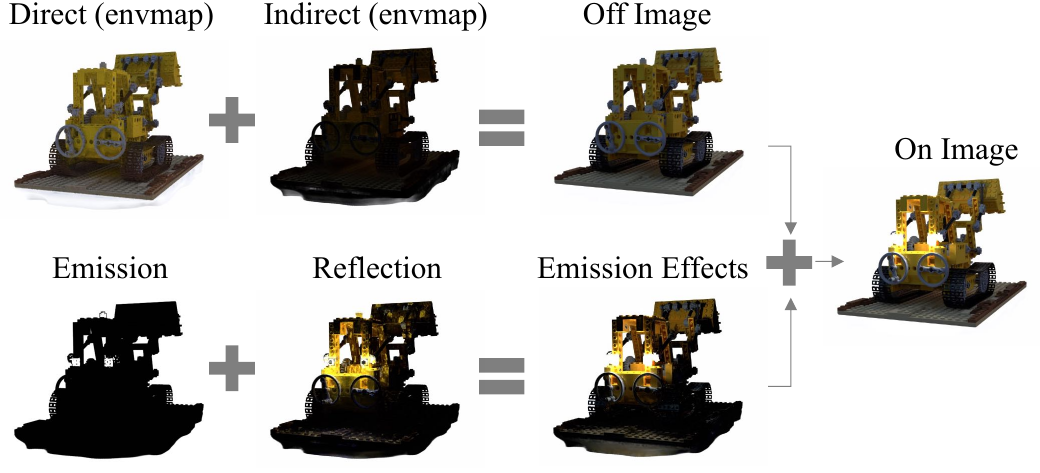} & \includegraphics[width=.48\linewidth]{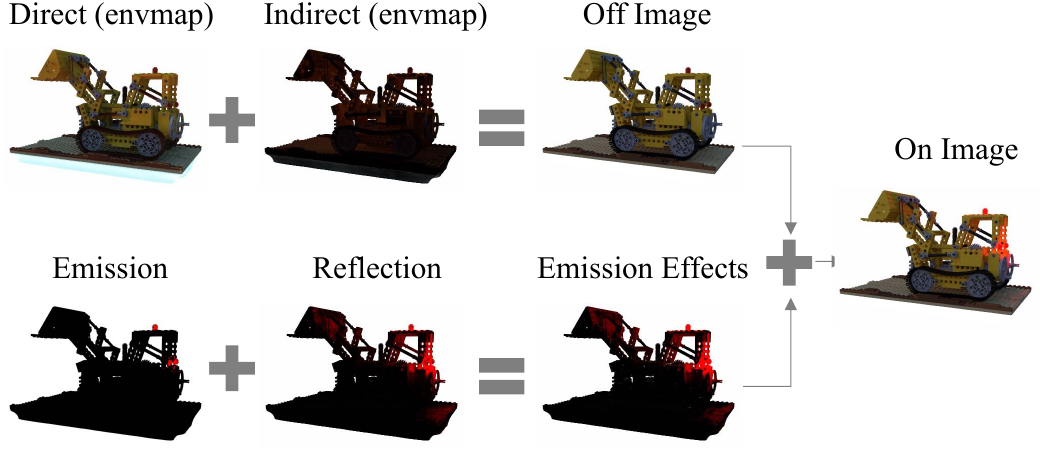} \\
        \end{tabular}
    \end{minipage}
    \begin{minipage}{\linewidth}
        \begin{tabular}{@{} >{\centering\arraybackslash}c : >{\centering\arraybackslash}c@{}}
            \includegraphics[width=.48\linewidth]{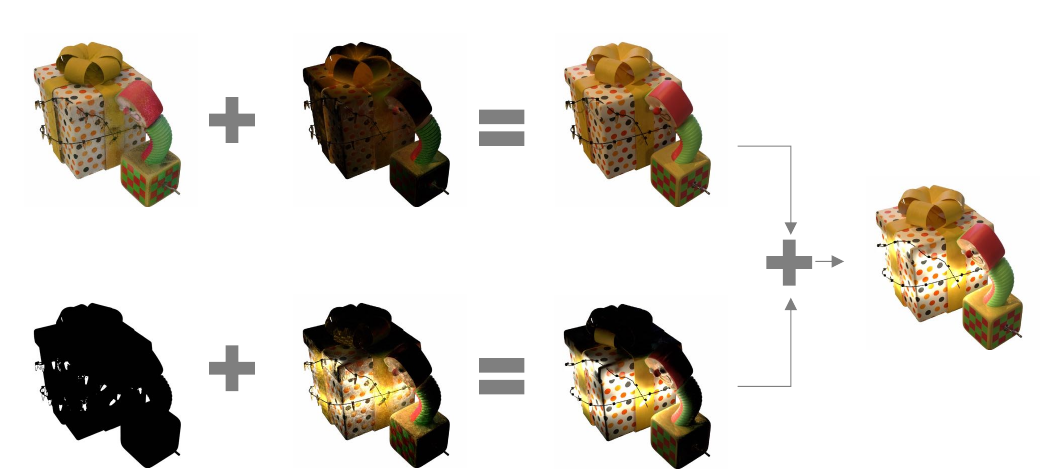} & \includegraphics[width=.48\linewidth]{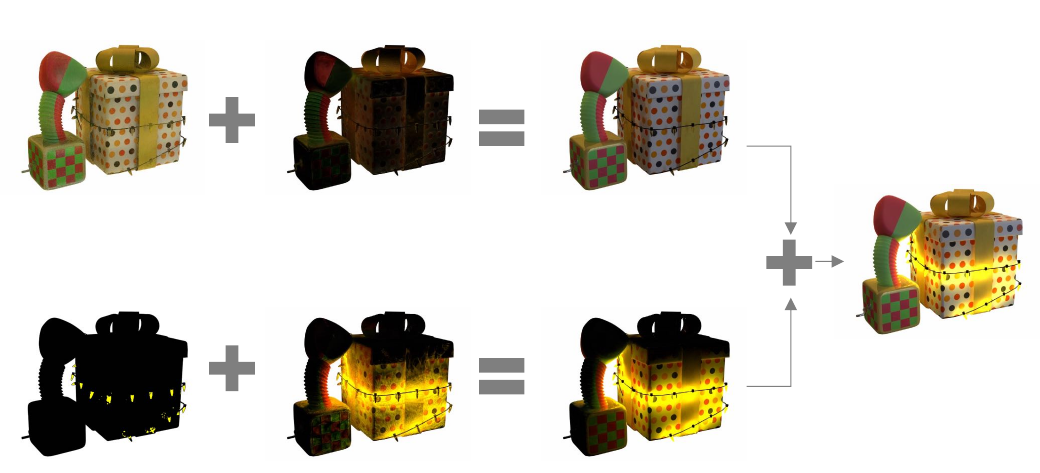} \\
        \end{tabular}
    \end{minipage}
    \begin{minipage}{\linewidth}
        \begin{tabular}{@{} >{\centering\arraybackslash}c : >{\centering\arraybackslash}c@{}}
            \includegraphics[width=.48\linewidth]{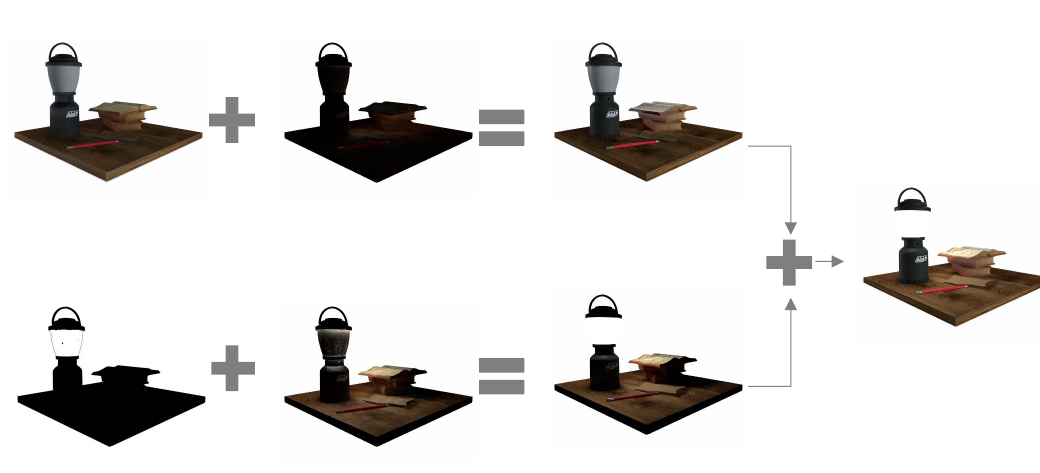} & \includegraphics[width=.48\linewidth]{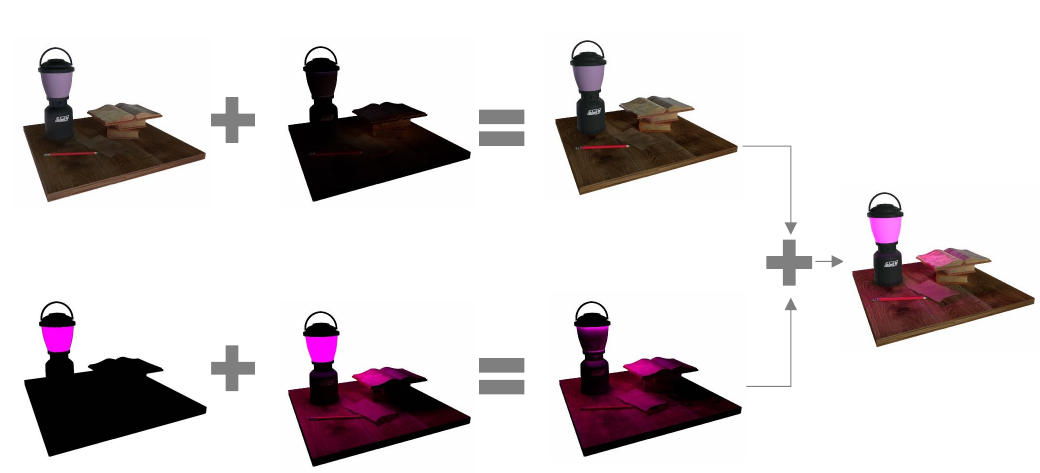} \\
        \end{tabular}
    \end{minipage}
    \begin{minipage}{\linewidth}
        \begin{tabular}{@{} >{\centering\arraybackslash}c : >{\centering\arraybackslash}c@{}}
            \includegraphics[width=.48\linewidth]{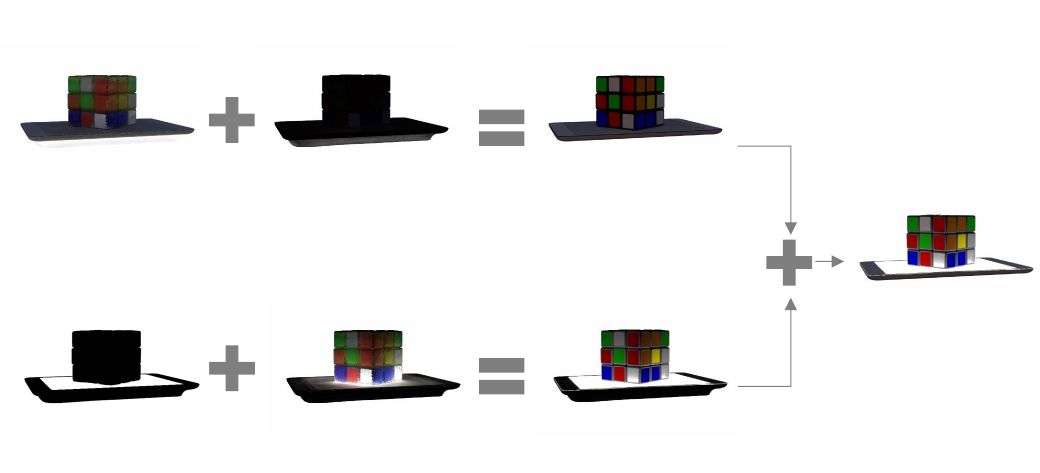} & \includegraphics[width=.48\linewidth]{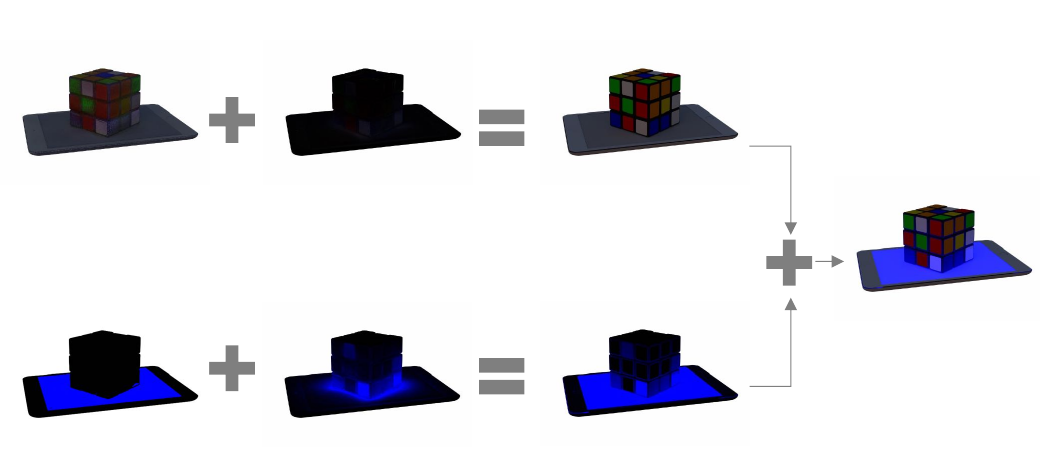} \\
        \end{tabular}
    \end{minipage}
    \begin{minipage}{\linewidth}
        \begin{tabular}{@{} >{\centering\arraybackslash}c : >{\centering\arraybackslash}c@{}}
            \includegraphics[width=.48\linewidth]{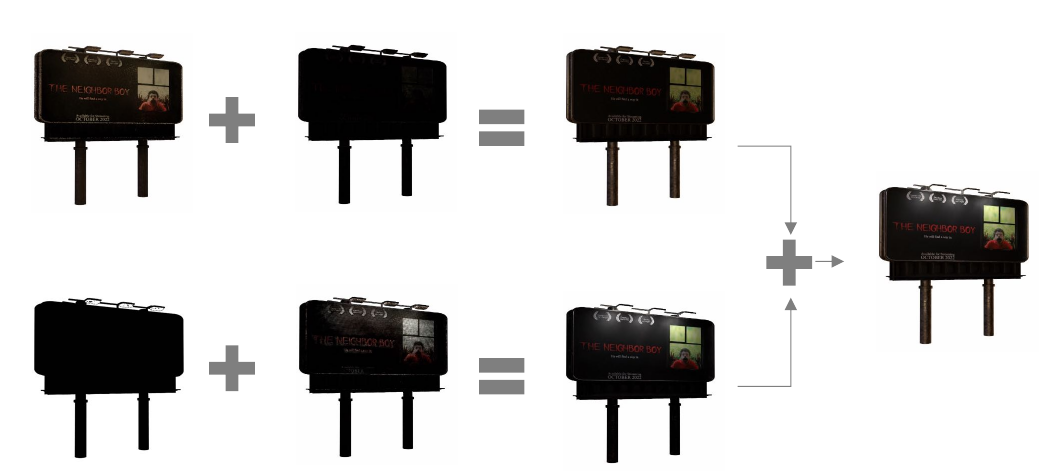} & \includegraphics[width=.48\linewidth]{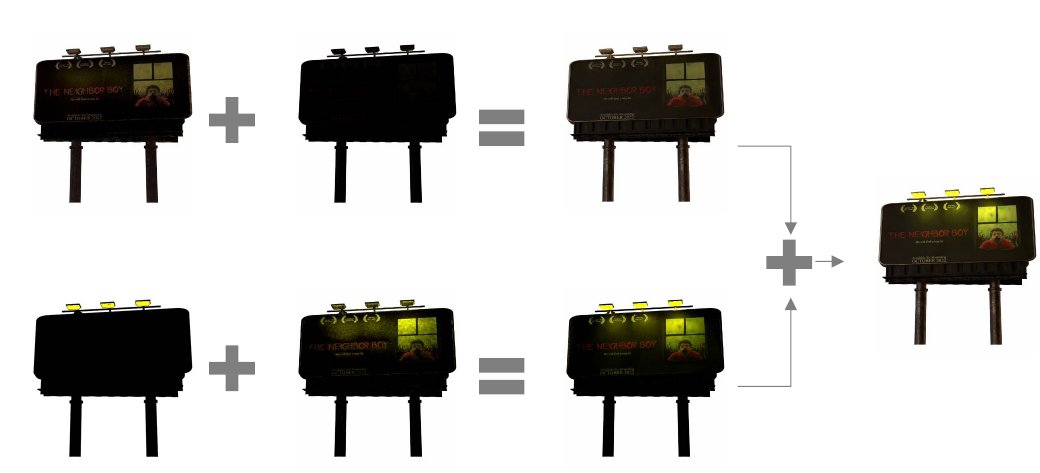} \\
        \end{tabular}
    \end{minipage}

    \caption{Illumination decomposition results. Left: scenes with white-colored, Right: scenes with vivid-colored emissive sources.
    }
    \label{fig:illum_decomp_1}
\end{figure*}

\begin{table*}
    \centering
    \scriptsize
    \setlength{\tabcolsep}{5pt}
    \begin{tabular}{@{}c*{15}{>{\centering\arraybackslash}c}>{\centering\arraybackslash}c}
                   & \multicolumn{16}{c}{White colored}                                                                                                                                                                                                                                                                                         \\
        \toprule
                   & \multicolumn{2}{c}{Lego (C)}       & \multicolumn{2}{c}{Lego (I)} & \multicolumn{2}{c}{Gift} & \multicolumn{2}{c}{Book} & \multicolumn{2}{c}{Cube} & \multicolumn{2}{c}{Billboard (C)} & \multicolumn{2}{c}{Billboard (I)} & \multicolumn{2}{c}{Balls}                                                                     \\
        \cmidrule(lr){2-3} \cmidrule(lr){4-5} \cmidrule(lr){6-7} \cmidrule(lr){8-9} \cmidrule(lr){10-11}\cmidrule(lr){12-13}\cmidrule(lr){14-15}\cmidrule(lr){16-17}
                   & PSNR                               & LPIPS                        & PSNR                     & LPIPS                    & PSNR                     & LPIPS                             & PSNR                              & LPIPS                     & PSNR  & LPIPS  & PSNR  & LPIPS  & PSNR  & LPIPS  & PSNR  & LPIPS  \\
        NV         & 37.77                              & 0.0082                       & 37.77                    & 0.0082                   & 37.72                    & 0.0060                            & 44.95                             & 0.0032                    & 43.60 & 0.0022 & 36.23 & 0.0109 & 36.23 & 0.0109 & 32.49 & 0.0190 \\
        NV + I     & 32.53                              & 0.0175                       & 29.50                    & 0.0261                   & 27.27                    & 0.0163                            & 30.29                             & 0.0166                    & 31.47 & 0.0097 & 29.50 & 0.0188 & 30.31 & 0.0216 & 29.03 & 0.0281 \\
\bottomrule
    \end{tabular}
    \caption{Editing performance on scenes with white-colored emissive sources by using the fine-tuning method. (C) denotes collective adjustments of emissive sources, while (I) represents individual adjustments of emissive sources. NV denotes novel view synthesis, I denotes intensity editing, and C denotes color editing. A higher PSNR or lower LPIPS value is better.}
    \label{tab:white_finetune}
\end{table*}

\begin{table*}
    \centering
    \scriptsize
    \setlength{\tabcolsep}{5pt}
    \begin{tabular}{@{}c*{15}{>{\centering\arraybackslash}c}>{\centering\arraybackslash}c}
                   & \multicolumn{16}{c}{White colored}                                                                                                                                                                                                                                                                                         \\
        \toprule
                   & \multicolumn{2}{c}{Lego (C)}       & \multicolumn{2}{c}{Lego (I)} & \multicolumn{2}{c}{Gift} & \multicolumn{2}{c}{Book} & \multicolumn{2}{c}{Cube} & \multicolumn{2}{c}{Billboard (C)} & \multicolumn{2}{c}{Billboard (I)} & \multicolumn{2}{c}{Balls}                                                                     \\
        \cmidrule(lr){2-3} \cmidrule(lr){4-5} \cmidrule(lr){6-7} \cmidrule(lr){8-9} \cmidrule(lr){10-11}\cmidrule(lr){12-13}\cmidrule(lr){14-15}\cmidrule(lr){16-17}
                   & PSNR                               & LPIPS                        & PSNR                     & LPIPS                    & PSNR                     & LPIPS                             & PSNR                              & LPIPS                     & PSNR  & LPIPS  & PSNR  & LPIPS  & PSNR  & LPIPS  & PSNR  & LPIPS  \\
        NV         & 37.77                              & 0.0082                       & 37.77                    & 0.0082                   & 37.72                    & 0.0060                            & 44.95                             & 0.0032                    & 43.60 & 0.0022 & 36.23 & 0.0109 & 36.23 & 0.0109 & 32.49 & 0.0190 \\
        NV + I     & 27.77                              & 0.0329                       & 29.00                    & 0.0293                   & 22.77                    & 0.0461                            & 28.85                             & 0.0327                    & 24.71 & 0.0382 & 26.49 & 0.0422 & 31.14 & 0.0249 & 28.41 & 0.0411 \\
\bottomrule
    \end{tabular}
    \caption{Editing performance on scenes with white-colored emissive sources by computing direct illumination from reconstructed emissive sources. (C) denotes collective adjustments of emissive sources, while (I) represents individual adjustments of emissive sources.  NV denotes novel view synthesis, I denotes intensity editing, and C denotes color editing. A higher PSNR or lower LPIPS value is better.}
    \label{tab:white_direct}
\end{table*}

\begin{table*}
    \centering
    \scriptsize
    \begin{tabular}{@{}c*{11}{>{\centering\arraybackslash}c}>{\centering\arraybackslash}c}
                   & \multicolumn{12}{c}{Vivid colored}                                                                                                                                                                                                     \\
        \toprule
                   & \multicolumn{2}{c}{Lego (C) }      & \multicolumn{2}{c}{Gift} & \multicolumn{2}{c}{Book} & \multicolumn{2}{c}{Cube} & \multicolumn{2}{c}{Billboard (C)} & \multicolumn{2}{c}{Balls}                                                    \\
        \cmidrule(lr){2-3} \cmidrule(lr){4-5} \cmidrule(lr){6-7} \cmidrule(lr){8-9} \cmidrule(lr){10-11}\cmidrule(lr){12-13}
                   & PSNR                               & LPIPS                    & PSNR                     & LPIPS                    & PSNR                              & LPIPS                     & PSNR  & LPIPS  & PSNR  & LPIPS  & PSNR  & LPIPS  \\
        \midrule
        NV         & 39.76                              & 0.0062                   & 38.31                    & 0.0055                   & 44.97                             & 0.0033                    & 45.13 & 0.0016 & 34.47 & 0.0165 & 32.78 & 0.0180 \\
        NV + I     & 35.00                              & 0.0154                   & 28.70                    & 0.0163                   & 32.86                             & 0.0141                    & 35.54 & 0.0072 & 28.56 & 0.0288 & 30.78 & 0.0231 \\
        NV + C     & 22.80                              & 0.0916                   & 26.00                    & 0.0312                   & 29.11                             & 0.0371                    & 22.50 & 0.0379 & 27.82 & 0.0347 & 30.05 & 0.0259 \\
\bottomrule
    \end{tabular}
    \caption{Editing performance on scenes with vivid-colored emissive sources by using the fine-tuning method. (C) denotes collective adjustments of emissive sources.  NV denotes novel view synthesis, I denotes intensity editing, and C denotes color editing. A higher PSNR or lower LPIPS value is better.}
    \label{tab:vivid_finetune}
\end{table*}

\begin{table*}
    \centering
    \scriptsize
    \begin{tabular}{@{}c*{11}{>{\centering\arraybackslash}c}>{\centering\arraybackslash}c}
                   & \multicolumn{12}{c}{Vivid colored}                                                                                                                                                                                                     \\
        \toprule
                   & \multicolumn{2}{c}{Lego (C) }      & \multicolumn{2}{c}{Gift} & \multicolumn{2}{c}{Book} & \multicolumn{2}{c}{Cube} & \multicolumn{2}{c}{Billboard (C)} & \multicolumn{2}{c}{Balls}                                                    \\
        \cmidrule(lr){2-3} \cmidrule(lr){4-5} \cmidrule(lr){6-7} \cmidrule(lr){8-9} \cmidrule(lr){10-11}\cmidrule(lr){12-13}
                   & PSNR                               & LPIPS                    & PSNR                     & LPIPS                    & PSNR                              & LPIPS                     & PSNR  & LPIPS  & PSNR  & LPIPS  & PSNR  & LPIPS  \\
        \midrule
        NV         & 39.76                              & 0.0062                   & 38.31                    & 0.0055                   & 44.97                             & 0.0033                    & 45.13 & 0.0016 & 34.47 & 0.0165 & 32.78 & 0.0180 \\
        NV + I     & 29.21                              & 0.0330                   & 24.97                    & 0.0378                   & 31.78                             & 0.0216                    & 27.44 & 0.0314 & 28.19 & 0.0356 & 27.20 & 0.0352 \\
        NV + C     & 27.41                              & 0.0330                   & 26.49                    & 0.0318                   & 33.95                             & 0.0198                    & 28.81 & 0.0260 & 28.51 & 0.0326 & 29.57 & 0.0301 \\
\bottomrule
    \end{tabular}
    \caption{Editing performance on scenes with vivid-colored emissive sources by computing direct illumination from reconstructed emissive sources. (C) denotes collective adjustments of emissive sources.  NV denotes novel view synthesis, I denotes intensity editing, and C denotes color editing. A higher PSNR or lower LPIPS value is better.}
    \label{tab:vivid_direct}
\end{table*}

\end{document}